\documentclass{article} 

\usepackage{amsmath,amsfonts,bm}

\def\eqref#1{equation~\ref{#1}}

\def\1{\bm{1}}

\DeclareMathAlphabet{\mathsfit}{\encodingdefault}{\sfdefault}{m}{sl}
\SetMathAlphabet{\mathsfit}{bold}{\encodingdefault}{\sfdefault}{bx}{n}

\usepackage{iclr2025_conference,times}

\usepackage[toc,page,header]{appendix}
\usepackage{minitoc}  

\usepackage{amssymb}  

\usepackage{amsthm}  

\usepackage{algorithm}
\usepackage{algpseudocode}
\usepackage{multicol}

\usepackage{booktabs}
\usepackage{multirow}
\usepackage{tabu}

\usepackage{enumitem}  

\usepackage{thmtools}  
\usepackage{thm-restate}

\usepackage[notquote]{hanging}

\usepackage{hyperref} 
\hypersetup{
    colorlinks=true,
    linkcolor=RedViolet!70!black,
    filecolor=Green,
    citecolor=Blue,
    urlcolor=OliveGreen,
}

\usepackage{graphicx}

\usepackage{diagbox}  

\usepackage{wrapfig} 

\usepackage[nameinlink]{cleveref}  

\usepackage{dblfloatfix}

\makeatletter
\newtheoremstyle{indented}
  {3pt}
  {3pt}
  {\addtolength{\@totalleftmargin}{5.0em}
   \addtolength{\linewidth}{-3.0em}  
   \parshape 1 1.5em \linewidth}
  {}
  {\bfseries}
  {.}
  {.8em} 
  {}

\newtheoremstyle{problem}
  {8pt} 
  {8pt} 
  {
    \parshape 1 1.5em \linewidth
    } 
  {} 
  {\bfseries}
  {} 
  {0.em plus .1em minus .1em} 
  {\thmname{#1} \thmnumber{#2} --- \thmnote{\textnormal{\scshape #3}}.} 

\newtheoremstyle{comp-problem}
  {4pt} 
  {-3pt} 
  {
    \parshape 1 0.5em \linewidth  
    } 
  {} 
  {\bfseries}
  {} 
  {0.em plus .1em minus .1em} 
  {\thmname{#1} \thmnumber{#2}.  \thmnote{\textnormal{\scshape #3}}} 
  
\newenvironment{compprob}[2]{
    \begin{comp-problem}[\textsc{#1 (#2)}]
            \noindent \begin{itemize}[labelwidth=30pt,align=left,leftmargin=45pt,labelindent=-15pt,itemindent=0pt,nosep,noitemsep, rightmargin=20pt]
            \setlength\itemsep{0.15em}
            \vspace{-5pt}
            \item[\textit{Input}:]
    \newcommand{\nextpart}{
         \item[\textit{Output}:]
        }
    }{%
        \end{itemize}
    \end{comp-problem}%
}

\newtheoremstyle{remark}
  {3pt}
  {3pt}
  {\addtolength{\@totalleftmargin}{2.5em}
   \addtolength{\linewidth}{-1.0em}
   \parshape 1 0.5em \linewidth}
  {}
  {\itshape}
  {}
  {.8em} 
  {}
  
\makeatother

\theoremstyle{indented}
\newtheorem{theorem}{Theorem}

\theoremstyle{indented}

\theoremstyle{indented}
\newtheorem{lemma}{Lemma}

\theoremstyle{indented}
\newtheorem{conjecture}{Conjecture}

\theoremstyle{indented}
\newtheorem{definition}{Definition}

\theoremstyle{problem}

\theoremstyle{comp-problem}
\newtheorem{comp-problem}{Problem}
\theoremstyle{remark}
\newtheorem{remark}{Remark}

\theoremstyle{indented}
\newtheorem{observation}{Observation}

\makeatletter
\def\namedlabel#1#2{\begingroup
   \def\@currentlabel{#2}%
   \label{#1}\endgroup
}
\makeatother

\makeatother

\let\mb\mathbf  
\let\mc\mathcal 

\newcommand{\la}{\langle}
\newcommand{\ra}{\rangle} 

\usepackage{url}

\title{The Computational Complexity of \\ Circuit Discovery for Inner Interpretability}

\author{Federico Adolfi \\
ESI Neuroscience, Max-Planck Society \\
\& University of Bristol\\
\texttt{fede.adolfi@bristol.ac.uk} \\
\And
Martina G. Vilas \\
Department of Computer Science \\
Goethe University Frankfurt \\
\texttt{martinagvilas@em.uni-frankfurt.de} \\
\And
Todd Wareham \\
Department of Computer Science \\
Memorial University of Newfoundland\\
\texttt{harold@mun.ca}
}

\iclrfinalcopy 

\begin{document}

\maketitle

\vspace{-5pt}
\begin{abstract}
Many proposed applications of neural networks in machine learning, cognitive/brain science, and society hinge on the feasibility of inner interpretability via circuit discovery. This calls for empirical and theoretical explorations of viable algorithmic options. Despite advances in the design and testing of heuristics, there are concerns about their scalability and faithfulness at a time when we lack understanding of the complexity properties of the problems they are deployed to solve. To address this, we study circuit discovery with classical and parameterized computational complexity theory: (1) we describe a conceptual scaffolding to reason about circuit finding queries in terms of affordances for description, explanation, prediction and control; (2) we formalize a comprehensive set of queries for mechanistic explanation, and propose a formal framework for their analysis; (3) we use it to settle the complexity of many query variants and relaxations of practical interest on multi-layer perceptrons. Our findings reveal a challenging complexity landscape. Many queries are intractable, remain fixed-parameter intractable relative to model/circuit features, and inapproximable under additive, multiplicative, and probabilistic approximation schemes. To navigate this landscape, we prove there exist transformations to tackle some of these hard problems with better-understood heuristics, and prove the tractability or fixed-parameter tractability of more modest queries which retain useful affordances. This framework allows us to understand the scope and limits of interpretability queries, explore viable options, and compare their resource demands on existing and future architectures.

\end{abstract}


\setlength{\tabcolsep}{2pt}
\renewcommand{\arraystretch}{1.2}  
\newcolumntype{P}[1]{>{\centering\arraybackslash}p{#1}}  
\newcommand{\CCG}{\cellcolor{lightgray!40!white}}
\newcommand{\CCB}{\cellcolor{CadetBlue!30!white}}
\newcommand{\CCO}{\cellcolor{Orange!20!white}}
\newcommand{\CCR}{\cellcolor{BrickRed!20!white}}
\newcommand{\CCM}{\cellcolor{OliveGreen!30!white}}
\newcommand{\CCF}{\cellcolor{Red!50!white}}
\newcommand{\CCFF}{\cellcolor{Red!30!white}}
\newcommand{\CCFFF}{\cellcolor{Red!10!white}}
\newcommand{\CCP}{\cellcolor{Purple!20!white}}
\newcommand{\CCT}{\cellcolor{Maroon!40!white}}
\newcommand{\CCY}{\cellcolor{YellowOrange!80!white}}

\doparttoc 
\faketableofcontents 

\section{Introduction}
\label{sec:intro}

As artificial neural networks (ANNs) grow in size and capabilities, \textit{Inner Interpretability} --- an emerging field tasked with explaining their inner workings \citep{raukerTransparentAISurvey2023a, vilasPositionInnerInterpretability2024} --- attempts to devise scalable, automated procedures to understand systems mechanistically.
Many proposed applications of neural networks in machine learning, cognitive and brain sciences, and society, hinge on the feasibility of inner interpretability.
For instance, we might have to rely on interpretability methods to improve system safety \citep{bereska2024mechanistic}, detect and control vulnerabilities \citep{garcia-carrascoDetectingUnderstandingVulnerabilities2024},
prune for efficiency \citep{hookerWhatCompressedDeep2021}, find and use task subnetworks \citep{zhangInstillingInductiveBiases2024a}, explain internal concepts underlying decisions \citep{leeNeuralActivationsConcepts2023}, experiment with neuro-cognitive models of language, vision, etc. \citep{lindsayGroundingNeuroscienceBehavioral2024,lindsayTestingMethodsNeural2023,pavlickSymbolsGroundingLarge2023}, describe determinants of ANN-brain alignment \citep{feghhiWhatAreLarge2024a,ootaSpeechLanguageModels2023}, improve architectures, and extract domain insights \citep{raukerTransparentAISurvey2023a}.
We will have to solve different instances of these interpretability problems, ideally automatically, for increasingly large models.
We therefore need efficient interpretability procedures, and this requires empirical and theoretical explorations of viable algorithmic options.

\textbf{Circuit discovery and its challenges.} Since top-down approaches to inner interpretability \citep[see][]{vilasPositionInnerInterpretability2024} work their way down from high-level concepts or algorithmic hypotheses \citep{lieberumDoesCircuitAnalysis2023}, there is interest in a complementary bottom-up methodology: \textit{circuit discovery} \citep[see][]{shiHypothesisTestingCircuit2024,tiggesLLMCircuitAnalyses2024}.
It starts from neuron- and circuit-level isolation or description \citep[e.g.,][]{hoang-xuanLLMassistedConceptDiscovery2024, leporiUncoveringCausalVariables2023} and attempts to build up higher-level abstractions.
The motivation is the \textit{circuit hypothesis}: models might implement their capabilities via small subnetworks \citep{shiHypothesisTestingCircuit2024}.
Advances in the design and testing of interpretability heuristics \citep[see][]{shiHypothesisTestingCircuit2024,tiggesLLMCircuitAnalyses2024} come alongside interest in the automation of circuit discovery  \citep[e.g.,][]{conmyAutomatedCircuitDiscovery2023a, ferrandoInformationFlowRoutes2024, syedAttributionPatchingOutperforms2023} and concerns about its feasibility \citep{vossVisualizingWeights2021, raukerTransparentAISurvey2023a}.
One challenge is scaling up methods to larger networks, more naturalistic datasets, and more complex tasks \citep[e.g.,][]{lieberumDoesCircuitAnalysis2023, marksSparseFeatureCircuits2024}, given their manual-intensive search over large spaces \citep{vossVisualizingWeights2021}.
A related issue is that current heuristics, though sometimes promising \citep[e.g.,][]{merulloCircuitComponentReuse2024}, often yield discrepant results \citep[see e.g.,][]{shiHypothesisTestingCircuit2024,niuWhatDoesKnowledge2023,zhangBestPracticesActivation2023}.
They often find circuits that are not functionally faithful \citep{yuFunctionalFaithfulnessWild2024} or lack the expected affordances \citep[e.g., effects on behavior;][]{shiHypothesisTestingCircuit2024}.
This questions whether certain localization methods yield results that inform editing \citep{haseDoesLocalizationInform2023}, and vice versa \citep{wangDoesEditingProvide2024}.
More broadly, we run into `interpretability illusions' \citep{friedmanInterpretabilityIllusionsGeneralization2024} when our simplifications (e.g., circuits) mimic the local input-output behavior of the system but lack global \textit{faithfulness} \citep{jacoviFaithfullyInterpretableNLP2020}.

\textbf{Exploring viable algorithmic options.} These challenges come at a time when, despite emerging theoretical frameworks \citep[e.g.,][]{vilasPositionInnerInterpretability2024, geigerCausalAbstractionTheoretical2024}, there are notable gaps in the formalization and analysis of the computational problems that interpretability heuristics attempt to solve \citep[see][\S 8]{wangDoesEditingProvide2024}.
Issues around scalability of circuit discovery and faithfulness have a natural formulation in the language of Computational Complexity Theory \citep{aroraComputationalComplexityModern2009,downeyFundamentalsParameterizedComplexity2013}.
A fundamental source of breakdown of scalability — which lack of faithfulness is one manifestation of — is the intrinsic resource demands of interpretability problems.
In order to design efficient and effective solutions, we need to understand the complexity properties of circuit discovery queries and the constraints that might be leveraged to yield the desired results.
Although experimental efforts have made promising inroads, the complexity-theoretic properties that naturally impact scalability and faithfulness remain open questions \citep[see e.g.,][\S 6C]{subercaseauxTESISPARAOPTAR}.
We settle them here by complementing these efforts with a systematic study of the computational complexity of circuit discovery for inner interpretability.
We present a framework that allows us to (a) understand the scope and limits of interpretability queries for description/explanation and prediction/control, (b) explore viable options, and (c) compare their resource demands among existing and future architectures.

\vspace{-10pt}
\subsection{Contributions}

\begin{itemize}[nosep,noitemsep]
    \item We present a conceptual scaffolding to reason about circuit finding queries in terms of affordances for description, explanation, prediction and control.
    \item We formalize a comprehensive set of queries that capture mechanistic explanation, and propose a formal framework for their analysis.
    \item We use this framework to settle the complexity of many query variants, parameterizations, approximation schemes and relaxations of practical interest on multi-layer perceptrons, relevant to various architectures such as transformers. 
    \item We demonstrate how our proof techniques can also be useful to draw links between interpretability and explainability by using them to improve existing results on the latter.
\end{itemize}

\vspace{-10pt}

\subsection{Overview of results}

\begin{itemize}[nosep,noitemsep]
     \item We uncover a challenging complexity landscape (see \Cref{tbl:results}) where many queries are intractable (NP-hard, $\Sigma^p_2$-hard), remain fixed-parameter intractable (W[1]-hard) when constraining model/circuit features (e.g., depth), and are inapproximable under additive, multiplicative, and probabilistic approximation schemes.
     \item We prove there exist transformations to potentially tackle some hard problems (NP- vs. $\Sigma^p_2$-complete) with better-understood heuristics, and prove the tractability (PTIME) or fixed-parameter tractability (FPT) of other queries of interest, and we identify open problems.
     \item We describe a quasi-minimality property of ANN circuits and exploit it to generate tractable queries which retain useful affordances as well as efficient algorithms to compute them.
     \item We establish a separation between local and global query complexity. Together with quasi-minimality, this explains interpretability illusions of faithfulness observed in experiments.
\end{itemize}

\vspace{-8pt}
\subsection{Related work}
\vspace{-5pt}

This paper gives the first systematic exploration of the computational complexity of \textit{inner interpretability} problems.\footnote{
This work expands on \href{https://fedeadolfi.github.io}{FA}'s PhD dissertation at University of Bristol \citep{adolfiComputationalMetaTheoryCognitive2023,adolfiComplexityTheoreticLimitsPromises2024a}.
}
An adjacent area is the complexity analysis of \textit{explainability} problems \citep{bassanFormalXAIFormally2023, ordyniakParameterizedComplexityFinding2023}.
It differs from our work in its focus on \textit{input queries} --- aspects of the input that explain model decisions --- as we look at the inner workings of neural networks via \textit{circuit queries}.
\cite{barcelo_model_2020} study the explainability of multi-layer perceptrons compared to simpler models through a set of input queries.
\cite{bassanLocalVsGlobal2024a} extend this idea with a comparison between \textit{local} and \textit{global} explainability.
None of these works formalize or analyze circuit queries \citep[although][identifies it as an open problem]{subercaseauxTESISPARAOPTAR}; we adapt the local versus global distinction in our framework and show how our proof techniques can tighten some results on explainability queries.
\cite{ramaswamyAlgorithmicBarrierNeural2019} and \cite{adolfiResourceDemandsImplementationist2023} explore a small set of circuit queries and only on abstract biological networks modeled as general graphs, which cannot inform circuit discovery in ANNs.
Efforts in characterizing the complexity of learning neural networks \citep[e.g.,][]{songComplexityLearningNeural2017a, chenLearningDeepReLU2020, livniComputationalEfficiencyTraining2014} might eventually connect to our work, although a number of differences between the formalizations makes results in one area difficult to predict from those in the other.
Likewise, efforts to settle the complexity of finding small circuits consistent with a truth table \citep{hitchcockNPCompletenessMinimumCircuit2015} are currently too general to be applicable to interpretability problems.
More generally, we join efforts to build a solid theoretical foundation for interpretability \citep{bassanFormalXAIFormally2023,geigerCausalAbstractionTheoretical2024,vilasPositionInnerInterpretability2024}.

\vspace{-10pt}
\section{Mechanistic understanding of neural networks}
\vspace{-5pt}

Mechanistic understanding is a contentious topic \citep{rossCausationNeuroscienceKeeping2024a}, but for our purposes it will suffice to adopt a pragmatic perspective.
In many cases of practical interest, we want our interpretability methods to output objects that allow us to, in some limited sense, (1) describe or explain succinctly, and (2) control or predict precisely.
Such objects (e.g., circuits) should be ‘efficiently queriable’; they are often referred to as ``a way of making an explanation tractable'' \citep{caoExplanatoryModelsNeuroscience2023}.
Roughly, this means that we would like short descriptions (e.g., small circuits) with useful affordances (e.g., to readily answer questions and perform interventions of interest).
Circuits have the potential to fulfill these criteria \citep{olahZoomIntroductionCircuits2020}.
Here we preview some special circuits with useful properties which we formalize and analyze later on.
\Cref{tbl:afford} maps the main circuits we study to their corresponding affordances for description, explanation, prediction and control. Formal definitions of circuit queries are given alongside results in \Cref{sec:results} (see also \ref{app}).


\vspace{-12pt}

\definecolor{mg}{gray}{0.94}
\setlength{\tabcolsep}{2pt}
\begin{table}[h]
    \caption{Circuit affordances for description, explanation, prediction, and control.}
    \centering
    \small{
    \begin{tabular}{p{0.17\linewidth} | P{0.4\linewidth} | P{0.4\linewidth}}
        \toprule
            \multirow{2}{*}{\textbf{Circuit}} & \multicolumn{2}{c}{\textbf{Affordance}}\\
             \cmidrule{2-3}
             & \textit{Description / Explanation} & \textit{Prediction / Control}\\
        \toprule
            \rowcolor{mg} Sufficient Circuit & Which neurons suffice in isolation to cause a behavior? \textit{Minimum:} shortest description. & Inference in isolation. \textit{Minimal:} ablating any neuron breaks behavior of the circuit. \\
            Quasi-minimal Sufficient Circuit & Which neurons suffice in isolation to cause a behavior and which is a breaking point? & Ablating the breaking point breaks behavior of the circuit. \\
            \rowcolor{mg} Necessary Circuit & Which neurons are part of all circuits for a behavior? Key subcomputations? & Ablating the neurons breaks behavior of any sufficient circuit in the network. \\
            Circuit Ablation \& Clamping & Which neurons are necessary in the current configuration of the network? & Ablating/Clamping the neurons breaks  behavior of the network. \\
            \rowcolor{mg} Circuit Robustness & How much redundancy supports a behavior? Resilience to perturbations. & Ablating any set of neurons of size below threshold does not break behavior.  \\
            Patched Circuit & Which neurons drive a behavior in a given input context, i.e., are control nodes? & Patching neurons changes network behavior for inputs of interest. Steering; Editing. \\
            \rowcolor{mg} Quasi-minimal Patched Circuit & Which neurons can drive a behavior in a given input context and which neuron is a breaking point? & Patching neurons causes target behavior for inputs of interest; Unpatching breaking point breaks target behavior. \\
            Gnostic Neurons & Which neurons respond preferentially to a certain concept? & Concept editing; guided synthesis. \\
         \bottomrule
    \end{tabular}
    }

    \namedlabel{tbl:afford}{table of circuit affordances}
\end{table}





\section{Inner interpretability queries as computational problems}
\vspace{-5pt}

We model post-hoc interpretability queries on neural networks as computational problems in order to analyze their intrinsic complexity properties.
These \textit{circuit queries} also formalize criteria for desired circuits, including those appearing in the literature, such as `faithfulness', `completeness', and `minimality' \citep{wangInterpretabilityWildCircuit2022, yuFunctionalFaithfulnessWild2024}.



\begin{figure}[h]
    \centering
    \includegraphics[width=0.6\linewidth]{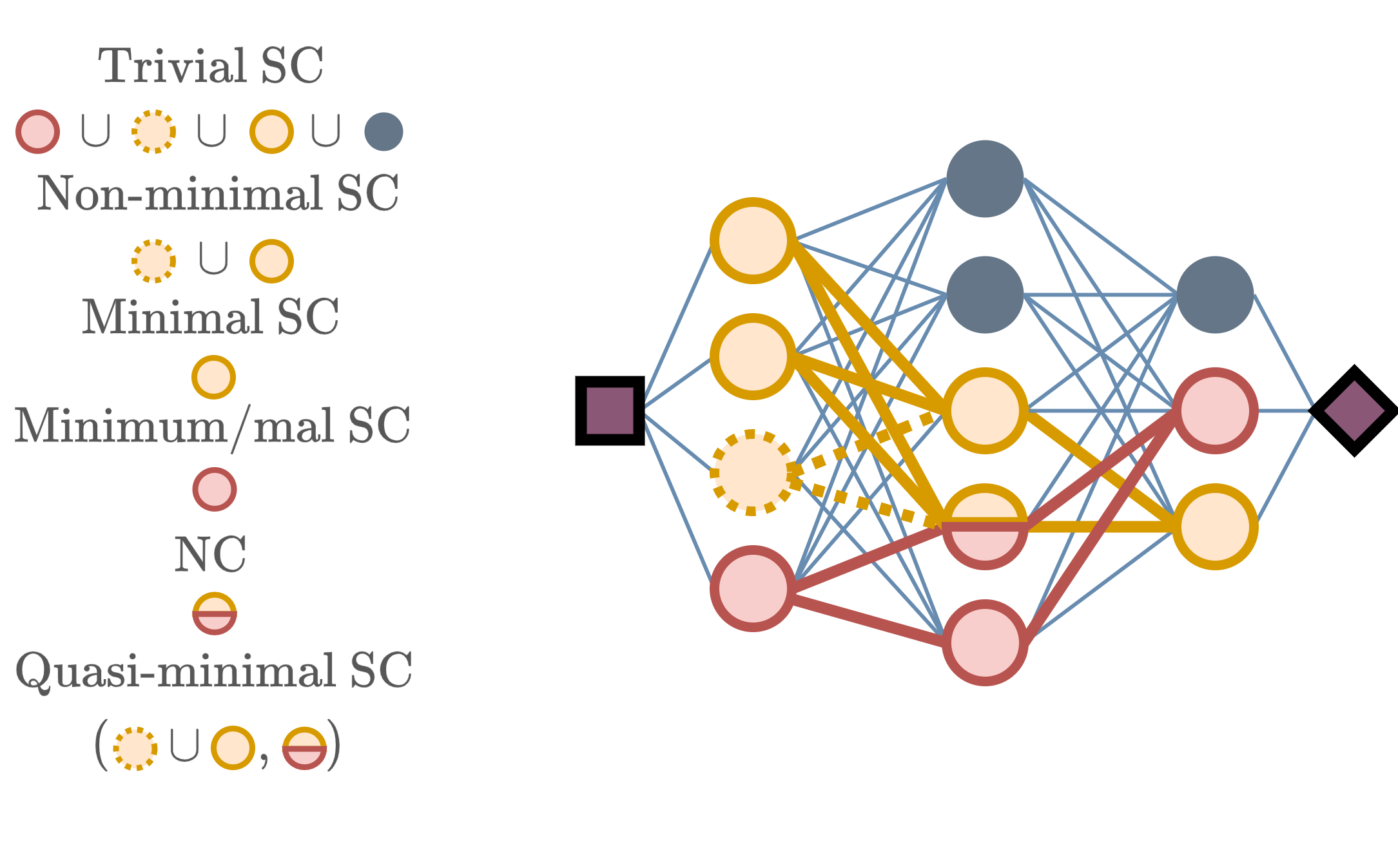}
    \vspace{-20pt}
    \caption{Relationships between circuit types. Sufficient Circuits (SCs) are faithful to the model. The entire network is a trivial SC. Necessary Circuits (NCs) are units shared by all minimal SCs. Quasi-minimal SCs contain a known breaking point (here, NC) and unknown superfluous units.}
    \label{fig:circuits}
\end{figure}

\textbf{Query variants: coverage, size and minimality.} The \textit{coverage} of a circuit is the domain over which it behaves in a certain way (e.g., faithful to the model's prediction).
\textit{Local} circuits do so over a finite set of known inputs and
\textit{global} circuits do so over all possible inputs.
The \textit{size} of a circuit is the number of neurons.
Some circuit queries require circuits of \textit{bounded} size whereas others leave the size \textit{unbounded}.
A circuit with a certain property (e.g., local sufficiency) is \textit{minimal} if there is no subset of its neurons that also has that property (cf. \textit{minimum} size among all such circuits present in the network; see \Cref{fig:circuits}).

\vspace{-10pt}

To fit our comprehensive suite of problems, we explain how to generate problem variants and later on only present one representative definition of each.

\begin{compprob}{\texttt{ProblemName}}{PN}
\namedlabel{prb:template}{\textsc{problem template}}
A multi-layer perceptron $\mc{M}$, {\small\texttt{CoverageIN}}, {\small\texttt{SizeIN}}.
\nextpart
A {\small\texttt{Property}} circuit $\mc{C}$ of  $\mc{M}$, {\small\texttt{SizeOUT}}, s.t. {\small\texttt{CoverageOUT}} $\mc{C}(\mb{x}) \! = \! \mc{M}(\mb{x})$, {\small\texttt{Suffix}}. 
\end{compprob}

\Cref{prb:template} and \Cref{tbl:template} illustrate how to generate problem variants using a template, and \texttt{ProblemName} = \textsc{Sufficient Circuit} as an example (e.g., the \texttt{Coverage[IN/OUT]} variables specify parts of the input/output description that vary according to whether the requested circuit must have global or local faithfulness).
Problem definitions will be given for \textit{search} (return specified circuits) or \textit{decision} (answer yes/no circuit queries) versions.
Others, including \textit{optimization} (return maximum/minimum-size circuits), can be generated by assigning variables.
Problems presented later on are obtained similarly.
We also explore various parameterizations, approximation schemes, and relaxations that we explain in the following sections as needed.

\vspace{-10pt}

\begin{table}[h]
    \caption{Generating query variants from problem templates. 
    }
    \centering
    \small{
    \begin{tabular}{p{0.15\linewidth} | P{0.14\linewidth} P{0.14\linewidth} P{0.10\linewidth} | P{0.14\linewidth}  P{0.14\linewidth} P{0.10\linewidth}}
        \toprule
             \multirow{3}{80pt}{\textbf{Description \newline variables}} & \multicolumn{6}{c}{\textbf{Query variants}} \\
        \cmidrule{2-7} 
              & \multicolumn{3}{c|}{\textit{Local}} & \multicolumn{3}{c}{\textit{Global}} \\
              \cmidrule(lr){2-4}  \cmidrule(lr){5-7}
               & \textit{Bounded} & \textit{Unbounded} &\textit{Optimal} & \textit{Bounded} & \textit{Unbounded} & \textit{Optimal} \\
        \toprule 
            \rowcolor{mg} \texttt{CoverageIN} &  an input $\mb{x}$ & an input $\mb{x}$ & an input $\mb{x}$ & \textcolor{lightgray}{``\_\_''} & \textcolor{lightgray}{``\_\_''} & \textcolor{lightgray}{``\_\_''} \\
            \texttt{CoverageOUT} & \textcolor{lightgray}{``\_\_''}  & \textcolor{lightgray}{``\_\_''} & \textcolor{lightgray}{``\_\_''} & $\forall_{\mb{x}}$ & $\forall_{\mb{x}}$ & $\forall_{\mb{x}}$ \\ 
            \rowcolor{mg} \texttt{SizeIN}  & int. $u \leq |\mc{M}|$ & \textcolor{lightgray}{``\_\_''} & \textcolor{lightgray}{``\_\_''} & int. $u \leq |\mc{M}|$ & \textcolor{lightgray}{``\_\_''} & \textcolor{lightgray}{``\_\_''} \\
            \texttt{SizeOUT} & size $|\mc{C}| \leq u$ & \textcolor{lightgray}{``\_\_''} & min. size & size $|\mc{C}| \leq u$ & \textcolor{lightgray}{``\_\_''} & min. size \\
            \rowcolor{mg} \texttt{Property} & minimal / \textcolor{lightgray}{``\_\_''} & minimal / \textcolor{lightgray}{``\_\_''} & \textcolor{lightgray}{``\_\_''} & minimal / \textcolor{lightgray}{``\_\_''} & minimal / \textcolor{lightgray}{``\_\_''} & \textcolor{lightgray}{``\_\_''} \\
            \texttt{Suffix} & if it exists, otherwise $\bot$ & \textcolor{lightgray}{``\_\_''} & \textcolor{lightgray}{``\_\_''} & if it exists, otherwise $\bot$ & \textcolor{lightgray}{``\_\_''} & \textcolor{lightgray}{``\_\_''} \\
         \bottomrule
    \end{tabular}
    }
    \namedlabel{tbl:template}{problem templates}
\end{table}


\subsection{Complexity analyses}
\vspace{-5pt}
\textbf{Classical and parameterized complexity.} We prove theorems about interpretability queries building on techniques from classical \citep{garey1979computers} and parameterized complexity \citep{downeyFundamentalsParameterizedComplexity2013}.
Given our limited knowledge of the problem space of interpretability, \textit{worst-case analysis} is appropriate to explore which problems might be solvable without requiring any additional assumptions \citep[e.g.,][]{bassanLocalVsGlobal2024a,barcelo_model_2020}, and experimental results suggest it captures a lower bound on real-world complexity \citep[e.g.,][]{friedmanInterpretabilityIllusionsGeneralization2024,shiHypothesisTestingCircuit2024,yuFunctionalFaithfulnessWild2024}.
Here we give a brief, informal overview of the main concepts underlying our analyses (see \ref{app} for extensive formal definitions).
We will explore beyond classical polynomial-time tractability (PTIME) by studying fixed-parameter tractability (FPT), a more novel and finer-grained look at the \textit{sources of complexity} of problems to test aspects that possibly make interpretability feasible in practice.
NP-hard queries are considered intractable because they cannot be computed by polynomial-time algorithms.
A relaxation is to allow unreasonable (e.g., exponential) resource demands to be confined to problem parameters that can be kept small in practice.
Parameterizing a given ANN and requested circuit leads to parameterized problems (see \Cref{tbl:params} for problem parameters we study later).
Parameterized queries in the class FPT admit fixed-parameter tractable algorithms.
W-hard queries (by analogy: to FPT as NP-hard is to PTIME), however, do not.
We study counting problems via analogous classes \#P and \#W[1].
We also investigate completeness for NP and classes higher up the polynomial hierarchy such as $\Sigma^p_2$ and $\Pi^p_2$ to identify aspects of hard problems that make them even harder, and to explore the possibility to tackle hard interpretability problems with better-understood methods for well-known NP-complete problems \citep{dehaanParameterizedComplexityClasses2017}. 
Most proofs involve reductions between computational problems which establish the complexity status of interpretability queries based on the known complexity of canonical problems in other areas.

\vspace{-15pt}

    \begin{table}[h]
        \caption{Model and circuit parameterizations.}
        \centering
        \small{
        \begin{tabular}{p{0.3\linewidth} | P{0.2\linewidth} P{0.2\linewidth}}
            \toprule
                {\textbf{Parameter}} & \textit{Model (given)} & \textit{Circuit (requested)}\\
            \midrule
                \CCG Number of layers (depth) & \CCG $\hat{L}$ & $\hat{l}$\\
                \CCG Maximum layer width & \CCG $\hat{L}_w$ & $\hat{l}_w$\\
                \CCB Total number of units\footnotemark & \CCB $\hat{U} = |\mc{M}| \leq \hat{L} \cdot \hat{L}_w$ & $|\mc{C}|=\hat{u}$\\
                Number of input units & $\hat{U}_I$ & $\hat{u}_I$\\
                Number of output units & $\hat{U}_O$ & $\hat{u}_O$\\
                Maximum weight & $\hat{W}$ & $\hat{w}$\\
                Maximum bias & $\hat{B}$ & $\hat{b}$\\
             \bottomrule
        \end{tabular}
        }
        \namedlabel{tbl:params}{parameter table}
    \end{table}

\vspace{-5pt}

\label{sec:approx}
\textbf{Approximation.} Although sometimes computing optimal solutions is intractable, it is conceivable we could devise tractable interpretability procedures to obtain \textit{approximate} solutions that are useful in practice. 
We consider 5 notions of approximation: additive, multiplicative, and three probabilistic schemes ($\mc{A} = \{c, \text{PTAS}, \text{3PA}\}$; see \ref{app} for formal definitions).
\textit{Additive approximation} algorithms return solutions at most a fixed distance $c$ away from optimal (e.g., from the minimum-sized circuit), ensuring that errors cannot get impractically large ($c$-approximability).
\textit{Multiplicative approximation} returns solutions at most a factor of optimal away.
Some hard problems allow for polynomial-time multiplicative approximation schemes (PTAS) where we can get arbitrarily close to optimal solutions as long as we expend increasing compute time \citep{ausielloComplexityApproximation1999}.
Finally, we consider three types of \textit{probabilistic polynomial-time approximability} (henceforth 3PA) that may be acceptable in situations where always getting the correct output for an input is not required: algorithms that (1) always run in polynomial time and produce the correct output for a given input in all but a small number of cases \citep{hemaspaandraSIGACTNewsComplexity2012}; (2) always run in polynomial time and produce the correct output for a given input with high probability \citep{motwaniRandomizedAlgorithms1995}; and (3) run in polynomial time with high probability but are always correct \citep{gillComputationalComplexityProbabilistic1977}.

\textbf{Model architecture.} The Multi-Layer Perceptron (MLP) is a natural first step in our exploration because (a) it is proving useful as a stepping stone in current experimental \citep[e.g.,][]{lampinenLearnedFeatureRepresentations2024} and theoretical work \cite[e.g.,][]{rossemWhenRepresentationsAlign2024, mcinerneyFeedforwardNeuralNetworks2023}; (b) it exists as a leading standalone architecture \citep{yuKANMLPFairer2024}, as the central element of all-MLP architectures \citep{tolstikhinMLPMixerAllMLPArchitecture2021}, and as a key component of state-of-the-art models such as transformers \citep{vaswaniAttentionAllYou2017}; (c) it is of active interest to the interpretability community \citep[e.g.,][]{gevaTransformerFeedForwardLayers2022a, gevaTransformerFeedForwardLayers2021a, daiKnowledgeNeuronsPretrained2022, mengLocatingEditingFactual2024, mengMassEditingMemoryTransformer2022,niuWhatDoesKnowledge2023,vilasAnalyzingVisionTransformers2024,hannaHowDoesGPT22023}; and (d) we can relate our findings in \textit{inner interpretability} to those in \textit{explainability}, which also begins with MLPs \cite[e.g.,][]{barcelo_model_2020,bassanLocalVsGlobal2024a}.
Although MLP blocks can be taken as units to simplify search, it is recommended to investigate MLPs by treating each neuron as a unit \citep[e.g.,][]{gurneeFindingNeuronsHaystack2023, cammarataThreadCircuits2020, olahFeatureVisualization2017}, as it better reflects the semantics of computations in ANNs \citep[][sec. 2.3.1]{lieberumDoesCircuitAnalysis2023}. We adopt this perspective in our analyses.
We write $\mc{M}$ for an MLP model and $\mc{M}(\mb{x})$ for its output on input vector $\mb{x}$.
Its size $|\mc{M}|$ is the number of neurons.
A circuit $\mc{C}$ is a subset of $|\mc{C}|$ neurons which induce a (possibly end-to-end) subgraph of $\mc{M}$ (see \ref{app} for formal definitions).


\vspace{-10pt}
\section{Results \& Discussion: the complexity of circuit queries}
\label{sec:results}
\vspace{-5pt}

In this section we present each circuit query with its computational problem and a discussion of the complexity profile we obtain across variants, relaxations, and parameterizations. For an overview of the results for all queries, see \Cref{tbl:results}.
Proofs of the theorems can be found in the \ref{app}.

\vspace{-8pt}
\subsection{Sufficient Circuit}
\label{sec:res-sc}
\vspace{-5pt}

Sufficient circuits (SCs) are sets of neurons connected end-to-end that suffice, in isolation, to reproduce some model behavior over an input domain \citep[see \textit{faithfulness};][]{wangInterpretabilityWildCircuit2022,yuFunctionalFaithfulnessWild2024}.
They are conceptually related to the desired outcome of zero-ablating components that do not contribute to the behavior of interest (small, parameter-efficient subnetworks).
Zero-ablation as a method (e.g., to find sufficient circuits) has been criticized on the grounds that the patched value (zero) is somewhat arbitrary and therefore can mischaracterize the functioning of the neuron/circuit when operating in the context of the rest of the network during inference.
This gives rise to alternative methods such as activation patching with activation means or specific input activations, which we study later.
SCs remain relevant as, despite valid criticisms of zero-ablation \citep[e.g.,][]{conmyAutomatedCircuitDiscovery2023a}, circuit discovery through pruning might be justified at least in some cases \citep[][]{yuFunctionalFaithfulnessWild2024}.

\begin{compprob}{Bounded Local Sufficient Circuit}{BLSC}
\namedlabel{prb:BLSC}{\textsc{BLSC}}
A multi-layer perceptron $\mc{M}$, an input vector $\mb{x}$, and an integer $u \leq |\mc{M}|$.
\nextpart
A circuit $\mc{C}$ in $\mc{M}$ of size $|\mc{C}| \leq u$, such that $\mc{C}(\mb{x}) = \mc{M}(\mb{x})$, if it exists, otherwise $\bot$. 
\end{compprob}

We find that many variants of SC are NP-hard (see \Cref{tbl:results}).
Counterintuitively, this intractability does not depend straightforwardly on parameters such as network depth (W[1]-hard relative to $\mc{P}$).
Therefore, hardness is not mitigated by keeping models shallow.
Given this barrier, we explore the possibility of obtaining approximate solutions but find that hard SC variants are inapproximable relative to all schemes in \Cref{sec:approx}.
An alternative is to consider the membership of these problems in a well-studied class whose solvers are better understood than interpretability heuristics \citep{dehaanParameterizedComplexityClasses2017}.
We prove that local versions of SC are NP-complete. 
This implies there exist efficient transformations from instances of SC to those of the satisfiability problem \citep[SAT;][]{biereHandbookSatisfiability2021}, opening up the possibility to borrow techniques that work reasonably well in practice for SAT that might be suitable for some versions of neural network problems.
Interestingly, this is not possible for the global version, which we prove is complete for a class higher up the complexity hierarchy ($\Sigma^p_2$-complete). 
This result establishes a formal separation between local and global query complexity that partly explains `interpretability illusions' \citep[][]{friedmanInterpretabilityIllusionsGeneralization2024,yuFunctionalFaithfulnessWild2024}, which we conjecture holds for other queries we investigate later. 
These illusions come about when an interpretability abstraction (e.g., a circuit) seems empirically faithful to its target (e.g., model behavior) by some criterion (e.g., `local' tests on a dataset), but actually lacks faithfulness in the way it generalizes to other criteria (e.g., tests of its `global' behavior outside the original distribution).

Next we explore whether we could diagnose if SCs with some desired property (e.g., minimality) are abundant, which would be informative of the ability of heuristic search to stumble upon one of them.
We analyze various queries where the output is a count of SCs (i.e., counting problems).
We find that both local and global, bounded and unbounded variants are \#P-complete and remain intractable (\#W[1]-hard) when parameterized by many network features including depth (\Cref{tbl:params}).

The hardness profile of SC over all these variants calls for exploring more substantial relaxations.
We introduce the notion of \textit{quasi-minimality} for this purpose (similar to \citealp{ramaswamyAlgorithmicBarrierNeural2019}) and later demonstrate its usefulness beyond this particular problem.
Any neuron in a \textit{minimal/minimum} SC is a breaking point in the sense that removing it will break the target behavior.
In \textit{quasi-minimal} SCs we are merely guaranteed to know at least one neuron that causes this breakdown.
By introducing this relaxation, which gives up some affordances but retains others of interest, we get a feasible interpretability query.

\begin{compprob}{Unbounded Quasi-minimal Local Sufficient Circuit}{UQLSC}
\namedlabel{prb:UQLSC}{\textsc{UQLSC}}
A multi-layer perceptron $\mc{M}$, and an input vector $\mb{x}$.
\nextpart
A circuit $\mc{C}$ in $\mc{M}$ and a neuron $v \in \mc{C}$ s.t. $\mc{C}(\mb{x})=\mc{M}(\mb{x})$ and $[ \mc{C}\setminus\{v\} ](\mb{x})\neq\mc{M}(\mb{x})$.
\end{compprob}

\ref{prb:UQLSC} is in PTIME.
We describe an efficient algorithm to compute it which can be heuristically biased towards finding smaller circuits and combined with techniques that exploit weights and gradients (see \ref{app}).

\vspace{-10pt}
\subsection{Gnostic Neuron}
\vspace{-5pt}

Gnostic neurons, sometimes called `grandmother neurons' in neuroscience \citep{galeAreThereAny2020} and `concept neurons' or `object detectors' in AI \citep[e.g.,][]{bauUnderstandingRoleIndividual2020}, are one of the oldest and still current interpretability queries of interest \citep[see also `knowledge neurons';][]{niuWhatDoesKnowledge2023}.

\begin{compprob}{Bounded Gnostic Neurons}{BGN}
\namedlabel{prb:BGN}{\textsc{BGN}}
A multi-layer perceptron $\mc{M}$ and two sets of input vectors $\mc{X}$ and $\mc{Y}$, an integer $k$, and an activation threshold $t$.
\nextpart
A set of neurons $V$ in $\mc{M}$ of size $|V| \geq k$ such that $\forall_{v \in V}$ it is the case that $\forall_{\mb{x} \in \mc{X}}$, $\mc{M}(\mb{x})$ produces activations $A^v_{\mb{x}} \geq t$ and $\forall_{\mb{y} \in \mc{Y}} : A^v_{\mb{y}} < t $, if it exists, else $\bot$.
\end{compprob}


\ref{prb:BGN} is in PTIME.
Alternatives might require GNs to have some behavioral effect when intervened; such variants would remain tractable.

\vspace{-10pt}
\subsection{Circuit Ablation and Clamping}
\vspace{-5pt}

The idea that some neurons perform key subcomputations for certain tasks naturally leads to the hypothesis that ablating them should have downstream effects on the corresponding model behaviors.
Searching for neuron sets with this property has been one strategy (i.e., \textit{zero-ablation}) to get at important circuits \citep{wangDoesEditingProvide2024}.
The circuit ablation (CA) problem formalizes this idea.

\begin{compprob}{Bounded Local Circuit Ablation}{BLCA}
\namedlabel{prb:BLCA}{\textsc{BLCA}}
A multi-layer perceptron $\mc{M}$, an input vector $\mb{x}$, and an integer $u \leq |\mc{M}|$.
\nextpart
A neuron set $\mc{C}$ in $\mc{M}$ of size $|\mc{C}| \leq u$, s.t. $[\mc{M} \setminus \mc{C}](\mb{x}) \neq \mc{M}(\mb{x})$, if it exists, else $\bot$. 
\end{compprob}


A difference between CAs and minimal SCs is that the former can be interpreted as a possibly non-minimal breaking set in the context of the whole network whereas the latter is by default a minimal breaking set when the SC is taken in isolation. In this sense, CA can be seen as a less stringent criterion for circuit affordances. A related idea is circuit clamping (CC): fixing the activations of certain neurons to a level that produces a change in the behavior of interest.

\begin{compprob}{Bounded Local Circuit Clamping}{BLCC}
\namedlabel{prb:BLCC}{\textsc{BLCC}}
A multi-layer perceptron $\mc{M}$, vector $\mb{x}$, value $r$, and an integer $u$ s.t. $1 < u \leq |\mc{M}|$.
\nextpart
A subset of neurons $\mc{C}$ in $\mc{M}$ of size $|\mc{C}| \leq u$, such that for the $\mc{M}^*$ induced by clamping all $c \in \mc{C}$ to value $r$, $\mc{M}^*(\mb{x}) \neq \mc{M}(\mb{x})$, if it exists, otherwise $\bot$. 
\end{compprob}


Despite these more modest criteria, we find that both the local and global variants of CA and CC are NP-hard, fixed-parameter intractable W[1]-hard relative to various parameters, and inapproximable in all 5 senses studied.
However, we prove these problems are NP-complete, which opens up practical options not available for other problems we study (see remarks in \Cref{sec:res-sc}).

\vspace{-8pt}

\subsection{Circuit Patching}

\vspace{-5pt}

A critique of zero-ablation is the arbitrariness of the value, leading to alternatives such as mean-ablation \citep[e.g.,][]{wangInterpretabilityWildCircuit2022}.
This contrasts studying circuits in isolation versus embedded in surrounding subnetworks.
Activation patching \citep{ghandehariounPatchscopesUnifyingFramework2024,zhangBestPracticesActivation2023,hannaHaveFaithFaithfulness2024} and path patching \citep{goldowsky-dillLocalizingModelBehavior2023} try to pinpoint which activations play an in-context role in model behavior, which inspires the circuit patching (CP) problem.


\begin{compprob}{Bounded Local Circuit Patching}{BLCP}
\namedlabel{prb:BLCP}{\textsc{BLCP}}
A multi-layer perceptron $\mc{M}$, an integer $k$, an input vector $\mb{y}$, and a vector set $\mc{X}$.
\nextpart
A subset $\mc{C}$ in $\mc{M}$ of size $|\mc{C}| \leq k$, such that for the $\mc{M}^*$ induced by patching $\mc{C}$ with activations from $\mc{M}(\mb{y})$ and $\mc{M} \setminus \mc{C}$ with activations from $\mc{M}(\mb{x})$, $\mc{M}^*(\mb{x}) = \mc{M}(\mb{y})$ for all $\mb{x} \in \mc{X}$, if it exists, otherwise $\bot$.
\end{compprob}

We find that local/global variants are intractable (NP-hard) in a way that does not depend on parameters such as network depth or size of the patched circuit (W[1]-hard), and are inapproximable ($\{c, \text{PTAS}, \text{3PA}\}$-inapprox.).
Although we also prove the local variant of CP is NP-complete and therefore approachable in practice with solvers for hard problems not available for the global variants (see remarks in \Cref{sec:res-sc}), these complexity barriers motivate exploring further relaxations.
With some modifications the idea of \textit{quasi-minimality} can be repurposed to do useful work here.

\begin{compprob}{Unbounded Quasi-minimal Local Circuit Patching}{UQLCP}
\namedlabel{prb:UQLCP}{\textsc{UQLCP}}
A multi-layer perceptron $\mc{M}$, an input vector $\mb{y}$, and a set $\mc{X}$ of input vectors.
\nextpart
A subset $\mc{C}$ in $\mc{M}$ and a neuron $v \in \mc{C}$, such that for the $\mc{M}^*$ induced by patching $\mc{C}$ with activations from $\mc{M}(\mb{y})$ and $\mc{M} \setminus \mc{C}$ with activations from $\mc{M}(\mb{x})$, \\ $\forall_{\mb{x} \in \mc{X}}: \mc{M}^*(\mb{x}) = \mc{M}(\mb{y})$, and for $\mc{M}'$ induced by patching identically \\ except for $v \in \mc{C}$, $\exists_{\mb{x} \in \mc{X}}: \mc{M}'(\mb{x}) \neq \mc{M}(\mb{y})$.
\end{compprob}

In this way we obtain a tractable query (PTIME) for quasi-minimal patching, sidestepping barriers while retaining some useful affordances (see \Cref{tbl:afford}).
We present an algorithm to compute \ref{prb:UQLCP} efficiently that can be combined with strategies exploiting weights and gradients (see \ref{app}).

\vspace{-10pt}
\newcommand\fbar{\kern1pt\rule[-\dp\strutbox]{1.pt}{\baselineskip}\kern1pt}

\begin{table}[H]
    \caption{Classical and parameterized complexity results by problem variant.}
    \centering
    \small{
    \begin{tabular}{p{0.34\linewidth} | P{0.15\linewidth} P{0.15\linewidth} | P{0.17\linewidth} P{0.15\linewidth}}
        \toprule \toprule
             \multirow{3}{180pt}{\textbf{Classical \& parameterized queries\footnotemark} 
             \newline\hfill $\mc{P} = \mc{P}_\mc{M} \cup \mc{P}_\mc{C}$
             \newline\hfill $\mc{P}_\mc{M} = \{ \hat{L}, \hat{U}_I, \hat{U}_O, \hat{W}, \hat{B} \}$
             \newline\hfill $\mc{P}_\mc{C} = \{ \hat{l}, \hat{l}_w, \hat{u}, \hat{u}_I,\hat{u}_O,\hat{w}, \hat{b} \}$} & \multicolumn{4}{c}{\textbf{Problem variants}} \\
        \cmidrule{2-5} 
              & \multicolumn{2}{c|}{\textit{Local}} & \multicolumn{2}{c}{\textit{Global}} \\
              \cmidrule(lr){2-3}  \cmidrule(lr){4-5}
               & \textit{Decision/Search} & \textit{Optimization} & \textit{Decision/Search} & \textit{Optimization} \\
        \toprule 
            \textsc{Sufficient Circuit (SC)} &  \CCT NP-complete & \CCP $\mc{A}$-inapprox. & \CCY $\Sigma^p_2$-complete & \CCP $\mc{A}$-inapprox. \\
             \hfill $\mc{P}$-SC   &  \CCR  W[1]-hard  & \CCP  $\mc{A}$-inapprox. & \CCR  W[1]-hard & \CCP $\mc{A}$-inapprox. \\
             \hfill Minimal SC & \CCT NP-complete & \CCG \textbf{?} & \CCO $\in \Sigma^p_2$ \fbar \ NP-hard & \CCG \textbf{?} \\
             \hfill $\mc{P}$-Minimal SC & \CCR  W[1]-hard & \CCG \textbf{?} & \CCR  W[1]-hard & \CCG \textbf{?}  \\
             \hfill Unbounded Minimal SC & \CCG \textbf{?} & \multirow{3}{20pt}{\slashbox{N}{A}} & \CCO \CCG \textbf{?} & \multirow{3}{20pt}{\slashbox{N}{A}} \\
             \hfill $\mc{P}$-Unbounded Minimal SC & \CCG \textbf{?} &  & \CCG \textbf{?} \\
             \hfill Unbounded Quasi-Minimal SC & \CCM PTIME &  & \CCG \textbf{?} &  \\
         \arrayrulecolor{lightgray} \midrule \arrayrulecolor{black}
            \hfill Count SC &  \CCF \#P-complete  & \multirow{5}{20pt}{\slashbox{N}{A}} & \CCFFF \#P-hard &  \multirow{5}{20pt}{\slashbox{N}{A}} \\
            \hfill $\mc{P}$-Count SC &  \CCFF \#W[1]-hard &  & \CCFF \#W[1]-hard & \\
            \hfill Count Minimal SC &  \CCF \#P-complete &  & \CCFFF \#P-hard &  \\
            \hfill $\mc{P}$-Count Minimal SC & \CCFF \#W[1]-hard &  & \CCFF \#W[1]-hard &  \\
            \hfill Count Unbounded Minimal SC &  \CCF \#P-complete &  & \CCFFF  \#P-hard &  \\
         \midrule
            \textsc{Gnostic Neuron} (GN) & \CCM PTIME & N/A & \CCG \textbf{?} & N/A \\
         \midrule
            \textsc{Circuit Ablation} (CA) &  \CCT NP-complete & \CCP $\mc{A}$-inapprox. & \CCO $\in \Sigma^p_2$ \fbar \ NP-hard & \CCP $\mc{A}$-inapprox. \\
            \hfill $\{ \hat{L}, \hat{U}_I, \hat{U}_O, \hat{W}, \hat{B}, \hat{u} \}$-CA &  \CCR W[1]-hard & \CCP $\mc{A}$-inapprox.  & \CCR W[1]-hard & \CCP  $\mc{A}$-inapprox. \\
        \midrule    
            \textsc{Circuit Clamping} (CC) & \CCT NP-complete & \CCP $\mc{A}$-inapprox. & \CCO $\in \Sigma^p_2$ \fbar \ NP-hard & \CCP  $\mc{A}$-inapprox. \\
            \hfill $\{ \hat{L}, \hat{U}_O, \hat{W}, \hat{B}, \hat{u} \}$-CC & \CCR W[1]-hard & \CCP  $\mc{A}$-inapprox. & \CCR W[1]-hard & \CCP $\mc{A}$-inapprox. \\
        \midrule
            \textsc{Circuit Patching} (CP) & \CCT NP-complete & \CCP $\mc{A}$-inapprox. & \CCO $\in \Sigma^p_2$ \fbar \ NP-hard & \CCP $\mc{A}$-inapprox. \\
            \hfill $\{ \hat{L}, \hat{U}_O, \hat{W}, \hat{B}, \hat{u} \}$-CP & \CCR W[2]-hard & \CCP $\mc{A}$-inapprox. & \CCR W[2]-hard & \CCP $\mc{A}$-inapprox. \\
            \hfill Unbounded Quasi-Minimal CP & \CCM PTIME &  N/A & \CCG \textbf{?} &  N/A \\
        \midrule
             \textsc{Necessary Circuit} (NC) & \CCO $\in \Sigma^p_2$ \fbar NP-hard & \CCP $\mc{A}$-inapprox. & \CCO $\in \Sigma^p_2$ \fbar \ NP-hard & \CCP $\mc{A}$-inapprox. \\
            \hfill $\{ \hat{L}, \hat{U}_I, \hat{U}_O, \hat{W}, \hat{u} \}$-NC & \CCR W[1]-hard & \CCP $\mc{A}$-inapprox. & \CCR W[1]-hard & \CCP $\mc{A}$-inapprox. \\ 
        \midrule
            \textsc{Circuit Robustness} (CR) & \CCT coNP-complete & \CCG \textbf{?} & \CCO $\in \Pi^p_2$ \fbar \ coNP-hard  & \CCG \textbf{?} \\
            \hfill $\{ \hat{L}, \hat{U}_I, \hat{U}_O, \hat{W}, \hat{B}, \hat{u} \}$-CR & \CCR coW[1]-hard & \CCG \textbf{?} &  \CCR coW[1]-hard & \CCG \textbf{?} \\
            \hfill \{$|H|$\}-CR & \CCM FPT & \CCM FPT &  \CCG \textbf{?} & \CCG \textbf{?} \\
            \hfill \{$|H|$, $\hat{U}_I$\}-CR & \CCM FPT & \CCM FPT & \CCM FPT & \CCM FPT \\
        \bottomrule
            \textsc{Sufficient Reasons} (SR) & \CCO $\in \Sigma^p_2$ \fbar \ NP-hard & \CCP 3PA-inapprox. & \multicolumn{2}{c}{\multirow{2}{20pt}{\slashbox{N}{A}}} \\
            \hfill $\{ \hat{L}, \hat{U}_O, \hat{W}, \hat{B}, \hat{u} \}$-SR & \CCR W[1]-hard & \CCP 3PA-inapprox. \\
         \bottomrule \bottomrule
    \end{tabular}
    }
    \namedlabel{tbl:results}{results table}
\end{table}

\footnotetext{Circuits are bounded-size unless otherwise stated. Each cell contains the complexity of the problem variant in terms of classical and FP (in)tractability, membership in complexity classes, and (in)approximability ($\mc{A} = \{c, \text{PTAS}, \text{3PA} \}$). `\textbf{?}' marks potentially fruitful open problems. `N/A' stands for not applicable.}


\subsection{Necessary Circuit}
\vspace{-5pt}

The criterion of \textit{necessity} is a stringent one, and consequently necessary circuits (NCs) carry powerful affordances (see \Cref{tbl:afford}). Since neurons in NCs collectively interact with all possible sufficient circuits for a target behavior, they are candidates to describe key task subcomputations and intervening on them is guaranteed to have effects even in the presence of high redundance.
This relates to the notion of \textit{circuit overlap} and therefore to efforts in identifying circuits shared by various tasks \citep[e.g.,][]{merulloCircuitComponentReuse2024}, and the link between overlap and faithfulness \citep[e.g.,][]{hannaHaveFaithFaithfulness2024}.

\begin{compprob}{Bounded Global Necessary Circuit}{BGNC}
\namedlabel{prb:BGNC}{\textsc{BGNC}}
A multi-layer perceptron $\mc{M}$, and an integer $k$.
\nextpart
A subset $\mc{S}$ of neurons in $\mc{M}$ of size $|\mc{S}| \leq k$, such that $\mc{S} \cap \mc{C} \neq \emptyset$ for every circuit $\mc{C}$ in $\mc{M}$ that is sufficient relative to all possible input vectors, if it exists, otherwise $\bot$.
\end{compprob}

Unfortunately both local and global versions of NC are NP-hard (in $\Sigma^p_2$; \Cref{tbl:results}), remain intractable even when keeping parameters such as network depth, number of input and output neurons, and others small (\Cref{tbl:params}), and does not admit any of the available approximation schemes (\Cref{sec:approx}).
Tractable versions of NC are unlikely unless substantial restrictions or relaxations are introduced.

\vspace{-8pt}
\subsection{Circuit Robustness}
\vspace{-5pt}

A behavior of interest might be over-determined or resilient in the sense that many circuits in the model implement it and one can take over when the other breaks down.
This is related to the notion of \textit{redundancy} used in neuroscience \citep[e.g.,][]{NandaRedundancy2023}.
Intuitively, when a model implements a task in this way, the behavior should be more robust to a number of perturbations.
The possibility of verifying this property experimentally motivates the circuit robustness (CR) problem, and a related interpretability effort is diagnosing nodes that are excluded by circuit discovery procedures but still have an impact on behavior \cite[false negatives;][]{kramarAtPEfficientScalable2024}.

\begin{compprob}{Bounded Local Circuit Robustness}{BLCR}
\namedlabel{prb:BLCR}{\textsc{BLCR}}
A multi-layer perceptron $\mc{M}$, a subset $H$ of $\mc{M}$, an input vector $\mb{x}$, and an integer $k$ with $1 \leq k \leq |H|$.
\nextpart
\texttt{<YES>} if for each $H' \subseteq H$, with $|H'| \leq k$, 
$\mc{M}(\mb{x}) = [\mc{M} \setminus H'](\mb{x})$, else \texttt{<NO>}.
\end{compprob}

We find that Local CR is coNP-complete while Global CR is in $\Pi^p_2$ and coNP-hard.
It remains fixed-parameter intractable (coW[1]-hard) relative to model parameters (\Cref{tbl:params}).
Pushing further, we explore parameterizing CR by $\{|H|\}$ and prove fixed-parameter tractability of $\{|H|\}$-CR which holds both for the local and global versions.
There exist algorithms for CR that scale well as long as $|H|$ is reasonable; a scenario that might be useful to probe robustness in practice. This wraps up our results for circuit queries.
We briefly digress into \textit{explainability} before discussing some implications.

\vspace{-8pt}
\subsection{Sufficient Reasons}
\vspace{-5pt}

Understanding the sufficient reasons (SR) for a model decision in terms of input features consists of knowledge of values of the input components that are enough to determine the output. Given a model decision on an input, the most interesting reasons are those with the least components.

\begin{compprob}{Bounded Local Sufficient Reasons}{BLSR}
\namedlabel{prb:BLSR}{\textsc{BLSR}}
A multi-layer perceptron $\mc{M}$, an input vector $\mb{x}$ of length $|\mb{x}| = \hat{u}_I$, and an integer $k$ with $1 \leq k \leq \hat{u}_I$.
\nextpart
A subset $\mb{x}^s$ of $\mb{x}$ of size $|\mb{x}^s| = k$, such that for every possible completion $\mb{x}^c$ of $\mb{x}^s$ $\mc{M}(\mb{x^c}) = \mc{M}(\mb{x})$, if it exists, otherwise $\bot$. 
\end{compprob}

To demonstrate the usefulness of our framework beyond inner interpretability, we show how it links to explainability.
Using our techniques for circuit queries, we significantly tighten existing results for SR \citep{barcelo_model_2020,WM+21} by proving that hardness (NP-hard, W[1]-hard, 3PA-inapprox.) holds even when the model has only one hidden layer.

\vspace{-8pt}
\section{Implications, limitations, and future directions}
\vspace{-5pt}

We presented a framework based on parameterized complexity to accompany experiments on inner interpretability with theoretical explorations of viable algorithms.
With this grasp of circuit query complexity, we can understand the challenges of scalability and the mixed outcomes of experiments with heuristics for circuit discovery.
There is ample complexity-theoretic evidence that there is a limit (often underestimated) to how good the performance of heuristics on intractable problems can be \citep{hemaspaandraSIGACTNewsComplexity2012}.
We can explain `interpretability illusions' \citep{friedmanInterpretabilityIllusionsGeneralization2024} due to lack of faithfulness, minimality \citep[e.g.,][]{shiHypothesisTestingCircuit2024,yuFunctionalFaithfulnessWild2024} and other affordances \citep{wangDoesEditingProvide2024, haseDoesLocalizationInform2023}, in terms of the kinds of circuits that our current heuristics are well-equipped to discover.
For instance, consider the algorithm for automated circuit discovery proposed by \citet{conmyAutomatedCircuitDiscovery2023a}, which eliminates one network component at a time if the consequence on behavior is reasonably small.
Since this algorithm runs in polynomial time, it is not likely to solve the problems proven hard here, such as \textsc{Minimal Sufficient Circuit}.
However, one reason we observe interesting results in some cases is because it is well-equipped to solve \textsc{Quasi-Minimal Circuit} problems.
As our conceptual and formal analyses show, quasi-minimal circuits can mimic various desirable aspects of sufficient circuits (\Cref{tbl:afford}), and the former can be found tractably (results for \Cref{prb:UQLSC} and \Cref{prb:UQLCP}).
At the same time, understanding these properties of circuit discovery heuristics helps us explain observed discrepancies: why we often see (1) lack of faithfulness (i.e., global coverage is out of reach for QMC algorithms), (2) non-minimality (i.e., QM circuits can have many non-breaking points), and (3) large variability in performance across tasks and analysis parameters \citep[e.g.,][]{shiHypothesisTestingCircuit2024,conmyAutomatedCircuitDiscovery2023a}.

Although we find that many queries of interest are intractable in the general case (and empirical results are in line with this characterization), this should not paralyze efforts to interpret neural network models.
As our exploration of the current complexity landscape shows, reasonable relaxations, restrictions and problem variants can yield tractable queries for circuits with useful properties.
Consider a few out of many possible avenues to continue these explorations.

\textbf{(i) Study query parameters.} Faced with an intractable query, we can investigate which parameters of the problem (e.g., network or circuit aspects) might be responsible for its core hardness.
If these problematic parameters can be kept small in real-world applications, this yields a fixed-parameter tractable query.
We have explored some, but more are possible as any aspect of the problem can be parameterized.
A close dialogue between theorists and experimentalists is important for this, as empirical regularities suggest which parameters might be fruitful to explore theoretically, and experiments test whether theoretically conjectured parameters can be kept small in practice.

\textbf{(ii) Generate novel queries.} 
Our formalization of quasi-minimal circuit problems illustrates the search for viable algorithmic options with examples of tractable problems for inner interpretability.
When the use case is well defined, efficient queries that return circuits with useful affordances for applications can be designed. 
Alternative circuits might also mimic the affordances for prediction/control of ideal circuits while avoiding intractability.

\textbf{(iii) Explore network output as axis of approximation.}
Some of our constructions use binary input/output \citep[following previous work; e.g.,][]{bassanLocalVsGlobal2024a,barcelo_model_2020}.
Although continuous output does not necessarily matter complexity-wise (see \ref{app} for [counter]examples), this is an interesting direction for future work, as it opens the door to studying the network output as an axis of approximation, which in turn might be a useful relaxation.

\textbf{(iv) Design more abstract queries.} A different path is to design queries that partially rely on mid-level abstractions \citep{vilasPositionInnerInterpretability2024} to bridge the gap between circuits and human-intelligible algorithms \citep[e.g., key-value mechanisms;][]{gevaTransformerFeedForwardLayers2022a, vilasAnalyzingVisionTransformers2024}.

\textbf{(v) Characterize actual network structure.} It is in principle possible that some real-world, trained neural networks possess internal structure that is benevolent to general (ideal) circuit queries (e.g., redundancy; see \ref{app}).
In such optimistic scenarios, general-purpose heuristics might work well.
The empirical evidence available to date, however, speaks against this.
In any case, it will always be important to characterize any such structure to use it explicitly to design algorithms with useful guarantees.

\textbf{(vi) Compare resource demands of interpretability/explainability across architectures.} Our results for inner interpretability complement those of explainability \citep[e.g.,][]{barcelo_model_2020,bassanLocalVsGlobal2024a,WM+21}.
These aspects can be studied together for different architectures to assess their intrinsic interpretability.
To some extent our results already transfer to some cases of interest. Since the transformer architecture contains MLPs, the complexity status of our circuit queries bears on neuron-level circuit discovery efforts in transformers (e.g., Large Language/Vision/Audio models).


\newpage

\section*{Acknowledgments}
We thank Ronald de Haan for comments on proving membership using alternating quantifier formulas.


\bibliography{iclr2025_conference}
\bibliographystyle{iclr2025_conference}

\newpage


\appendix




\part{Appendix: Definitions, Theorems and Proofs} 
\parttoc 
\namedlabel{app}{Appendix}


\section{Preliminaries}
Each section of this appendix is self-contained except for the following definitions.
We re-state interpretability query definitions in a more detailed form in each section for convenience.
To err here on the side of rigor, here we will use more cumbersome notation that we avoided in the main manuscript for succinctness.

\subsection{Computational problems of known complexity}

Some of our proofs construct reductions from the following computational problems.

\noindent
{\sc Clique} \citep[Problem GT19]{garey1979computers} \\
{\em Input}: An undirected graph $G = (V, E)$ and a positive integer $k$. \\
{\em Question}: Does $G$ have a clique of size at least $k$, i.e., 
                 a subset $V' \subseteq V$, $|V'| \geq k$, such that for 
                 all pairs $v, v' \in V'$, $(v, v') \in E$?

\vspace*{0.1in}

\noindent
{\sc Vertex cover} (VC) \citep[Problem GT1]{garey1979computers} \\
{\em Input}: An undirected graph $G = (V, E)$ and a positive integer $k$. \\
{\em Question}: Does $G$ contain a vertex cover of size at most $k$, i.e., 
                 a subset $V' \subseteq V$, $|V'| \leq k$, such that for 
                 all $(u, v) \in E$, at least one of $u$ or $v$ is in $V'$? 

\vspace*{0.1in}

\noindent
{\sc Dominating set} (DS) \citep[Problem GT2]{garey1979computers} \\
{\em Input}: An undirected graph $G = (V, E)$ and a positive integer $k$. \\
{\em Question}: Does $G$ contain a dominating set of size at most $k$, i.e.,
                 a subset $V' \subseteq V$, $|V'| \leq k$, such that for all
                 $v \in V$, either $v \in V'$ or there is at least one
                 $v' \in V'$ such that $(v, v') \in E$?

\vspace*{0.1in}

\noindent
{\sc Hitting set} (HS) \citep[Problem SP8]{garey1979computers} \\
{\em Input}: A collection of subsets $C$ of a finite set $S$ and a
              positive integer $k$. \\
{\em Question}: Is there a subset $S'$ of $S$, $|S'| \leq k$, such that
              $S'$ has a non-empty intersection with each set in $C$?

\vspace*{0.1in}

\noindent
{\sc Minimum DNF Tautology} (3DT) \cite[Problem L7]{SU02} \\
{\em Input}: A 3-DNF tautology $\phi$ with $T$ terms over a set of variables
              $V$ and a positive integer $k$. \\
{\em Question}: Is there a 3-DNF formula $\phi'$ made up of $\leq k$ of the
             terms in $\phi$ that is a also a tautology?

\vspace*{0.1in}



\subsection{Classical and parameterized complexity}

\begin{definition}[Polynomial-time tractability]\label{def:polytime}
An algorithm is said to run in \textit{polynomial-time} if the number of steps it performs is $O(n^c)$, where $n$ is a measure of the input size and $c$ is some constant. A problem $\Pi$ is said to be \textit{tractable} if it has a \textit{polynomial-time algorithm}. P denotes the class of such problems.
\end{definition}

Consider a more fine-grained look at the \textit{sources of complexity} of problems.
The following is a relaxation of the notion of tractability, where unreasonable resource demands are allowed as long as they are constrained to a set of problem parameters.

\begin{definition}[Fixed-parameter tractability]\label{def:fpt-tractability}
    Let $\mc{P}$ be a set of problem parameters. A problem $\mc{P}$-$\Pi$ is \textit{fixed-parameter tractable} relative to $\mc{P}$ if there exists an algorithm that computes solutions to instances of $\mc{P}$-$\Pi$ of any size $n$ in time $f(\mc{P}) \cdot n^c$, where $c$ is a constant and $f(\cdot)$ some computable function. FPT denotes the class of such problems and includes all problems in P. 
\end{definition}


\subsection{Hardness and reductions} 

Most proof techniques in this work involve reductions between computational problems.

\begin{definition} [Reducibility]\label{def:reducibility}
A problem $\Pi_1$ is \textit{polynomial-time reducible} to $\Pi_2$ if there exists a polynomial-time algorithm (\textit{reduction}) that transforms instances of $\Pi_1$ into instances of $\Pi_2$ such that solutions for $\Pi_2$ can be transformed in polynomial-time into solutions for $\Pi_1$. 
This implies that if a tractable algorithm for $\Pi_2$ exists, it can be used to solve $\Pi_1$ tractably.
\textit{Fpt-reductions} transform an instance $(x,k)$ of some problem parameterized by $k$ into an instance $(x',k')$ of another problem, with $k'\leq g(k)$, in time $f(k)\cdot p(|x|)$ where $p$ is a polynomial and $g(\cdot)$ is an arbitrary function.
These reductions analogously transfer fixed-parameter tractability results between problems.

\end{definition}

\noindent Hardness results are generally conditional on two conjectures with extensive theoretical and empirical support.
Intractability statements build on these as follows. 

\begin{conjecture}
\label{cnj:p-np}
    P $\neq$ NP.
\end{conjecture}

\begin{definition}[Polynomial-time intractability] \label{def:poly-intractability}
The class NP contains all problems in P and more. Assuming \Cref{cnj:p-np}, 
NP-hard problems lie outside P.
These problems are considered \textit{intractable} because they cannot be solved in polynomial-time \citep[unless \Cref{cnj:p-np} is false; see][]{fortnow2009status}.
\end{definition}

\begin{conjecture}
\label{cnj:fpt-w}
    FPT $\neq$ W[1].
\end{conjecture}

\begin{definition}[Fixed-parameter intractability] \label{def:fp-intractability}
    The class W[1] contains all problems in the class FPT and more.
    Assuming \Cref{cnj:fpt-w}, W[1]-hard parameterized problems lie outside FPT.
    These problems are considered \textit{fixed-parameter intractable}, relative to a given parameter set, because no fixed-parameter tractable algorithm can exist to solve them \citep[unless \Cref{cnj:fpt-w} is false; see][]{downeyFundamentalsParameterizedComplexity2013}.
\end{definition}

The following two easily-proven lemmas will be useful in our parameterized complexity proofs.

\begin{lemma}
\citep[Lemma 2.1.30]{warehamSystematicParameterizedComplexity1999}
If problem $\Pi$ is fp-tractable relative to aspect-set $K$ then $\Pi$ is fp-tractable for any aspect-set $K'$ such that $K \subset K'$.
\label{LemPrmFPProp1}
\end{lemma}

\begin{lemma}
\citep[Lemma 2.1.31]{warehamSystematicParameterizedComplexity1999}
If problem $\Pi$ is fp-intractable relative to aspect-set $K$ then $\Pi$ is fp-intractable for any aspect-set $K'$ such that $K' \subset K$.
\label{LemPrmFPProp2}
\end{lemma}


\subsection{Approximation}
\label{app:approx}

Although sometimes computing optimal solutions might be intractable, it is still conceivable that we could devise tractable procedures to obtain \textit{approximate} solutions that are useful in practice. 
We consider two natural notions of additive and multiplicative approximation and three probabilistic schemes.

\subsubsection{Multiplicative approximation} 
For a minimization problem $\Pi$, let $OPT_{\Pi}(I)$ be an optimal solution for $\Pi$ on instance $I$, $A_{\Pi}(I)$ be a solution for $\Pi$ returned by an algorithm $A$, and $m(OPT_{\Pi}(I))$ and $m(A_{\Pi}(I))$ be the values of these solutions.

\begin{definition}[Multiplicative approximation algorithm]
{[\citealt[Def. 3.5]{ausielloComplexityApproximation1999}]}.
Given a minimization problem $\Pi$, an algorithm $A$ is a \textit{multiplicative} $\epsilon$-\textit{approximation algorithm} for $\Pi$ if for each instance $I$ of $\Pi$, $m(A_{\Pi}(I)) - m(OPT_{\Pi}(I)) \leq \epsilon \times m(OPT_{\Pi}(I))$.
\label{def:mult-approx-alg}
\end{definition}

\noindent It would be ideal if one could obtain approximate solutions for a problem
$\Pi$ that are arbitrarily close to optimal if one is willing to allow
extra algorithm runtime.

\begin{definition}[Multiplicative approximation scheme]
{[\citealp[Adapted from][Def. 3.10]{ausielloComplexityApproximation1999}]}.
Given a minimization problem $\Pi$, a \textit{polynomial-time approximation scheme}
(PTAS) for $\Pi$ is a set $\cal{A}$ of algorithms such that for each integer 
$k > 0$, there is a $\frac{1}{k}$-approximation algorithm $A^k_{\Pi} \in \cal{A}$ 
that runs in time polynomial in $|I|$.
\label{def:mult-approx-scheme}
\end{definition}

\subsubsection{Additive approximation}
It would be useful to have guarantees that an approximation algorithm for our problems returns solutions at most a fixed distance away from optimal.
This would ensure errors cannot get impractically large. 

\begin{definition}[Additive approximation algorithm] {[\citealp[Adapted from][Def. 3.3]{ausielloComplexityApproximation1999}]}. An algorithm $A_{\Pi}$ for a problem $\Pi$ is a \textit{d-additive approximation algorithm} ($d$-AAA) if there exists a constant $d$ such that for all instances $x$ of $\Pi$ the error between the value $m(\cdot)$ of an optimal solution $optsol(x)$ and the output $A_{\Pi}(x)$ is such that $| \ m(optsol(x)) - m(A_{\Pi}(x)) \ | \leq d$.
\label{def:add-approx}
\end{definition}

\subsubsection{Probabilistic approximation}
Finally, consider three other types of probabilistic polynomial-time approximability (henceforth 3PA) that may be acceptable in situations where always getting the correct output for an input is not required: (1) algorithms that always run in polynomial time and produce the correct output for a given input in all but a small number of cases \citep{hemaspaandraSIGACTNewsComplexity2012}; (2) algorithms that always run in polynomial time and produce the correct output for a given input with high probability \citep{motwaniRandomizedAlgorithms1995}; and (3) algorithms that run in polynomial time with high probability but are always correct \citep{gillComputationalComplexityProbabilistic1977}.

\subsection{Model architecture}

\begin{definition}[Multi-Layer Perceptron]
\label{def:mlp}
    {[Adapted from \citealt{barcelo_model_2020}]}.
     A \textit{multi-layer perceptron} (MLP) is a neural network model $\mc{M}$, with $\hat{L}$ layers, defined by sequences of 
    weight matrices $(\mb{W}_1, \mb{W}_2, \dots, \mb{W}_{\hat{L}}), \quad \mb{W}_i \in \mathbb{Q}^{d_{i-1} \times d_i}$, 
    bias vectors $(\mb{b}_1, \mb{b}_2, \dots, \mb{b}_{\hat{L}}), \quad \mb{b}_i \in \mathbb{Q}^{d_i}$,  
    and (element-wise) ReLU functions $(f_1, f_2, \dots, f_{\hat{L}-1}), \quad f_i(x):= \max(0, x).$
    The final function is, without loss of generality, the binary step function $f_{\hat{L}}(x):= 1 \ \text{if} \ x \geq 0, \text{otherwise} \ 0.$ 
    The computation rules for $\mc{M}$ are given by 
    $\mb{h}_i := f_i(\mb{h}_{i-1} \mb{W} + \mb{b}_i), \quad \mb{h}_0:=\mb{x},$ 
    where $\mb{x}$ is the input. The output of $\mc{M}$ on $\mb{x}$ is defined as $\mc{M}(\mb{x}):= \mb{h}_{\hat{L}}$. The graph $G_{\mc{M}} = (V, E)$ of $\mc{M}$ has a vertex for each component of each $\mb{h}_i$. All vertices in layer $i$ are connected by edges to all vertices of layer $i+1$, with no intra-layer connections. Edges carry weights according to $\mb{W}_i$, and vertices carry the components of $\mb{b}_i$ as biases. The size of $\mc{M}$ is defined as $|\mc{M}|:= |V|$.
\end{definition}

\subsection{Preliminary remarks}

As is the case for other work in this area \citep[(which might be called ``Applied Complexity Theory''][]{bassanLocalVsGlobal2024a,barcelo_model_2020}, we are not aiming at developing new mathematical techniques but rather deploying existing mathematical tools to answer important questions that connect to applications.
Part of the technical challenge we take up is to formalize problems of practical interest in simple (and if possible, elegant) ways that are readily understandable, and to prove their complexity properties efficiently (i.e., obtaining a one to many relation between proof constructions and meaningful results). This allows us to gain insights into the sources of complexity of problems, an investigation where the difficulty/complexity/intricacy of proofs are a liability.

We use `input queries' to refer to computational problems in \textit{explainability} and `circuit queries' for circuit discovery in \textit{inner interpretability}.
We make no claims as to whether one or the other query relates more to intuitive ideas of explanation or interpretation and merely use the latter as familiar pointers to the literature.

All of our proofs for local problem variants assume a particular input vector I, be it the all-0 or all-1 vector. Note that we can simulate these vectors by having zero weights on the input lines and putting appropriate 0 and 1 biases on the input neurons (a technique developed and used in our later-derived proofs but readily applicable to earlier ones). This causes the input to be `ignored', which renders our proofs correct under both integer and continuous inputs. Note this construction is in line with previous work \citep[e.g.,][]{bassanLocalVsGlobal2024a,barcelo_model_2020}.
Importantly, this highlights that the characteristics of the input are not important but rather there is a combinatorial 'heart' beating at the center of our circuit problems; namely, the selection of a subcircuit from exponential number of subcircuits. This combinatorial core can, if not tamed by appropriate restrictions on network and input structure, give rise to non-polynomial worst-case algorithmic complexity.

Proofs for global problem variants often employ constructions similar to those for local variants, with minor to medium (though crucial) differences. For completeness, the full construction is stated again to minimize errors and the need to check proofs other than those being examined.

\textbf{On the issue of real-world structure and formal complexity.} One example scenario where real-world statistics might act as mitigating forces with respect to computational hardness (of the general problems) is the case of high redundancy (related to our Circuit Robustness problem).
Redundancy can in some sense make circuit finding easier (as solutions are more abundant), but the benefit comes at a cost for interpretability through introducing identifiably issues.
As circuits supporting a particular behavior are more numerous (i.e., there is more redundancy), it might get easier to find them with heuristics, but since they are more numerous, they potentially represent competing explanations, which leads to the issue of identifiability.
This redundancy would be important to diagnose and characterize, an issue that our Circuit Robustness problem touches on.

\section{Local and Global Sufficient Circuit}

\noindent
{\sc Minimum locally sufficient circuit} (MLSC) \\
{\em Input}: A multi-layer perceptron $M$ of depth $cd_g$ with $\#n_{tot,g}$ neurons
              and maximum layer width $cw_g$, connection-value matrices $W_1, W_2, \ldots, 
              W_{cd_g}$, neuron bias vector $B$, a Boolean input vector $I$ of length 
              $\#n_{g,in}$, and integers $d$, $w$, and $\#n$ such that $1 \leq d \leq 
              cd_g$, $1 \leq w \leq cw_g$, and $1 \leq \#n \leq \#n_{tot,g}$. \\
{\em Question}: Is there a subcircuit $C$ of $M$ of depth $cd_r \leq d$ with $\#n_{tot,r}
              \leq \#n$ neurons and maximum layer width $cw_r \leq w$ that produces the
              same output on input $I$ as $M$?

\vspace*{0.1in}

\noindent
{\sc Minimum globally sufficient circuit} (MGSC) \\
{\em Input}: A multi-layer perceptron $M$ of depth $cd_g$ with $\#n_{tot,g}$ neurons
              and maximum layer width $cw_g$, connection-value matrices $W_1, W_2, \ldots, 
              W_{cd_g}$, neuron bias vector $B$, 
              and integers $d$, $w$, and $\#n$ such that 
              $1 \leq d \leq cd_g$, $1 \leq w \leq cw_g$, and $1 \leq \#n
              \leq \#n_{tot,g}$. \\
{\em Question}: Is there a subcircuit $C$ of $M$ of depth $cd_r \leq d$ with $\#n_{tot,r}
              \leq \#n$ neurons and maximum layer width $cw_r \leq w$ that produces the
              same output as $M$ on every possible Boolean input vector of
              length $\#_{in,g}$?

\vspace*{0.1in}


\vspace*{0.1in}

\noindent
Given a subset $x$ of the neurons in $M$, the subcircuit $C$ of $M$ based on
$x$ has the neurons in $x$ and all connections in $M$ among these neurons.
Note that in order for the output of $C$ to be equal to the output of $M$ on
input $I$, the numbers $\#n_{in,g}$ and $\#n_{out,g}$ of input and output 
neurons in $M$ must exactly equal the numbers $\#n_{in,r}$ and $\#n_{out,r}$
of input and output neurons in $C$; hence, no input or output neurons can be
deleted from $M$ in creating $C$. Following \citealt[page 4]{barcelo_model_2020}, all 
neurons in $M$ use the ReLU activation function and the output $x$ of each 
output neuron is stepped as necessary to be Boolean, i.e, $step(x) = 0$ if 
$x \leq 0$ and is $1$ otherwise.








\noindent
For a graph $G = (V, E)$, we shall assume an ordering on the vertices and edges in $V$ 
and $E$, respectively. For each vertex $v \in V$, let the complete neighbourhood $N_C(v)$ of $v$ be the set
composed of $v$ and the set of all vertices in $G$ that are adjacent to $v$
by a single edge, i.e., $v \cup \{ u ~ | ~ u ~ \in V ~ \rm{and} ~ (u,v) \in E\}$.
Finally, let VC$_B$ be the version of VC in which each vertex in 
$G$ has degree at most $B$.

\begin{figure}[t]
\begin{center}
\includegraphics[width=4.5in]{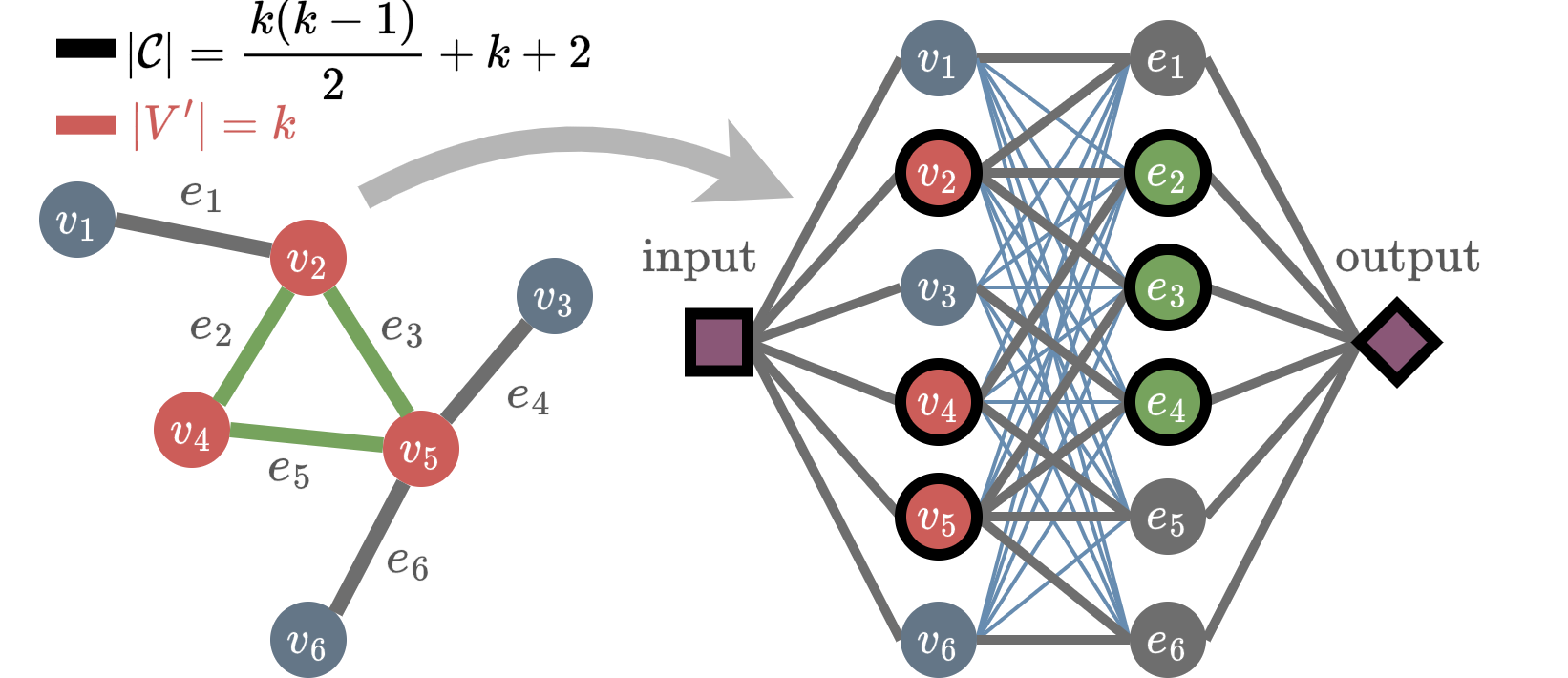}
\end{center}
\caption{This figure summarizes the reduction from \textsc{Clique}.}
%
\label{FigClique}
\end{figure}

We will prove various classical and parameterized results for MLSC and MGSC
using reductions from {\sc Clique} (Theorem \ref{ThmMLSCCli} and \ref{ThmMGSCCli}).
These reductions are summarized in Figure \ref{FigClique} and the
parameterized results are proved relative to the parameters in 
Table \ref{TabPrmMSCR}. Additional reductions from VC and DS 
(Theorems \ref{ThmMLSCVC} and \ref{ThmMSRDS}) use specialized ReLU logic
gates described in \citealt[Lemma 13]{barcelo_model_2020}. These gates assume Boolean
neuron input and output values of 0 and 1 and are structured as follows:

\begin{enumerate}
\item NOT ReLU gate:  A ReLU gate with one input connection weight of value
       $-1$ and a 
       bias of 1. This gate has output 1 if the input is 0 and 0 otherwise.
\item $n$-way AND ReLU gate: A ReLU gate with $n$ input connection weights 
       of value 1 and a bias of $-(n - 1)$. This gate has output 1 if all 
       inputs have value 1 and 0 otherwise.
\item $n$-way OR ReLU gate: A combination of an $n$-way AND ReLU gate with
       NOT ReLU gates on all of its inputs and a NOT ReLU gate on its 
       output that uses DeMorgan's Second Law to implement 
       $(x_1 \vee x_2 \vee \ldots x_n)$ as $\neg(\neg x_1 \wedge \neg x_2 
       \wedge \ldots \neg x_n)$. This gate has 
       output 1 if any input has value 1 and 0 otherwise.
\end{enumerate}

\begin{table}[t]
\caption{Parameters for the minimum sufficient subcircuit and reason 
         problems.
}
\vspace*{0.1in}
\label{TabPrmMSCR}
\centering
\begin{tabular}{| c || l | c |}
\hline
Parameter     & Description & Problem \\
\hline\hline
$cd_g$        & \# layers in given MLP & All \\
\hline
$cw_g$        & max \# neurons in layer in given MLP & All \\
\hline
$\#n_{tot,g}$ & total \# neurons in given MLP & All \\
\hline
$\#n_{in,g}$  & \# input neurons in given MLP & All \\
\hline
$\#n_{out,g}$ & \# output neurons in given MLP & All \\
\hline
$B_{\max,g}$  & max neuron bias in given MLP & All \\
\hline
$W_{\max,g}$  & max connection weight in given MLP & All \\
\hline\hline
$cd_r$        & \# layers in requested subcircuit & M\{L,G\}SC \\
\hline
$cw_r$        & max \# neurons in layer in requested subcircuit & M\{L,G\}SC \\
\hline
$\#n_{tot,r}$ & total \# neurons in requested subcircuit & M\{L,G\}SC \\
\hline
$\#n_{in,r}$  & \# input neurons in requested subcircuit & M\{L,G\}SC \\
\hline
$\#n_{out,r}$ & \# output neurons in requested subcircuit & M\{L,G\}SC \\
\hline
$B_{\max,r}$  & max neuron bias in requested subcircuit & M\{L,G\}SC \\
\hline
$W_{\max,r}$  & max connection weight in requested subcircuit & M\{L,G\}SC \\
\hline\hline
$k$           & Size of requested subset of input vector & MSR \\
\hline
\end{tabular} 
\end{table}

\subsection{Results for MLSC}
Towards proving NP-completeness, we first prove membership and then follow up with hardness.
Membership of MLSC in NP can be proven via the definition of the polynomial hierarchy and the following alternating quantifier formula:
$$\exists [\mc{C} \subseteq \mc{M}] : \mc{C}(\mb{x}) = \mc{M}(\mb{x})$$

\begin{theorem}
If MLSC is polynomial-time tractable then $P = NP$.
\label{ThmMLSCCli}
\end{theorem}
\begin{proof}
Consider the following reduction from {\sc Clique} to MLSC.
Given an instance $\la G = (V,E), k\ra$ of {\sc Clique}, construct the following 
instance $\la M, I, d, w, \#n\ra$ of MLSC:
Let $M$ be an MLP based on $\#n_{tot,g} = |V| + |E| + 2$ neurons spread
across four layers:

\begin{enumerate}
\item {\bf Input layer}: The single input neuron $n_{in}$ (bias 0).
\item {\bf Hidden vertex layer}: The vertex neurons $nv_1, nv_2, \ldots nv_{|V|}$
        (all with bias 0).
\item {\bf Hidden edge layer}: The edge neurons $ne_1, ne_2, \ldots ne_{|E|}$
        (all with bias $-1$).
\item {\bf Output layer}: The single output neuron $n_{out}$ (bias $-(k(k - 1)/2 - 1)$).
\end{enumerate}

\noindent
The non-zero weight connections between adjacent layers are as follows:

\begin{itemize}
\item The input neuron $n_{in}$ is connected to each vertex neuron with weight 1.
\item Each vertex neuron $nv_i$, $1 \leq i \leq |V|$, is connected to each edge neuron 
       whose corresponding edge has an endpoint $v_i$ with weight 1.
\item Each edge neuron $ne_i$, $1 \leq i \leq |E|$, is connected to the output neuron
       $n_{out}$ with weight 1.
\end{itemize}

\noindent
All other connections between neurons in adjacent layers have weight 0.
Finally, let $I = (1)$, $d = 4$, $w = k(k - 1)/2$, and $\#n = k(k - 1)/2 + k + 2$.
Observe that this instance of MLSC can be created in time polynomial in the size of the 
given instance of {\sc Clique}. Moreover, the output behaviour of the neurons in 
$M$ from the presentation of input $I$ until the output is generated is as follows:

\begin{center}
\begin{tabular}{| c || l |}
\hline
timestep & neurons (outputs) \\
\hline\hline
0 & --- \\
\hline
1 & $n_{in} (1)$ \\
\hline
2 & $nv_1 (1), nv_2 (1), \ldots nv_{|V|} (1)$ \\
\hline
3 & $ne_1 (1), ne_2 (1), \ldots ne_{|E|} (1)$ \\
\hline
4 & $n_{out} (|E| - (k(k - 1)/2 - 1))$ \\
\hline
\end{tabular}
\end{center}
 
\noindent
Note that it is the stepped output of $n_{out}$ in timestep 4 that yields output 1.

We now need to show the correctness of this reduction by proving that the answer
for the given instance of {\sc Clique} is ``Yes'' if and only if the answer for the 
constructed instance of MLSC is ``Yes''. We prove the two 
directions of this if and only if separately as follows:

\begin{description}
\item [$\Rightarrow$]: Let $V' = \{v'_1, v'_2, \ldots, v'_k\} \subseteq V$ be a clique
       in $G$ of size $k$. Consider the subcircuit $C$ based on
       neurons $n_{in}$, $n_{out}$, $\{nv' ~|~ v' \in V'\}$, and $\{ne' ~ | ~
       e' = (x, y) ~ {\rm and} ~ vx, vy \in V'\}$. Observe that in this subcircuit, $cd_r
       = d = 4$, $cw_r = r = k(k - 1)/2$, and $\#n_{tot,r} = \#n = k(k - 1)/2 + k + 2$.
       The output behaviour of the neurons in $C$ from the 
       presentation of input $I$ until the output is generated is as follows:

\begin{center}
\begin{tabular}{| c || l |}
\hline
timestep & neurons (outputs) \\
\hline\hline
0 & --- \\
\hline
1 & $n_{in} (1)$ \\
\hline
2 & $nv'_1 (1), nv'_2 (1), \ldots nv'_{|V'|} (1)$ \\
\hline
3 & $ne'_1 (1), ne'_2 (1), \ldots ne'_{k(k - 1)/2} (1)$ \\
\hline
4 & $n_{out} (1)$ \\
\hline
\end{tabular}
\end{center}
 
       \noindent
       The output of $n_{out}$ in timestep 4 is stepped to 1, which means that $C$
       is behaviorally equivalent to $M$ on $I$.
\item [$\Leftarrow$]: Let $C$ be a subcircuit of $M$ that is behaviorally equivalent to $M$
       on input $I$ and has $\#n_{tot,r} \leq \#n = k(k -1)/2 + k + 2$ neurons. As neurons
       in all four layers in $M$ must be present in $C$ to produce the required output,
       $cd_r = d = cd_g$ and both $n_{in}$ and $n_{out}$ are in $C$. In order for $n_{out}$
       to produce a non-zero output, there must be at least $k(k - 1)/2$ edge neurons in
       $C$, each of which must be activated by the inclusion of the vertex neurons 
       corresponding to both of their endpoint vertices. This requires the inclusion of at 
       least $k$ vertex neurons in $C$, as a set $V''$ of vertices in graph can have at 
       most $|V''|(|V''| - 1)/2$ distinct edges between them (with this maximum occurring 
       if all pairs of vertices in $V''$ have an edge between them). As $\#n_{tot,r} \leq 
       k(k - 1)/2 + k + 2$, all of the above implies that there must be exactly 
       $k(k - 1)/2$ edge neurons and exactly $k$ vertex neurons in $C$ and the vertices
       in $G$ corresponding to these vertex neurons must form a clique of size $k$ in $G$.
\end{description}

\noindent
As {\sc Clique} is $NP$-hard \citep{garey1979computers}, the reduction above establishes that MLSC is also
$NP$-hard.  The result follows from the definition of $NP$-hardness.
\end{proof}

\begin{theorem}
If $\la cd_g, \#n_{in,g}, \#n_{out,g}, B_{\max,g}, W_{\max,g}, cd_r, cw_r, \#n_{in,r}, 
        \#n_{out,r}, \#n_{tot,r},$ \linebreak $B_{\max,r}, W_{\max,r}\ra$-MLSC is fixed-parameter 
        tractable then $FPT = W[1]$.
\label{ThmMLSC_fpi1}
\end{theorem}
\begin{proof}
Observe that in the instance of MLSC constructed in the reduction in the proof of
Theorem \ref{ThmMLSCCli}, $cd_g = cd_r = 4$, $\#n_{in,g} = \#n_{in,r} = \#n_{out,r} =
\#n_{out,r} = W_{\max,g} = W_{\max,r} = 1$, and $B_{\max,g}, B_{\max,r}, \#n_{tot,r},$
and $cw_r$ are all functions of $k$ in the given instance of {\sc Clique}. The result
then follows from the fact that $\la k \ra$-{\sc Clique} is $W[1]$-hard \citep{downeyParameterizedComplexity1999a}.
\end{proof}

\begin{theorem}
$\la \#n_{tot,g}\ra$-MLSC is fixed-parameter tractable.
\label{ThmMLSC_fpt1}
\end{theorem}
\begin{proof}
Consider the algorithm that generates each possible subcircuit of $M$ and checks if that
subcircuit is behaviorally equivalent to $M$ on input $I$. If such a subcircuit is
found, return ``Yes''; otherwise, return ``No''. As each such subcircuit can be
run on $I$ in time polynomial in the size of the given instance of MLSC and the total
number of subcircuits that need to be checked is at most $2^{\#n_{tot,g}}$, the above is
a fixed-parameter tractable algorithm for MLSC relative to parameter-set $\{\#n_{tot,g}\}$.
\end{proof}

\begin{theorem}
$\la cd_g, cw_g\ra$-MLSC is fixed-parameter tractable.
\label{ThmMLSC_fpt2}
\end{theorem}
\begin{proof}
Follows from the observation that $\#n_{tot,g} \leq cd_g \times cw_g$ and the
algorithm in the proof of Theorem \ref{ThmMLSC_fpt1}.
\end{proof}

\vspace*{0.15in}

\begin{figure}[t]
\begin{center}
\includegraphics[width=4.5in]{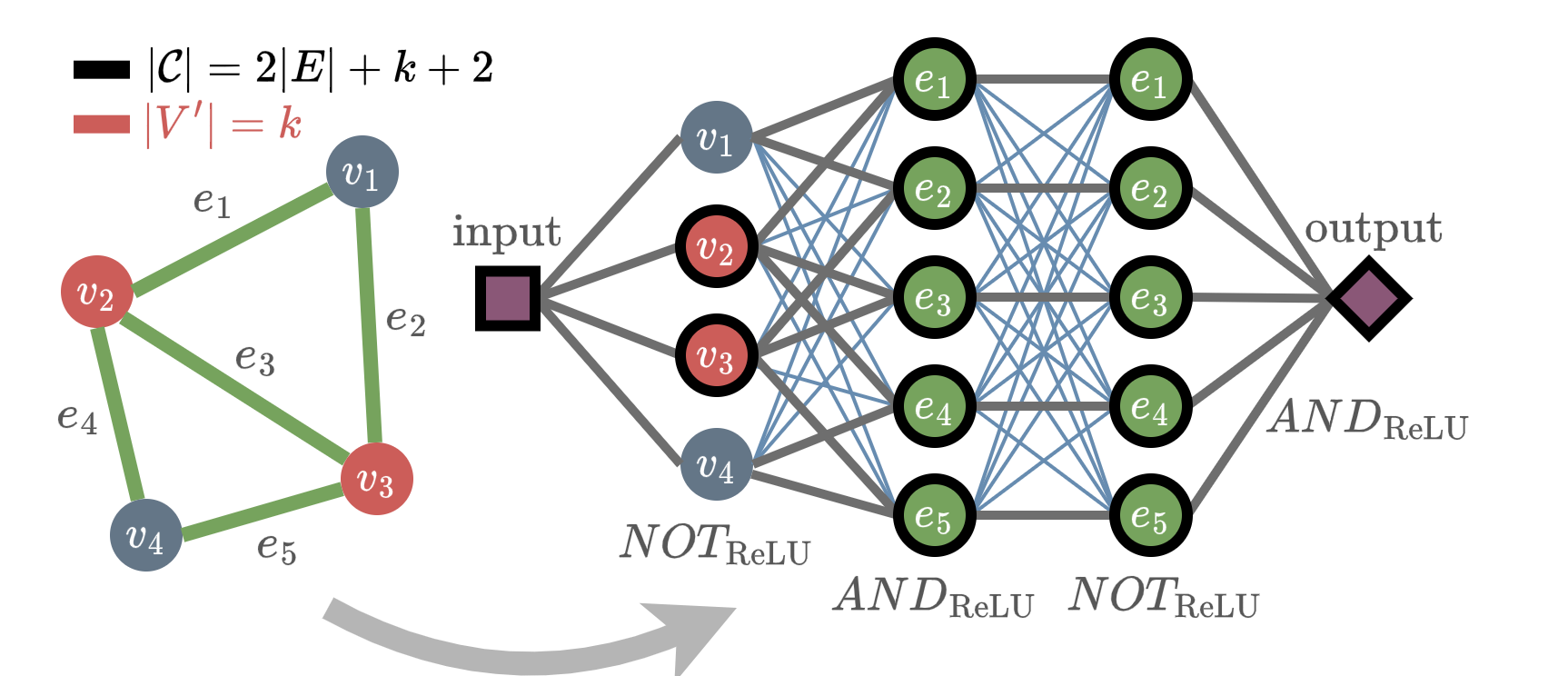}
\end{center}
\caption{This figure summarizes the reduction from \textsc{Vertex Cover}}
\label{FigVC}
\end{figure}

Though we have already proved the polynomial-time intractability of MLSC in Theorem 
\ref{ThmMLSCCli}, the reduction in the proof of the following theorem will be useful in 
proving a certain type of polynomial-time inapproximability for MLSC (see Figure \ref{FigVC}).

\begin{theorem}
If MLSC is polynomial-time tractable then $P = NP$.
\label{ThmMLSCVC}
\end{theorem}
\begin{proof}
Consider the following reduction from VC to MLSC.
Given an instance $\la G = (V,E), k\ra$ of VC, construct the following 
instance $\la M, I, d, w, \#n\ra$ of MLSC:
Let $M$ be an MLP based on $\#n_{tot,g} = |V| + 2|E| + 2$ neurons spread
across five layers:

\begin{enumerate}
\item {\bf Input layer}: The single input neuron $n_{in}$ (bias 0).
\item {\bf Hidden vertex layer}: The vertex neurons $nvN_1, nvN_2, \ldots nvN_{|V|}$, all of which are NOT ReLU gates.
\item {\bf Hidden edge layer I}: The edge AND neurons $neA_1, neA_2, \ldots neA_{|E|}$, all of which are 2-way AND ReLU gates.
\item {\bf Hidden edge layer II}: The edge NOT neurons $neN_1, neN_2, \ldots neN_{|E|}$, all of which are NOT ReLU gates.
\item {\bf Output layer}: The single output neuron $n_{out}$, which is an
       $|E|$-way AND ReLU gate.
\end{enumerate}

\noindent
The non-zero weight connections between adjacent layers are as follows:

\begin{itemize}
\item The input neuron $n_{in}$ is connected to each vertex NOT neuron with weight 1.
\item Each vertex NOT neuron $nvN_i$, $1 \leq i \leq |V|$, is connected to each edge AND neuron 
       whose corresponding edge has an endpoint $v_i$ with weight 1.
\item Each edge AND neuron $neA_i$, $1 \leq i \leq |E|$, is connected to its corresponding edge NOT neuron $neN_i$ with weight 1.
\item Each edge NOT neuron $neN_i$, $1 \leq i \leq |E|$, is connected to the output neuron
       $n_{out}$ with weight 1.
\end{itemize}

\noindent
All other connections between neurons in adjacent layers have weight 0.
Finally, let $I = (1)$, $d = 5$, $w = |E|$, and $\#n = 2|E| + k + 2$.
Observe that this instance of MLSC can be created in time polynomial in the size of the 
given instance of VC, Moreover, the output behaviour of the neurons in 
$M$ from the presentation of input $I$ until the output is generated is as follows:

\begin{center}
\begin{tabular}{| c || l |}
\hline
timestep & neurons (outputs) \\
\hline\hline
0 & --- \\
\hline
1 & $n_{in} (1)$ \\
\hline
2 & $nvN_1 (0), nvN_2 (0), \ldots nvN_{|V|} (0)$ \\
\hline
3 & $neA_1 (0), neA_2 (0), \ldots neA_{|E|} (0)$ \\
\hline
4 & $neN_1 (1), neN_2 (1), \ldots neN_{|E|} (1)$ \\
\hline
5 & $n_{out} (1)$ \\
\hline
\end{tabular}
\end{center}
 
We now need to show the correctness of this reduction by proving that the answer
for the given instance of VC is ``Yes'' if and only if the answer for the 
constructed instance of MLSC is ``Yes''. We prove the two 
directions of this if and only if separately as follows:

\begin{description}
\item [$\Rightarrow$]: Let $V' = \{v'_1, v'_2, \ldots, v'_k\} \subseteq V$ be a vertex cover
       in $G$ of size $k$. Consider the subcircuit $C$ based on
       neurons $n_{in}$, $n_{out}$, $\{nv'N ~|~ v' \in V'\}$, 
       $\{neA_1, neA_2, \ldots, neA_{|E|}\}$, and
       $\{neN_1, neN_2, \ldots, neN_{|E|}\}$.  
       Observe that in this subcircuit,
       $cd_r = d = 5$, $cw_r = w = |E|$, and $\#n_{tot,r} = \#n = 2|E| + k + 2$.
       The output behaviour of the neurons in $C$ from the 
       presentation of input $I$ until the output is generated is as follows:

\begin{center}
\begin{tabular}{| c || l |}
\hline
timestep & neurons (outputs) \\
\hline\hline
0 & --- \\
\hline
1 & $n_{in} (1)$ \\
\hline
2 & $nvN'_1 (0), nvN'_2 (0), \ldots nvN'_{|V'|} (0)$ \\
\hline
3 & $neA_1 (0), neA_2 (0), \ldots neA_{|E|} (0)$ \\
\hline
4 & $neN_1 (1), neN_2 (1), \ldots neN_{|E|} (1)$ \\
\hline
5 & $n_{out} (1)$ \\
\hline
\end{tabular}
\end{center}
 
       \noindent
       This means that $C$ is behaviorally equivalent to $M$ on $I$.
\item [$\Leftarrow$]: Let $C$ be a subcircuit of $M$ that is behaviorally equivalent to $M$
       on input $I$ and has $\#n_{tot,r} \leq \#n = 2|E| + k + 2$ neurons. As neurons in
       all five layers in $M$ must be present in $C$ to produce the required output,
       $cd_r = cd_g$ and both $n_{in}$ and $n_{out}$ are in $C$. In order for $n_{out}$
       to produce a non-zero output, there must be at least $|E|$ edge NOT neurons and $|E|$ AND neurons in $C$, and each
       of the latter must be connected to at least one of the vertex NOT neurons corresponding to 
       their endpoint vertices. As $\#n_{tot,r} \leq 2|E| + k + 2$, there must be 
       exactly $|E|$ edge NOT neurons, $|E|$ edge AND neurons, and $k$ vertex NOT neurons in $C$ and the vertices in $G$ 
       corresponding to these vertex NOT neurons must form a vertex cover of size $k$ in $G$.
\end{description}

\noindent
As VC is $NP$-hard \citep{garey1979computers}, the reduction above establishes that MLSC is also
$NP$-hard.  The result follows from the definition of $NP$-hardness.
\end{proof}

\vspace*{0.15in}

\noindent
We now define our two notions of polynomial-time approximation.
For a minimization problem $\Pi$, let $OPT_{\Pi}(I)$ be an optimal solution
for $\Pi$, $A_{\Pi}(I)$ be a solution for$\Pi$ returned by an algorithm $A$,
and $m(OPT_{\Pi}(I))$ and $m(A_{\Pi}(I))$ be the values of these solutions.
Consider the following alternative to approximation algorithms that give
solutions that are within an additive factor of optimal.

\begin{definition}
\citep[Definition 3.5]{ausielloComplexityApproximation1999}
Given a minimization problem $\Pi$, an algorithm $A$ is a {\bf (multiplicative)
$\epsilon$-approximation algorithm for $\Pi$} if for each instance $I$ of $\Pi$,
$m(A_{\Pi}(I)) - m(OPT_{\Pi}(I)) \leq \epsilon \times m(OPT_{\Pi}(I))$.
\end{definition}

\noindent
It would be ideal if one could obtain approximate solutions for a problem
$\Pi$ that are arbitrarily close to optimal if one is willing to allow
extra algorithm runtime. This is encoded in the following entity.

\begin{definition}
(Adapted from Definition 3.10 in \citealt{ausielloComplexityApproximation1999})
Given a minimization problem $\Pi$, a {\bf polynomial-time approximation scheme
(PTAS) for $\Pi$} is a set $\cal{A}$ of algorithms such that for each integer 
$k > 0$, there is a $\frac{1}{k}$-approximation algorithm $A^k_{\Pi} \in \cal{A}$ 
that runs in time polynomial in $|I|$.
\end{definition}

\noindent
The question of whether or not a problem has a PTAS can be answered using
the following type of approximation-preserving reducibility.

\begin{definition}
\cite[page 427]{PY91}
Given two minimization problems $\Pi$ and $\Pi'$, {\bf $\Pi$ L-reduces to $\Pi'$,
i.e., $\Pi \leq_L \Pi'$} if there are polynomial-time algorithms $f$ and $g$
and constants $\alpha, \beta > 0$ such that for each instance $I$ of $\Pi$

\begin{description}
\item[(L1)]{Algorithm $f$ produces an instance $I'$ of $\Pi'$ such that
             $m(OPT_{\Pi'}(I')) \leq \alpha \times m(OPT_{\Pi}(I))$; and
}
\item[(L2)]{For any solution for $I'$ with value $v'$, algorithm $g$ produces
             a solution for $I$ of value $v$ such that $v - m(OPT_{\Pi}(I))
             \leq \beta \times (v' - m(OPT_{\Pi'}(I')))$.
}
\end{description}
\end{definition}

\begin{lemma}
\cite[Theorem 1.2.2]{AL+98}
If an optimization problem that is $MAX ~ SNP$-hard under L-reductions has
a PTAS then $P = NP$.
\label{LemPTAS}
\end{lemma}

\begin{theorem}
If MLSC has a PTAS then $P = NP$.
\end{theorem}

\begin{proof}
We prove that the reduction from VC to MLSC in the proof of Theorem \ref{ThmMLSCVC} is
also an L-reduction from VC$_B$ to MLSC as follows:

\begin{itemize}
\item{Observe that $m(OPT_{VC_B}(I)) \geq |E|/B$ (the best case in which $G$ is a 
       collection of $B$-star subgraphs such that each edge is uniquely covered by 
       the central vertex of its associated star) and $m(OPT_{MLSC}(I')) \leq 2|E|
       + 2|E| + 2 = 4|E| + 2$ (the worst case in which the vertex neurons
       corresponding to the two endpoints of every edge in $G$ are selected).
       This gives us

\begin{eqnarray*}
m(OPT_{MLSC}(I')) & \leq & 4Bm(OPT_{VC_B}(I)) + 2 \\
                 & \leq & 4Bm(OPT_{VC_B}(I)) + 2Bm(OPT_{VC_B}(I)) \\
                 & \leq & 6Bm(OPT_{VC_B}(I)) \\
\end{eqnarray*}

              \noindent
              which satisfies condition L1 with $\alpha = 6B$.
}
\item{Observe that any solution $S'$ for for the constructed instance $I'$ of MLSC
       of value $k + 2|E| + 2$ implies a solution $S$ for the given instance $I$ of VC$_B$ 
       of size $k$, i.e., the vertices in $V$ corresponding to the selected vertex neurons 
       in $S$. Hence, it is the case that $m(S) - m(OPT_{VC_B}(I)) = m(S') - 
       m(OPT_{MLSC}(I'))$, which satisfies condition L2 with $\beta = 1$.
}
\end{itemize}

\noindent
As VC$_B$ is $MAX ~ SNP$-hard under L-reductions \cite[Theorem 2(d)]{PY91}, the L-reduction
above proves that MLSC is also $MAX ~ SNP$-hard under L-reductions. The result follows from
Lemma \ref{LemPTAS}.
\end{proof}

\subsection{Results for MGSC}

\begin{theorem}
If MGSC is polynomial-time tractable then $P = NP$.
\label{ThmMGSCCli}
\end{theorem}
\begin{proof}
Consider the following reduction from {\sc Clique} to MGSC. Given an instance $\la G = 
(V,E), k\ra$ of {\sc Clique}, construct an instance $\la M, d, w, \#n\ra$ of MGSC
as in the reduction in the proof of Theorem \ref{ThmMLSCCli}, omitting input vector $I$.
Observe that this instance of MGSC can be created in time polynomial in the size of the 
given instance of {\sc Clique}. As $\#n_{in,g} = 1$, there are only two possible Boolean
input vectors, $(0)$ and $(1)$. Given input vector $(1)$, as MLP $M$ in this reduction is
the same as $M$ in the proof of Theorem \ref{ThmMLSCCli}, the output in
timestep 4 is once again 1; moreover, given input vector $(0)$, no vertex or edge neurons
can have output 1 and hence the output in timestep 4 is 0.

We now need to show the correctness of this reduction by proving that the answer
for the given instance of {\sc Clique} is ``Yes'' if and only if the answer for the 
constructed instance of MGSC is ``Yes''. We prove the two 
directions of this if and only if separately as follows:

\begin{description}
\item [$\Rightarrow$]: Let $V' = \{v'_1, v'_2, \ldots, v'_k\} \subseteq V$ be a clique
       in $G$ of size $k$. Consider the subcircuit $C$ based on
       neurons $n_{in}$, $n_{out}$, $\{nv' ~|~ v' \in V'\}$, and $\{ne' ~ | ~
       e' = (x, y) ~ {\rm and} ~ vx, vy \in V'\}$. Observe that in this subcircuit,
       $cd_r = 4$, $cw_r = k(k - 1)/2$, and $\#n_{tot,r} = k(k - 1)/2 + k + 2$.
       Given input $(1)$, the output behaviour of the neurons in $C$ from the 
       presentation of input until the output is generated is as follows:

\begin{center}
\begin{tabular}{| c || l |}
\hline
timestep & neurons (outputs) \\
\hline\hline
0 & --- \\
\hline
1 & $n_{in} (1)$ \\
\hline
2 & $nv'_1 (1), nv'_2 (1), \ldots nv'_{|V'|} (1)$ \\
\hline
3 & $ne'_1 (1), ne'_2 (1), \ldots ne'_{k(k - 1)/2} (1)$ \\
\hline
4 & $n_{out} (1)$ \\
\hline
\end{tabular}
\end{center}
 
       \noindent
       Moreover, given input $(0)$, no vertex or edge neurons in $C$ can have output 1
       and the output of $C$ at timestep 4 is 0. This means that $C$ is behaviorally 
       equivalent to $M$ on all possible Boolean input vectors
\item [$\Leftarrow$]: Let $C$ be a subcircuit of $M$ that is behaviorally equivalent to $M$
       on all possible Boolean input vectors and has $\#n_{tot,r} \leq \#n = k(k -1)/2 + k
       + 2$ neurons. Consider the case of input vector $(1)$. This vector must cause $C$ to
       generate output 1 at timestep 4 as $C$ is behaviorally equivalent to $M$ on all
       Boolean input vectors. As neurons in all four layers in $M$ must be present in $C$ 
       to produce the required output, $cd_r = cd_g$ and both $n_{in}$ and $n_{out}$ are in
       $C$. In order for $n_{out}$ to produce a non-zero output, there must be at least 
       $k(k - 1)/2$ edge neurons in $C$, each of which must be activated by the inclusion 
       of the vertex neurons corresponding to both of their endpoint vertices. This 
       requires the inclusion of at least $k$ vertex neurons in $C$, as a set $V''$ of 
       vertices in graph can have at most $|V''|(|V''| - 1)/2$ distinct edges between them
       (with this maximum occurring if all pairs of vertices in $V''$ have an edge between
       them). As $\#n_{tot,r} \leq k(k - 1)/2 + k + 2$, all of the above implies that there
       must be exactly $k(k - 1)/2$ edge neurons and exactly $k$ vertex neurons in $C$ and
       the vertices in $G$ corresponding to these vertex neurons must form a clique of 
       size $k$ in $G$.
\end{description}

\noindent
As {\sc Clique} is $NP$-hard \citep{garey1979computers}, the reduction above establishes that MGSC is also
$NP$-hard.  The result follows from the definition of $NP$-hardness.
\end{proof}

\begin{theorem}
If $\la cd_g, \#n_{in,g}, \#n_{out,g}, B_{\max,g}, W_{\max,g}, cd_r, cw_r, \#n_{in,r}, 
        \#n_{out,r}, \#n_{tot,r},$ \linebreak $B_{\max,r}, W_{\max,r}\ra$-MGSC is fixed-parameter 
        tractable then $FPT = W[1]$.
\label{ThmMGSC_fpi1}
\end{theorem}
\begin{proof}
Observe that in the instance of MGSC constructed in the reduction in the proof of
Theorem \ref{ThmMGSCCli}, $cd_g = cd_r = 4$, $\#n_{in,g} = \#n_{in,r} = \#n_{out,r} =
\#n_{out,r} = W_{\max,g} = W_{\max,r} = 1$, and $B_{\max,g}, B_{\max,r}, \#n_{tot,r},$
and $cw_r$ are all functions of $k$ in the given instance of {\sc Clique}. The result
then follows from the fact that $\la k \ra$-{\sc Clique} is $W[1]$-hard \citep{downeyParameterizedComplexity1999a}.
\end{proof}

\begin{theorem}
$\la \#n_{tot,g}\ra$-MGSC is fixed-parameter tractable.
\label{ThmMGSC_fpt1}
\end{theorem}
\begin{proof}
Consider the algorithm that generates each possible subcircuit of $M$ and checks if that
subcircuit is behaviorally equivalent to $M$ on all possible Boolean input vectors of 
length $\#n_{in,g}$. If such a subcircuit is found, return ``Yes''; otherwise, return 
``No''. There are $2^{\#n_{in,g}}$ possible Boolean input vectors and the total number of 
subcircuits that need to be checked is at most $2^{\#n_{tot,g}}$. As $\#n_{in,g} \leq 
\#n_{tot,g}$ and each such subcircuit can be run on an input vector in time polynomial in 
the size of the given instance of MGSC, the above is a fixed-parameter tractable algorithm
for MGSC relative to parameter-set $\{\#n_{tot,g}\}$.
\end{proof}

\begin{theorem}
$\la cd_g, cw_g\ra$-MGSC is fixed-parameter tractable.
\label{ThmMGSC_fpt2}
\end{theorem}
\begin{proof}
Follows from the observation that $\#n_{tot,g} \leq cd_g \times cw_g$ and the
algorithm in the proof of Theorem \ref{ThmMGSC_fpt1}.
\end{proof}

\vspace*{0.15in}

Let us now consider the PTAS-approximability of MGSC. A first thought
would be to -reuse the reduction in the proof of Theorem
\ref{ThmMLSCVC} if the given MLP $M$ and VC subcircuit are 
behaviorally equivalent under both possible input vectors, $(1)$ and $(0)$.
We already know the former is true. With respect to the latter, observe
that the output behaviour of the neurons in $M$ from the presentation of 
input $(0)$ until the output is generated is as follows:

\begin{center}
\begin{tabular}{| c || l |}
\hline
timestep & neurons (outputs) \\
\hline\hline
0 & --- \\
\hline
1 & $n_{in} (0)$ \\
\hline
2 & $nvN_1 (1), nvN_2 (1), \ldots nvN_{|V|} (1)$ \\
\hline
3 & $neA_1 (1), neA_2 (1), \ldots neA_{|E|} (1)$ \\
\hline
4 & $neN_1 (0), neN_2 (0), \ldots neN_{|E|} (0)$ \\
\hline
5 & $n_{out} (0)$ \\
\hline
\end{tabular}
\end{center}
 
\noindent
However, in the VC subcircuit, we are no longer guaranteed that both
endpoint vertex NOT neurons for any edge AND neuron (let alone the endpoint
vertex NOT neurons for all edge AND neurons) will be the vertex cover 
encoded in the subcircuit. This means that all
edge AND neurons could potentially output 0, which would cause
$M$ to output 1 at timestep 4.

This problem can be fixed if we can modify the given VC graph $G$ to create
a graph $G'$ such that 

\begin{enumerate}
\item we can guarantee that both endpoint vertex NOT neurons for at least 
       one edge AND neuron are present in a VC subcircuit $C$ constructed 
       for $G'$ (which would make at least one edge AND neuron output 1 
       and cause $C$ to output 0 at timestep 4); and
\item we can easily extract a graph vertex cover of size at most $k$ for 
       $G$ from any vertex cover of a particular size for $G'$.
\end{enumerate}

\begin{figure}[t]
\begin{center}
\includegraphics[width=4.5in]{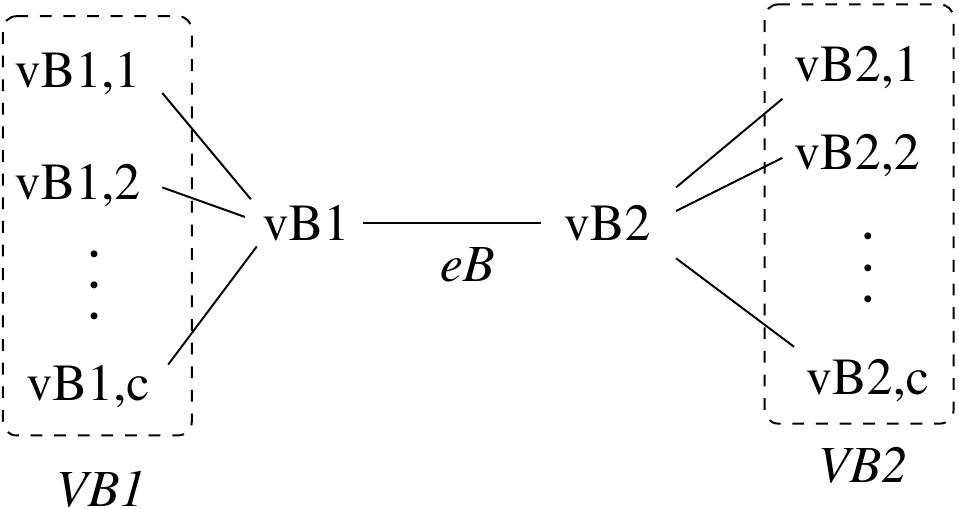}
\end{center}
\caption{The $c$-way Bowtie Graph $B_c$.}
\label{FigBowtie}
\end{figure}

\noindent
To do this, we shall use the {\bf $c$-way bowtie graph $B_c$}. For $c > 0$,
$B_c$ consists of a central edge $e_B$ between vertices $v_{B1}$ 
and $v_{B2}$ such that $c$ edges radiate outwards from $v_{B1}$ and 
$v_{B2}$ to the $c$-sized vertex-sets $V_{B1} = \{v_{B1,1}, v_{B1,2}, \ldots, 
v_{B1,c}\}$ and $V_{B2} = \{v_{B2,1}, v_{B2,2}, \ldots, v_{B2,c}\}$, respectively 
(see Figure \ref{FigBowtie}). Note that such a graph has $2c + 2$ vertices
and $2c + 1$ edges. Given a graph $G = (V,E)$ with no isolated vertices 
such that $|V| \leq 2|E|$ (with the minimum occurring in a graph consisting
of $|E|$ endpoint-disjoint edges), let $Bow(G) = B_{4|E|} \cup G$. This
graph has the following useful property.

\begin{lemma}
Given a graph $G = (V,E)$ and a positive integer $k \leq |V|$, if $Bow(G)$ 
has a vertex cover $V'$ of size at most $k + 2$ then (1) $\{v_{B1}, 
v_{B2}\} \in V'$ and (2) $G$ has a vertex cover of size at most $k$.
\label{LemBowVC}
\end{lemma}
\begin{proof}
Let us prove the two consequent clauses as follows:

\begin{enumerate}
\item Any vertex cover $V'$ of $Bow(G)$ must cover all the edges in both
       $B_{4|E|}$ and $G$. Suppose $\{v_{B1}, v_{B2}\} \not\in V'$. In
       order to cover the edges in $B_{4|E|}$, all $8|E|$ vertices in
       $V_{B1} \cup V_{B2}$ must be in $V'$. This is however impossible
       as $|V'| \leq k + 2 \leq |V| + 2 \leq 2|E| + 2 \leq 4|E < 8|E|$.
       Similarly, suppose only one of $v_{B1}$ and $v_{B2}$ is in $V'$;
       let us assume it is $v_{B1}$. In that case, all vertices in $V_{B2}$
       must be in $V'$. However, this too is impossible as $|V' - 
       \{v_{B1}\}| \leq k + 1 \leq |V| + 1 \leq 2|E| + 1 \leq 3|E| < 4|E| =
       |V_{B2}|$. Hence, both $v_{B1}$ and $v_{B2}$ must be in $V'$.
\item Given (1), $k' \leq k$ vertices remain in $V'$ to cover $G$. All $k'$
       of these vertices need not be in $G$, e.g., some may be scattered 
       over $V_{B1}$ and $V_{B2}$. That being said, it is still the case 
       that $G$ must have a vertex cover of size at most $k$.
\end{enumerate}

\noindent
This concludes the proof.
\end{proof}

\begin{theorem}
If MGSC is polynomial-time tractable then $P = NP$.
\label{ThmMGSCVC}
\end{theorem}
\begin{proof}
Consider the following reduction from VC to MGSC. Given an instance $\la 
G = (V,E), k\ra$ of VC, construct the following instance $\la M, d, w, 
\#n\ra$ of MGSC based on $G' = (V', E') = Bow(G)$: Let $M$ be an MLP based
on $\#n_{tot,g} = |V'| + 2|E'| + 2$ neurons spread
across five layers:

\begin{enumerate}
\item {\bf Input layer}: The single input neuron $n_{in}$ (bias 0).
\item {\bf Hidden vertex layer}: The vertex neurons $nvN_1, nvN_2, \ldots nvN_{|V'|}$, all of which are NOT ReLU gates.
\item {\bf Hidden edge layer I}: The edge AND neurons $neA_1, neA_2, \ldots neA_{|E'|}$, all of which are 2-way AND ReLU gates.
\item {\bf Hidden edge layer II}: The edge NOT neurons $neN_1, neN_2, \ldots neN_{|E'|}$, all of which are NOT ReLU gates.
\item {\bf Output layer}: The single output neuron $n_{out}$, which is an
       $|E'|$-way AND ReLU gate.
\end{enumerate}

\noindent
The non-zero weight connections between adjacent layers are as follows:

\begin{itemize}
\item The input neuron $n_{in}$ is connected to each vertex NOT neuron with weight 1.
\item Each vertex NOT neuron $nvN_i$, $1 \leq i \leq |V'|$, is connected to each edge AND neuron 
       whose corresponding edge has an endpoint $v'_i$ with weight 1.
\item Each edge AND neuron $neA_i$, $1 \leq i \leq |E'|$, is connected to its corresponding edge NOT neuron $neN_i$ with weight 1.
\item Each edge NOT neuron $neN_i$, $1 \leq i \leq |E'|$, is connected to the output neuron
       $n_{out}$ with weight 1.
\end{itemize}

\noindent
All other connections between neurons in adjacent layers have weight 0.
Finally, let $d = 5$, $w = |E'|$, and $\#n = 2|E'| + (k + 2) + 2 =
2|E'| = k + 4$. Observe that this instance of MGSC can be created in time 
polynomial in the size of the given instance of VC, the output behaviour of
the neurons in $M$ from the presentation of input $(1)$ until the output is
generated is as follows:

\begin{center}
\begin{tabular}{| c || l |}
\hline
timestep & neurons (outputs) \\
\hline\hline
0 & --- \\
\hline
1 & $n_{in} (1)$ \\
\hline
2 & $nvN_1 (0), nvN_2 (0), \ldots nvN_{|V'|} (0)$ \\
\hline
3 & $neA_1 (0), neA_2 (0), \ldots neA_{|E'|} (0)$ \\
\hline
4 & $neN_1 (1), neN_2 (1), \ldots neN_{|E'|} (1)$ \\
\hline
5 & $n_{out} (1)$ \\
\hline
\end{tabular}
\end{center}
 
\noindent
and the output behaviour of the neurons in $M$ from the presentation of 
input $(0)$ until the output is generated is as follows:

\begin{center}
\begin{tabular}{| c || l |}
\hline
timestep & neurons (outputs) \\
\hline\hline
0 & --- \\
\hline
1 & $n_{in} (0)$ \\
\hline
2 & $nvN_1 (1), nvN_2 (1), \ldots nvN_{|V'|} (1)$ \\
\hline
3 & $neA_1 (1), neA_2 (1), \ldots neA_{|E'|} (1)$ \\
\hline
4 & $neN_1 (0), neN_2 (0), \ldots neN_{|E'|} (0)$ \\
\hline
5 & $n_{out} (0)$ \\
\hline
\end{tabular}
\end{center}
 
We now need to show the correctness of this reduction by proving that the 
answer for the given instance of VC is ``Yes'' if and only if the answer 
for the constructed instance of MGSC is ``Yes''. We prove the two 
directions of this if and only if separately as follows:

\begin{description}
\item [$\Rightarrow$]: Let $V'' = \{v''_1, v''_2, \ldots, v''_k\} \subseteq
       V$ be a vertex cover in $G$ of size $k$. Consider the subcircuit $C$
       based on neurons $n_{in}$, $n_{out}$, $\{nv''N ~|~ v'' \in V''\}
       \cup \{nv_{B1}N, nv_{B2}B\}$, $\{neA_1, neA_2, \ldots, 
       neA_{|E'|}\}$, and $\{neN_1, neN_2, \ldots, neN_{|E'|}\}$. Observe 
       that in this subcircuit, $cd_r = d = 5$, $cw_r = w = |E'|$, and 
       $\#n_{tot,r} = \#n = 2|E'| + k + 4$. The output behaviour of the neurons 
       in $C$ from the presentation of input $(1)$ until the output is 
       generated is as follows:

\begin{center}
\begin{tabular}{| c || l |}
\hline
timestep & neurons (outputs) \\
\hline\hline
0 & --- \\
\hline
1 & $n_{in} (1)$ \\
\hline
2 & All vertex NOT neurons (0) \\
\hline
3 & $neA_1 (0), neA_2 (0), \ldots neA_{|E'|} (0)$ \\
\hline
4 & $neN_1 (1), neN_2 (1), \ldots neN_{|E'|} (1)$ \\
\hline
5 & $n_{out} (1)$ \\
\hline
\end{tabular}
\end{center}
 
       \noindent
       Moreover, the output behaviour of the neurons in $C$ from the 
       presentation of input $(0)$ until the output is generated is as 
       follows:

\begin{center}
\begin{tabular}{| c || l |}
\hline
timestep & neurons (outputs) \\
\hline\hline
0 & --- \\
\hline
1 & $n_{in} (0)$ \\
\hline
2 & All vertex NOT neurons (1) \\
\hline
3 & At least one edge AND neuron has output 1, e.g., $ne_{B}A$ \\
\hline
4 & At least one edge NOT neuron has output 0, e.g., $ne_{B}N$ \\
\hline
5 & $n_{out} (0)$ \\
\hline
\end{tabular}
\end{center}
 
       \noindent
       This means that $C$ is behaviorally equivalent to $M$ on all
       possible Boolean input vectors.
\item [$\Leftarrow$]: Let $C$ be a subcircuit of $M$ that is behaviorally 
       equivalent to $M$ on all possible Boolean input vectors and has 
       $\#n_{tot,r} \leq \#n = 2|E'| + k + 4$ neurons. As neurons in
       all five layers in $M$ must be present in $C$ to produce the 
       required output, $cd_r = cd_g$ and both $n_{in}$ and $n_{out}$ are 
       in $C$. In order for $n_{out}$ to produce a non-zero output, there 
       must be at least $|E'|$ edge NOT neurons and $|E'|$ AND neurons in 
       $C$, and each of the latter must be connected to at least one of the
       vertex NOT neurons corresponding to their endpoint vertices. As 
       $\#n_{tot,r} \leq 2|E'| + k + 4$, there must be exactly $|E|$ edge 
       NOT neurons, $|E'|$ edge AND neurons, and $k + 2$ vertex NOT neurons
       in $C$ and the vertices in $G'$ corresponding to these vertex NOT 
       neurons must form a vertex cover $V''$ of size $k + 2$ in $G'$. 
       However, as $G' = Bow(G)$, Lemma \ref{LemBowVC} implies not only 
       that $\{nv_{B1}N, nv_{B2}N\} \in V''$ but that $G$ has a vertex 
       cover of size at most $k$.
\end{description}

\noindent
As VC is $NP$-hard \citep{garey1979computers}, the reduction above establishes that MGSC is also
$NP$-hard.  The result follows from the definition of $NP$-hardness.
\end{proof}

\begin{theorem}
If MGSC has a PTAS then $P = NP$.
\end{theorem}

\begin{proof}
We prove that the reduction from VC to MLSC in the proof of Theorem 
\ref{ThmMGSCVC} is also an L-reduction from VC$_B$ to MGSC as follows:

\begin{itemize}
\item{Observe that $m(OPT_{VC_B}(I)) \geq |E|/B$ (the best case in which 
       $G$ is a collection of $B$-star subgraphs such that each edge is 
       uniquely covered by the central vertex of its associated star) and 
       $m(OPT_{MGSC}(I')) \leq 2|E'| + 2|E'| + 2 = 4|E'| + 2$ (the worst 
       case in which the vertex neurons corresponding to the two endpoints
       of every edge in $G$ are selected). As $|E'| = 9|E| + 1$, this 
       gives us

\begin{eqnarray*}
m(OPT_{MLSC}(I')) & \leq & 36|E| + 4 + 2 \\
                  & \leq & 36Bm(OPT_{VC_B}(I)) + 6 \\
                  & \leq & 36Bm(OPT_{VC_B}(I)) + 6Bm(OPT_{VC_B}(I)) \\
                  & \leq & 42Bm(OPT_{VC_B}(I)) \\
\end{eqnarray*}

              \noindent
              which satisfies condition L1 with $\alpha = 42B$.
}
\item{Observe that any solution $S'$ for for the constructed instance $I'$
       of MGSC of value $k + 2|E'| + 4$ implies a solution $S$ for the 
       given instance $I$ of VC$_B$ of size $k$ in $S$. Hence, it is the 
       case that $m(S) - m(OPT_{VC_B}(I)) = m(S') - m(OPT_{MGSC}(I'))$, 
       which satisfies condition L2 with $\beta = 1$.
}
\end{itemize}

\noindent
As VC$_B$ is $MAX ~ SNP$-hard under L-reductions \cite[Theorem 2(d)]{PY91},
the L-reduction above proves that MGSC is also $MAX ~ SNP$-hard under 
L-reductions. The result follows from Lemma \ref{LemPTAS}.
\end{proof}

\section{Sufficient Circuit Search and Counting Problems}

\begin{definition}
An entity $x$ with property $P$ is {\em minimal} if there is no non-empty
subset of elements in $x$ that can be deleted to create an entity $x'$ with
property $P$.
\end{definition}

\noindent
We shall assume here that all subcircuits are non-trivial, i.e., the
subcircuit is of size $< | M|$.

Consider the following search problem templates:

\vspace*{0.1in}

\noindent
{\sc {\bf Name} local sufficient circuit} ({\bf Acc}LSC) \\
{\em Input}: A multi-layer perceptron $M$ of depth $cd_g$ with $\#n_{tot,g}$ neurons
              and maximum layer width $cw_g$, connection-value matrices $W_1, W_2, \ldots, 
              W_{cd_g}$, neuron bias vector $B$, a Boolean input vector $I$ of length 
              $\#n_{g,in}$, integers $d$ and $w$ such that $1 \leq d \leq 
              cd_g$ and $1 \leq w \leq cw_g$ {\bf PrmAdd}. \\
{\em Output}: A {\bf CType} subcircuit $C$ of $M$ of depth $cd_r \leq d$ 
               with maximum layer width $cw_r \leq w$ that $C(I) = M(I)$,
               if such a subcircuit exists, and special symbol $\bot$ 
               otherwise.

\vspace*{0.1in}

\noindent
{\sc {\bf Name} global sufficient circuit} ({\bf Acc}GSC) \\
{\em Input}: A multi-layer perceptron $M$ of depth $cd_g$ with $\#n_{tot,g}$ neurons
              and maximum layer width $cw_g$, connection-value matrices $W_1, W_2, \ldots, 
              W_{cd_g}$, neuron bias vector $B$, integers $d$ and $w$
	      such that $1 \leq d \leq cd_g$ and $1 \leq w \leq cw_g$ {\bf PrmAdd}. \\
{\em Output}: A {\bf CType} subcircuit $C$ of $M$ of depth $cd_r \leq d$ with
               maximum layer width $cw_r \leq w$ such that $C(I) = M(I)$
               for every possible Boolean input vector $I$ of
               length $\#_{in,g}$, if such a subcircuit exists, and special symbol $\bot$ otherwise..

\vspace*{0.1in}

\noindent
{\sc {\bf Name} local necessary circuit} ({\bf Acc}LNC) \\
{\em Input}: A multi-layer perceptron $M$ of depth $cd$ with $\#n_{tot}$
              neurons and maximum layer width $cw$, connection-value 
              matrices $W_1, W_2, \ldots, W_{cd}$, neuron bias vector $B$,
              a Boolean input vector $I$ of length $\#n_{in}$ {\bf PrmAdd}. \\
{\em Output}: A {\bf CType} subcircuit $C$ of $M$ 
               such that $C \cap C' \neq \emptyset$ for every 
	       sufficient circuit $C'$ of $M$ relative to $I$, if such a s
               subcircuit exists, and special symbol $\bot$ otherwise.

\vspace*{0.1in}

\noindent
{\sc {\bf Name} global necessary circuit} ({\bf Acc}GNC) \\
{\em Input}: A multi-layer perceptron $M$ of depth $cd$ with $\#n_{tot}$
              neurons and maximum layer width $cw$, connection-value 
              matrices $W_1, W_2, \ldots, W_{cd}$, neuron bias vector 
	      $B$ {\bf PrmAdd}. \\
{\em Output}: A {\bf CType} subcircuit $C$ of $M$
               such that for for every possible Boolean input
	       vector $I$ of length $\#n_{in}$, $C \cap C' \neq \emptyset$
	       for every sufficient circuit $C'$ of $M$ relative to $I$,
               if such a subcircuit exists, and special symbol $\bot$
               otherwise.

\vspace*{0.1in}

\noindent
{\sc {\bf Name} local circuit ablation} ({\bf Acc}LCA) \\
{\em Input}: A multi-layer perceptron $M$ of depth $cd$ with $\#n_{tot}$
              neurons and maximum layer width $cw$, connection-value 
              matrices $W_1, W_2, \ldots, W_{cd}$, bias vector $B$,
              a Boolean input vector $I$ of length $\#n_{in}$ {\bf PrmAdd}. \\
{\em Output}: A {\bf CType} subcircuit $C$ of $M$
               such that $(M/C)(I) \neq M(I)$, if such a subcircuit exists,
               and special symbol $\bot$ otherwise.

\vspace*{0.1in}

\noindent
{\sc {\bf Name} global circuit ablation} ({\bf Acc}GCA) \\
{\em Input}: A multi-layer perceptron $M$ of depth $cd$ with $\#n_{tot}$
              neurons and maximum layer width $cw$, connection-value 
              matrices $W_1, W_2, \ldots, W_{cd}$, neuron bias 
	      vector $B$ {\bf PrmAdd}. \\
{\em Output}: A {\bf CType} subcircuit $C$ of $M$ such that $(M/C)(I) \neq
               M(I)$ every possible Boolean input vector $I$ of length 
               $\#n_{in}$, if such a subcircuit exists, and special
               symbol $\bot$ otherwise.

\vspace*{0.1in}

\noindent
Each these templates can be filled out to create six search problem
variants as follows:

\begin{enumerate}
\item{
Exact-size Problem:

\begin{itemize}
\item {\bf Name} $=$ ``Exact'' 
\item {\bf Acc} $=$ ``Ex'' 
\item {\bf PrmAdd} $=$ ``, and a positive integer $k < |M|$'' 
\item{\bf CType} $=$ ``size-$k$''
\end{itemize}
}
\item{
Bounded-size Problem:

\begin{itemize}
\item {\bf Name} $=$ ``Bounded'' 
\item {\bf Acc} $=$ ``B'' 
\item {\bf PrmAdd} $=$ ``, and a positive integer $k < |M|$'' 
\item{\bf CType} $=$ ``size-$\leq k$''
\end{itemize}
}
\item{
Minimal Problem:

\begin{itemize}
\item {\bf Name} $=$ ``Minimal''
\item {\bf Acc} $=$ ``Mnl''
\item {\bf PrmAdd} $=$ ``''
\item{\bf CType} $=$ ``minimal''
\end{itemize}
}
\item{
Minimal Exact-size Problem:

\begin{itemize}
\item {\bf Name} $=$ ``Minimal exact'' 
\item {\bf Acc} $=$ ``MnlEx'' 
\item {\bf PrmAdd} $=$ ``, and a positive integer $k < |M|$'' 
\item{\bf CType} $=$ ``minimal size-$k$''
\end{itemize}
}
\item{
Minimal Bounded-size Problem:

\begin{itemize}
\item {\bf Name} $=$ ``Minimal bounded'' 
\item {\bf Acc} $=$ ``MnlB'' 
\item {\bf PrmAdd} $=$ ``, and a positive integer $k < |M|$'' 
\item{\bf CType} $=$ ``minimal size-$\leq k$''
\end{itemize}
}
\item{
Minimum Problem:

\begin{itemize}
\item {\bf Name} $=$ ``Minimum''
\item {\bf Acc} $=$ ``Min''
\item {\bf PrmAdd} $=$ ``''
\item{\bf CType} $=$ ``minimum''
\end{itemize}
}
\end{enumerate}

\noindent
We will use previous results for the following problems
to prove our results for the problems above.

\vspace*{0.1in}

\noindent
{\sc Exact clique} (ExClique) \\
{\em Input}: An undirected graph $G = (V, E)$ and a positive integer 
              $k \leq |V|$. \\
{\em Output}: A $k$-size vertex subset $V'$ of $G$ that is a clique of
               size $k$ in $G$, if such a $V'$ exists, and special 
               symbol $\bot$ otherwise.

\vspace*{0.1in}

\noindent
{\sc Exact vertex cover} (ExVC) \\
{\em Input}: An undirected graph $G = (V, E)$ and a positive integer 
              $k \leq |V|$. \\
{\em Output}: A $k$-size vertex subset $V'$ of $G$ that is a vertex
               cover of size $k$ in $G$, if such a $V'$ exists, and special 
               symbol $\bot$ otherwise.

\vspace*{0.1in}

\noindent
{\sc Minimal vertex cover} (MnlVC) \citep[Problem 4]{valiantComplexityEnumerationReliability1979} \\
{\em Input}: An undirected graph $G = (V, E)$. \\
{\em Output}: A minimal subset $V'$ of the vertices in $G$ that is a 
               vertex cover of $G$.

\vspace*{0.1in}

\noindent
Given any search problem {\bf X} above, let \#{\bf X} be the problem that 
returns the number of solution outputs. To assess the complexity of
these counting problems, we will use the following definitions in
\citealt[Section 7.3]{garey1979computers}, adapted from those originally given in 
\citealt{valiantComplexityEnumerationReliability1979}.

\begin{definition}
\citep[p. 168]{garey1979computers}
A counting problem \#{$\Pi$} is in \#P if there is a nondeterministic 
algorithm such that for each input $I$ of $\Pi$, (1) the number of `distinct 
``guesses'' that lead to acceptance of $I$ exactly equals the number 
solutions of $\Pi$ for input $I$ and (2) the length of the longest accepting
computation is bounded by a polynomial in $|I|$.
\end{definition}

\noindent
\#P contains very hard problems, as it is known every class in the
Polynomial Hierarchy (which include $P$ and $NP$ as its lowest members)
Turing reduces to \#P, i.e., $PH \subseteq P^{\#P}$ \citep{todaPPHardPolynomialTime1991}. We
use the following type of reduction to isolate problems that are
the hardest (\#P-complete) and at least as hard as the hardest
(\#P-hard) in \#P.

\begin{definition}
\cite[p. 168-169]{garey1979computers}
Given two search problems $\Pi$ and $\Pi'$, a {\em (polynomial time)
parsimonious reduction from $\Pi$ to $\Pi'$} is a function $f : I_{\Pi}
\rightarrow I_{\Pi'}$ that can be computed in polynomial time such that
for every $I \in{\Pi}$, the number of solutions of $\Pi$ for input $I$ 
is exactly equal to the number of $\Pi'$ for input $f(I)$.
\end{definition}

\noindent
We will also derive parameterized counting results using the framework 
given in \citealt[Chapter 14]{flumParameterizedComplexityTheory2006}. The definition of class \#$W[1]$ 
\cite[Definition 14.11]{flumParameterizedComplexityTheory2006} is rather intricate and need not concern us 
here. We will use the following type of reduction to isolate problems that 
are at least as hard as the hardest (\#$W[1]$-hard) in \#$W[1]$.

\begin{definition}
(Adapted from \citealt[Definition 14.10.a]{flumParameterizedComplexityTheory2006})
Given two parameterized search problems $\la k\ra$-$\Pi$ and 
$\la K\ra$-$\Pi'$, a {\em (fpt) parsimonious reduction from 
$\la k \ra$-$\Pi$ to $\la K \ra$-$\Pi'$} is a function $f : I_{\Pi}
\rightarrow I_{\Pi'}$ computable in fixed-parameter time relative to
parameter $k$ such that for every $I \in{\Pi}$ (1) the number of solutions
of $\Pi$ for input $I$ is exactly equal to the number of $\Pi'$ for input 
$f(I)$ and (2) for every parameter $k' \in K$, $k' \leq g_{k'}(k)$ for
some function $g_{k'}()$.
\end{definition}

\noindent
Reductions are often established to be parsimonious by proving bijections 
between solution-sets for $I$ and $f(I)$, i.e., each solution to $I$ 
corresponds to exactly one solution for $f(I)$ and vice versa.

We will prove various parameterized results for our problems using
reductions from ExClique. The
parameterized results are proved relative to the parameters in 
Table \ref{TabPrmCSNC}. Lemmas \ref{LemPrmFPProp1} and \ref{LemPrmFPProp2} will be useful in deriving additional parameterized results from proved ones.



\begin{table}[t]
\caption{Parameters for the sufficient circuit, necessary circuit, circuit 
ablation problems. Note that for sufficient circuit problems, there are
two versions of each parameter $k$ -- namely, those describing the given
MLP $M$ and the derived subcircuit $C$ (distinguished by $g$- ands
$r$-subscripts, respectively).
}
\vspace*{0.1in}
\label{TabPrmCSNC}
\centering
\begin{tabular}{| c || l | }
\hline
Parameter     & Description \\
\hline\hline
$cd$        & \# layers in given MLP \\
\hline
$cw$        & max \# neurons in layer in given MLP \\
\hline
$\#n_{tot}$ & total \# neurons in given MLP \\
\hline
$\#n_{in}$  & \# input neurons in given MLP \\
\hline
$\#n_{out}$ & \# output neurons in given MLP \\
\hline
$B_{\max}$  & max neuron bias in given MLP \\
\hline
$W_{\max}$  & max connection weight in given MLP \\
\hline
$k$           & Size of requested neuron subset \\
\hline
\end{tabular} 
\end{table}

\subsection{Results for Sufficient Circuit Problems}

\begin{theorem}
For $\Pi \in L = \{ {\bf V}LSC, {\bf V}GSC ~ | ~ {\bf V} \in \{Ex, B, MnlEx,
MnlB\}\}$, \linebreak ExClique polynomial-time parsimoniously reduces to
$\Pi$.
\label{ThmPolyParRedSCk}
\end{theorem}
\begin{proof}
Consider first the local sufficient circuit problem variants. Observe
that for the reduction from Clique to MLSC in the proof of Theorem \ref{ThmMLSCCli} (1) the reduction is also from ExClique, (2) each
clique of size $k$ in the given instance of ExClique has exactly
one corresponding sufficient circuit of size $k' = k + k(k -1)/2 + 2$ in the
constructed instance of MLSC and vice versa, and (3) courtesy of the bias in
neuron $n_{out}$ and the structure of the MLP $M$ in the constructed
instance of MLSC, no sufficient circuit can have size $< k'$ and
hence problem variants ExLSC, BLSC, MnlExLSC, and MnlBLSC (when $k' 
= k + k(k - 1)/2 + 2$) have the same set of sufficient circuit solutions
Hence, this reduction is also a polynomial-time parsimonious reduction from
ExClique to each local sufficient circuit problem variant in $L$.

As for the global sufficient circuit problem variants, it was pointed
out that in the proof of Theorem \ref{ThmMGSCCli} that the
reduction above is also a reduction from Clique to MGSC; moreover,
all three properties above also hold modulo MGSC, ExGSC, BGSC, MnlExGSC,
and MnlBGSC. Hence, this reduction is also a polynomial-time parsimonious
reduction from ExClique to each global sufficient circuit problem variant 
in $L$.
\end{proof}

\vspace*{0.1in}

\begin{theorem}
For $\Pi \in L = \{ {\bf V}LSC, {\bf V}GSC ~ | ~ {\bf V} \in \{Ex, B, MnlEx,
MnlB\}\}$, $\la k \ra$-ExClique fpt parsimoniously reduces to $\la cd_g, 
\#n_{in,g}, \#n_{out,g}, B_{\max,g}, W_{\max,g}, cd_r, cw_r,$ \linebreak
$\#n_{in,r}, \#n_{out,r}, \#n_{tot,r}, B_{\max,r}, W_{\max,r}\ra$-$\Pi$.
\label{ThmFPTParRedSCk}
\end{theorem}
\begin{proof}
Observe that in the instance of MLSC constructed in the reduction in the 
proof of Theorem \ref{ThmMLSCCli}, $cd_g = cd_r = 4$, $\#n_{in,g} = 
\#n_{in,r} = \#n_{out,r} = \#n_{out,r} = W_{\max,g} = W_{\max,r} = 1$, and 
$B_{\max,g}, B_{\max,r}, \#n_{tot,r},$ and $cw_r$ are all functions of $k$ 
in the given instance of {\sc Clique}. The result then follows by the
reasoning in the proof of Theorem \ref{ThmPolyParRedSCk}.
\end{proof}

\vspace*{0.1in}

\begin{theorem}
For $\Pi \in L = \{ {\bf V}LSC, {\bf V}GSC ~ | ~ {\bf V} \in \{Ex, B, 
MnlEx, MnlB\}\}$, if $\Pi$ is polynomial-time solvable then $P = NP$.
\label{ThmNonPolySCk}
\end{theorem}
\begin{proof}
Suppose there is a polynomial-time algorithm $A$ for some $\Pi \in L$.
Let $R$ be the polynomial-time algorithm underlying the polynomial-time
parsimonious reduction from ExClique to $\Pi$ specified in the proof
of Theorem \ref{ThmPolyParRedSCk}. Construct an algorithm $A'$ for the
decision version of ExClique as follows: Given an input $I$ for 
ExClique$_D$, create input $I'$ for $\Pi$ using $R$, and apply $A$ to $I'$ 
to create solution $S$. If $S = \bot$, return ``No''; otherwise, return 
``Yes''.  Algorithm $A'$ is a polynomial-time algorithm for ExClique$_D$; 
however, as ExClique$_D$ is $NP$-complete \citep[Problem GT19]{garey1979computers}, this 
implies that $P = NP$, giving the result.
\end{proof}

\vspace*{0.1in}

\begin{theorem}
if MinLSC or MinGSC is polynomial-time solvable then $P = NP$.
\label{ThmNonPolySCM}
\end{theorem}
\begin{proof}
Suppose there is a polynomial-time algorithm $A$ for MinLSC (MinGSC).
Let $R$ be the polynomial-time algorithm underlying the polynomial-time
parsimonious reduction from ExClique to BLSC (BGSC) specified in the proof
of Theorem \ref{ThmPolyParRedSCk}. Construct an algorithm $A'$ for the
decision version of ExClique as follows: Given an input $I$ for 
ExClique$_D$, create input $I'$ for BSC )LSC) using $R$, and apply $A$ to 
$I'$ to create solution $S$. If $|S| \leq k$, return ``Yes''; otherwise, 
return ``No''.  Algorithm $A'$ is a polynomial-time algorithm for 
ExClique$_D$; however, as ExClique$_D$ is $NP$-complete 
\citep[Problem GT19]{garey1979computers}, this implies that $P = NP$, giving the result.
\end{proof}

\vspace*{0.1in}

\begin{theorem}
For $\Pi \in L = \{ {\bf V}LSC, {\bf V}GSC ~ | ~ {\bf V} \in \{Ex, B, MnlEx,
MnlB\}\}$ and $K = \{cd_g, \#n_{in,g}, \#n_{out,g}, B_{\max,g}, W_{\max,g}, 
cd_r, cw_r, \#n_{in,r}, \#n_{out,r}, \#n_{tot,r}, B_{\max,r},$ \linebreak $W_{\max,r}\}$,
if $\la K \ra$-$\Pi$ is fixed-parameter tractable then $FPT = W[1]$.
\label{ThmNonFPTSCk}
\end{theorem}
\begin{proof}
Suppose there is a fixed-parameter tractable algorithm $A$ for 
$\la K \ra$-$\Pi$ for some $\Pi \in L$. Let $R$ be the fixed-parameter algorithm
underlying the fpt parsimonious reduction from $\la k \ra$-ExClique to 
$\la  K \ra$-$\Pi$ specified in the proof of Theorem \ref{ThmFPTParRedSCk}.
Construct an algorithm $A'$ for the decision version of $\la k \ra$-ExClique
as follows: Given an input $I$ for $\la k \ra$-ExClique$_D$, 
create input $I'$ for $\Pi$ using $R$, and apply $A$ to $I'$ to create 
solution $S$. If $S = \bot$, return ``No''; otherwise, return ``Yes''.
Algorithm $A'$ is a fixed-parameter tractable algorithm for 
$\la k \ra$-ExClique$_D$; however, as $\la k \ra$-ExClique$_D$ is 
$W[1]$-complete \citep{downeyParameterizedComplexity1999a}, this implies that $FPT = W[1]$, giving the 
result.
\end{proof}

\vspace*{0.1in}

\begin{theorem}
For $K = \{cd_g, \#n_{in,g}, \#n_{out,g}, B_{\max,g}, W_{\max,g}, 
cd_r, cw_r, \#n_{in,r}, \#n_{out,r},$ \linebreak $\#n_{tot,r}, B_{\max,r}, W_{\max,r}\}$,
if $\la K \ra$-MinLSC or $\la K]\ra$-MinGSC is fixed-parameter tractable 
then $FPT = W[1]$.
\label{ThmNonFPTSCM}
\end{theorem}
\begin{proof}
Suppose there is a fixed-parameter tractable algorithm $A$ for 
$\la K \ra$-MinLSC ($\la K \ra$-MinGSC). Let $R$ be the 
fixed-parameter algorithm underlying the fpt parsimonious reduction from 
$\la k \ra$-ExClique to $\la K\ra$-BLSC ($\la K\ra$-BGSC) specified in the 
proof of Theorem \ref{ThmFPTParRedSCk}. Construct an algorithm $A'$ for the
decision version of $\la k \ra$-ExClique as follows: Given an input $I$ for
$\la k \ra$-ExClique$_D$, create input $I'$ for $\Pi$ using $R$, and apply 
$A$ to $I'$ to create solution $S$. If $|S| \leq k$, return ``Yes''; 
otherwise, return ``No''. Algorithm $A'$ is a fixed-parameter tractable 
algorithm for $\la k \ra$-ExClique$_D$; however, as 
$\la k \ra$-ExClique$_D$ is $W[1]$-complete \citep{downeyParameterizedComplexity1999a}, this implies that 
$FPT = W[1]$, giving the result.
\end{proof}

\vspace*{0.1in}

\begin{theorem}
For $\Pi \in L = \{ {\bf V}LSC ~ | ~ {\bf V} \in \{Ex, B, 
MnlEx, MnlB\}\}$, \#$\Pi$ is \linebreak \#P-complete.
\label{ThmSharpPcSCk}
\end{theorem}
\begin{proof}
As \#ExVC is \#P-complete (\citealt[Page 781]{provanComplexityCountingCuts1983}; see also 
\citealt[Page 169]{garey1979computers}), \#ExClique is
\#P-hard by the polynomial-time parsimonious reduction from
ExVC to ExClique implicit in \citealt[Lemma 3.1]{garey1979computers}. The \#P-hardness
of \#$\Pi$ then follows from the appropriate polynomial-time
parsimonious reduction from ExClique to $\Pi$ specified in the proof
of Theorem \ref{ThmPolyParRedSCk}. Membership of \#$\Pi$ in \#P and the 
result follows from the nondeterministic algorithm for 
\#$\Pi$ that, on each computation path, guesses a subcircuit $C$ of $M$
and then verified that $C$ satisfies the properties required by $\Pi$
relative to MLP $M$ input $I$.
\end{proof}

\vspace*{0.1in}

\begin{theorem}
For $\Pi \in L = \{ {\bf V}GSC ~ | ~ {\bf V} \in \{Ex, B, 
MnlEx, MnlB\}\}$, \#$\Pi$ is \linebreak \#P-hard.
\label{ThmSharpPhSCkC}
\end{theorem}
\begin{proof}
The result follows from the \#P-hardness of \#ExClique noted in the
proof of Theorem \ref{ThmSharpPhSCkC} and the appropriate polynomial-time
parsimonious reduction from ExClique to $\Pi$ specified in the proof
of Theorem \ref{ThmPolyParRedSCk}. 
\end{proof}

\vspace*{0.1in}

\begin{theorem}
\#MnlLSC is \#P-complete.
\label{ThmMnlLSCsPc}
\end{theorem}
\begin{proof}
Consider the reduction from MnlVC to MnlLSC created by modifying the 
reduction
from VC to MLSC given in the proof of Theorem \ref{ThmMLSCVC} such
that the bias and input-line weight of input neuron $n_{in}$ are changed
to 0 and 1 to ensure that all possible input vectors (namely, $\{0\}$ and
$\{1\}$) cause $n_{in}$ to output 1. Observe that this modified reduction
runs in time polynomial in the size of the given instance of MnlVC.
We now need to show that this reduction is parsimonious, i.e., this
reduction creates a bijection between the
solution-sets of the given instance of MnlVC and the constructed instance
of MnlLSC.  We prove the two directions of this bijection separately as 
follows:

\begin{description}
\item [$\Rightarrow$]: Let $V' = \{v'_1, v'_2, \ldots, v'_k\} \subseteq V$ 
       be a minimal vertex cover in $G$. Consider the subcircuit $C$ based 
       on neurons $n_{in}$, $n_{out}$, $\{nv'N ~|~ v' \in V'\}$, 
       $\{neA_1, neA_2, \ldots, neA_{|E|}\}$, and
       $\{neN_1, neN_2, \ldots, neN_{|E|}\}$.  
       As shown in the $\Rightarrow$-portion of the proof of correctness
       of the reduction in the proof of Theorem \ref{ThmMLSCVC},
       $C$ is behaviorally equivalent to $M$ on $I$.
       As $V'$ is minimal and only vertex NOT neurons can be
       deleted from $M$ to create $C$, $C$ must itself be 
       minimal. Moreover, note that any such set $V'$ in $G$ is associated
       with exactly one set of vertex NOT neurons in (and thus exactly
       one sufficient circuit of) $M$.
\item [$\Leftarrow$]: Let $C$ be a minimal subcircuit of $M$ that is 
       behaviorally equivalent to $M$ on input $I$. As neurons in all five 
       layers in $M$ must be present in $C$ to produce the required output,
       both $n_{in}$ and $n_{out}$ are in $C$. In order for $n_{out}$
       to produce a non-zero output, all $|E|$ edge NOT neurons and all
       $|E|$ AND neurons must also be in $C$, and each of the latter must 
       be connected to at least one of the vertex NOT neurons corresponding
       to their endpoint vertices. Hence, the vertices in $G$ corresponding
       to the vertex NOT neurons in $C$ must form a vertex cover in $G$.
       As $C$ is minimal and only vertex NOT neurons can be
       deleted from $M$ to create $C$, this vertex cover must itself be 
       minimal. Moreover, note that any such set of vertex NOT neurons in
       $M$ is associated with exactly one set of vertices in (and hence
       exactly one vertex cover of) $G$.
\end{description}

\noindent
As \#MnlVC is \#P-complete \citep[Theorem 1(4)]{valiantComplexityEnumerationReliability1979}, the reduction 
above establishes that \#MnlLSC is \#P-hard.  Membership of \#MnlLSC in 
\#P and hence the result follows from the nondeterministic algorithm for 
\#MnlLSC that, on each computation path, guesses a subcircuit $C$ of $M$
and then verifies that $C$ is minimal and $C(I) = M(I)$.
\end{proof}

\vspace*{0.1in}

\begin{theorem}
\#MnlGSC is \#P-hard.
\label{ThmMnlGSCsPh}
\end{theorem}
\begin{proof}
Recall that the parsimonious reduction from MnlVC to MnlLSC in the proof of
Theorem \ref{ThmMnlLSCsPc} creates an instance of MnlLSC whose MLP $M$
has the same output for every possible input vector; hence, this reduction
is also a parsimonious reduction from MnlVC to MnlGSC.
As \#MnlVC is \#P-complete \citep[Theorem 1(4)]{valiantComplexityEnumerationReliability1979}, this reduction 
establishes that \#MnlGSC is \#P-hard, giving the result.  
\end{proof}

\vspace*{0.1in}

\begin{theorem}
For $\Pi \in L = \{ {\bf V}LSC, {\bf V}GSC ~ | ~ {\bf V} \in \{Ex, B, 
MnlEx, MnlB\}\}$, if \#$\Pi$ is polynomial-time solvable then $P = NP$.
\label{ThmNonPolySCkC}
\end{theorem}
\begin{proof}
Suppose there is a polynomial-time algorithm $A$ for \#$\Pi$ for
some $\Pi \in L$. Let $R$ be the polynomial-time algorithm underlying the 
polynomial-time parsimonious reduction from ExClique to $\Pi$ specified in 
the proof of Theorem \ref{ThmPolyParRedSCk}. Construct an algorithm $A'$ for
the decision version of ExClique as follows: Given an input $I$ for 
ExClique$_D$, create input $I'$ for $\Pi$ using $R$, and apply $A$ to $I'$ 
to create solution $S$. If $S = 0$, return ``No''; otherwise, return 
``Yes''.  Algorithm $A'$ is a polynomial-time algorithm for ExClique$_D$; 
however, as ExClique$_D$ is $NP$-complete \citep[Problem GT19]{garey1979computers}, this 
implies that $P = NP$, giving the result.
\end{proof}

\vspace*{0.1in}

\begin{theorem}
For $\Pi \in L = \{ {\bf V}LSC, {\bf V}GSC ~ | ~ {\bf V} \in \{Ex, B, MnlEx,
MnlB\}\}$ and $K = \{cd_g, \#n_{in,g}, \#n_{out,g}, B_{\max,g}, W_{\max,g}, 
cd_r, cw_r, \#n_{in,r}, \#n_{out,r}, \#n_{tot,r}, B_{\max,r}, W_{\max,r}\}$,
$\la K \ra$-\#$\Pi$ is \#$W[1]$-hard.
\label{ThmSharpW1hSCkC}
\end{theorem}
\begin{proof}
The result follows from the \#$W[1]$-hardness of $\la k \ra$-\#ExClique
\citep[Theorem 14.18]{flumParameterizedComplexityTheory2006}
and the appropriate fpt parsimonious reduction from $\la k \ra$-ExClique to
$\la K \ra$-$\Pi$ specified in the proof of Theorem \ref{ThmFPTParRedSCk}. 
\end{proof}

\vspace*{0.1in}

\begin{theorem}
For $\Pi \in L = \{ {\bf V}LSC, {\bf V}GSC ~ | ~ {\bf V} \in \{Ex, B, MnlEx,
MnlB\}\}$ and $K = \{cd_g, \#n_{in,g}, \#n_{out,g}, B_{\max,g}, W_{\max,g}, 
cd_r, cw_r, \#n_{in,r}, \#n_{out,r}, \#n_{tot,r}, B_{\max,r}, W_{\max,r}\}$,
if $\la K \ra$-\#$\Pi$ is fixed-parameter tractable then $FPT = \#W[1]$.
\label{ThmNonFPTSCkC}
\end{theorem}
\begin{proof}
Suppose there is a fixed-parameter tractable algorithm $A$ for 
$\la K \ra$-\#$\Pi$ for some $\Pi \in L$. Let $R$ be the fixed-parameter algorithm
underlying the fpt parsimonious reduction from $\la k \ra$-ExClique to 
$\la  K \ra$-$\Pi$ specified in the proof of Theorem \ref{ThmFPTParRedSCk}.
Construct an algorithm $A'$ for $\la k \ra$-\#ExClique
as follows: Given an input $I$ for $\la k ]ra$-\#ExClique, 
create input $I'$ for \#$\Pi$ using $R$, and apply $A$ to $I'$ to create 
solution $S$. If $S = 0$, return ``No''; otherwise, return ``Yes''.
Algorithm $A'$ is a fixed-parameter tractable algorithm for 
$\la k \ra$-\#ExClique$_D$; however, as $\la k \ra$-ExClique$_D$ is 
$\#W[1]$-complete \citep[Theorem 14.18]{flumParameterizedComplexityTheory2006}, this implies that $FPT = 
\#W[1]$, giving the result.
\end{proof}

\vspace*{0.1in}

\section{Global Sufficient Circuit Problem (sigma completeness)}

\noindent
{\sc Minimum global sufficient circuit} (MGSC) \\
{\em Input}: A multi-layer perceptron $M$ of depth $cd$ with $\#n_{tot}$ 
              neurons  and maximum layer width $cw$, connection-value 
	      matrices $W_1, W_2, \ldots, W_{cd_g}$, neuron bias vector $B$,
              and a positive integer $k$. \\
{\em Question}: Is there a subcircuit $C$ of $M$ based on $\leq k$ neurons
              from $M$ such that for every possible input $I$ of $M$,
              $C(I) = M(I)$?

\vspace*{0.1in}

\noindent
Given a subset $N$ of the neurons in $M$, the subcircuit $C$ of $M$ based on
$x$ has the neurons in $x$ and all connections in $M$ among these neurons.
Note that in order for the output of $C$ to be equal to the output of $M$ on
input $I$, the numbers $\#n_{in}$ and $\#n_{out}$ of input and output 
neurons in $M$ must exactly equal the numbers 
of input and output neurons in $C$; hence, no input or output neurons can be
deleted from $M$ in creating $C$. Following \citealt[page 4]{barcelo_model_2020}, all neurons in $M$ use the ReLU activation function and the output $x$ of each output neuron is stepped as necessary to be Boolean, i.e, $step(x) = 0$ if $x \leq 0$ and is $1$ otherwise.

\vspace*{0.1in}

We will prove our result for MGSC using a polynomial-time reduction from
the problem \textsc{Minimum DNF Tautology}. Given a DNF formula $\phi$ over a set $V$ of 
variables, $\phi$ is a {\bf tautology} if $\phi$ evaluates to $True$ for 
every possible truth-assignment to the variables in $V$.

\vspace*{0.1in}



\noindent
Our reduction will use specialized ReLU logic
gates described in \citealt[Lemma 13]{barcelo_model_2020}. These gates assume Boolean
neuron input and output values of 0 and 1 and are structured as follows:

\begin{enumerate}
\item NOT ReLU gate:  A ReLU gate with one input connection weight of value
       $-1$ and a 
       bias of 1. This gate has output 1 if the input is 0 and 0 otherwise.
\item $n$-way AND ReLU gate: A ReLU gate with $n$ input connection weights 
       of value 1 and a bias of $-(n - 1)$. This gate has output 1 if all 
       inputs have value 1 and 0 otherwise.
\item $n$-way OR ReLU gate: A combination of an $n$-way AND ReLU gate with
       NOT ReLU gates on all of its inputs and a NOT ReLU gate on its 
       output that uses DeMorgan's Second Law to implement 
       $(x_1 \vee x_2 \vee \ldots x_n)$ as $\neg(\neg x_1 \wedge \neg x_2 
       \wedge \ldots \neg x_n)$. This gate has 
       output 1 if any input has value 1 and 0 otherwise.
\end{enumerate}

\begin{theorem}
MGSC is $\Sigma^p_2$-complete.
\label{ThmMGSCSp2}
\end{theorem}
\begin{proof}
Let us first show the membership of MGSC in $\Sigma^p_2$. Using the
alternating-quantifier definition of classes in the polynomial hierarchy, 
membership of a decision problem $\Pi$ in $\Sigma^p_2$ can be proved by 
showing that solving $\Pi$ for input $I$ is equivalent to solving a 
quantified formula of the form $\exists (x) \forall (y) : p(x, y)$ 
where both the sizes of $x$ and $y$ and the evaluation time of
predicate formula $p()$ are upper-bounded by polynomials in $|I|$. Such a formula for MGSC is
$$
\exists (\mc{C} \subseteq \mc{M}) \forall \mb{x} \in \{0,1\}^{\#n_{in}} : \mc{C}(\mb{x}) = \mc{M}(\mb{x})
$$

We now show the $\Sigma^p_2$-hardness of MGSC.
Consider the following reduction from 3DT to MGSC. Given an instance $\la 
\phi, T, V, k\ra$ of 3DT, construct the following instance $\la M, k'\ra$
of MGSC: Let $M$ be an MLP based on $3|V| + 2T + 2$ neurons spread
across five layers:

\begin{enumerate}
\item {\bf Input neuron layer}: The input neurons $ni_1, ni_2, \ldots,
       ni_{|V|}$ (all with bias 0).
\item {\bf Hidden layer I}: The unnegated variable identity neurons $nvU_1, 
       nvU_2, \ldots, nvU_{|V|}$ (all with bias 0) and negated variable 
       NOT neurons $nvN_1, nvN_2, \ldots, nvN_{|V|}$ (all with bias 1).
\item{ {\bf Hidden layer II}: 
       \begin{enumerate}
       \item The term 3-way AND neurons $nT_1, xT_2, \ldots, xT_{|T|}$
              (all with bias -2).
       \item The gadget neuron $n_g$ (bias $|V| - 1$).
       \end{enumerate}
}
\item {\bf Hidden layer III}: The modified term 2-way AND neurons
       $nTm_1, nTM_2, \ldots, nTM_{|T|}$ (all with bias $-1$).
\item {\bf Output layer}: The stepped output neuron $n_{out}$.
\end{enumerate}

\noindent
The non-zero weight connections between adjacent layers are as follows:

\begin{itemize}
\item Each input neuron $ni_i$, $1 \leq i \leq |V|$, is input-connected with
       weight 1 to its corresponding input line and output-connected with
       weights 1 and -1 to unnegated and negated variable neurons
       $nvU_i$ and $nvN_i$, respectively.
\item Each term neuron $nT_i$, $1 \leq i \leq |T|$, is input-connected with
       weight 1 to each of the 3 variable neurons corresponding to that
       term's literals.
\item The gadget neuron $n_g$ is input-connected with weight 1 to all of the
       unnegated and negated variable neurons.
\item Each modified term neuron $nTm_i$, $1 \leq i \leq |T|$, is 
       input-connected with weight 1 to the term neuron $nT_i$ and the 
       gadget neuron $n_g$ and output-connected with weight 1 to the
       output neuron $n_{out}$.
\end{itemize}

\noindent
All other connections between neurons in adjacent layers have weight 0.
Finally, let $k' = 3|V|+ 2k + 2$. Observe that this instance of MGSC can be
constructed time polynomial in the size of the given instance of 3DT.

The following observations about the MLP $M$ constructed above will
be of use:

\begin{itemize}
\item The input to $M$ is exactly that of the 3-DNF formula $\phi$.
\item The output neuron of $M$ outputs 1 if and only if one or more of the 
       modified term neurons output 1.
\item Modified term neuron $nTm_i$ outputs 1 if and only both the 
       term neuron $nT_i$ and the gadget neuron $n_g$ output 1.
\item The gadget neuron $n_g$ outputs 1 for input $I$ to a subcircuit $C$
       of $M$ if and only if
       the negated and unnegated variable neurons corresponding to
       $I$ each output 1; hence, $n_g$ outputs 1 for all possible
       inputs to $C$ if and only if all negated and unnegated variable
       neurons in hidden layer I are part of $C$.
\end{itemize}

\noindent
As $\phi$ is a tautology, the above implies that (1) $M$ outputs 1
for every possible input and (2) every global sufficient circuit of $M$
must include all input and variable neurons, the gadget and output neurons,
and at least one term /  modified term neuron-pair.

We now need to show the correctness of this reduction by proving that the 
answer for the given instance of 3DT is ``Yes'' if and only if the answer 
for the constructed instance of MGSC is ``Yes''. We prove the two 
directions of this if and only if separately as follows:

\begin{description}
\item [$\Rightarrow$]: Let $T'$, $|T'| = k$ be a subset of the terms
        in $\phi$ that is a tautology. As noted above, any global
	sufficient circuit $C$ for the constructed MLP $M$ must include all
	input and variable neurons, the gadget and output neurons, and at 
	least one term /  modified term neuron-pair. Let $C$ contain
	the term / modified term neuron-pairs corresponding to the
	terms in $T'$. Such a $C$ is therefore a global sufficient
	circuit for $M$ of size $k' = 3|V| = 2k + 2$.
\item [$\Leftarrow$]: Let $C$ be a global sufficient circuit for $M$
       of size $k' = 3|V| + 2k + 2$. As noted above, $C$ must include all
	input and variable neurons, the gadget and output neurons, and $k$ 
	term /  modified term neuron-pairs. Let $T'$ be the subset of
	$k$ terms in $\phi$ corresponding to these neuron-pairs. Given
	the input-output equivalence $\phi$ and $M$ (and hence any
	sufficient circuit for $M$), the disjunction of the $k$ terms in
	$T'$ must be a tautology.
\end{description}

\noindent
As 3DT is $\Sigma^p_2$-hard \cite[Problem L7]{SU02}, the reduction above 
establishes that MGSC is also $\Sigma^p_2$-hard. The result then
follows from the membership of MGSC in $\Sigma^p_2$ shown at the beginning
of this proof.
\end{proof}

\section{Quasi-Minimal Sufficient Circuit Problem} \noindent
{\sc Quasi-Minimal Sufficient Circuit} (QMSC) \\
{\em Input}: A multi-layer perceptron $M$ of depth $cd$ with $\#n_{tot}$
              neurons and maximum layer width $cw$, connection-value 
              matrices $W_1, W_2, \ldots, W_{cd}$, neuron bias vector $B$, a set $\mc{X}$ of input vectors of length $\#n_{in}$. \\
{\em Output}: a circuit $\mc{C}$ in $\mc{M}$ and a neuron $v \in \mc{C}$ such that $\mc{C}(\mb{x}) = \mc{M}(\mb{x})$ and $[ \mc{C} \setminus \{v\} ](\mb{x}) \neq \mc{M}(\mb{x})$

\vspace*{0.1in}

\begin{theorem}
\label{ThmQMCS}
QMSC is in PTIME (i.e., polynomial-time tractable).
\end{theorem}

\begin{proof}
Consider the following algorithm for QMSC.
Build a sequence of MLPs by taking $\mc{M}$ with all neurons labeled 1, and generating subsequent $\mc{M}_i$ in the sequence by labeling an additional neuron with 0 each time (this choice can be based on any heuristic strategy, for instance, one based on gradients).
The first MLP, $\mc{M}_1$, obtained by removing all neurons labeled 0 (i.e., none) is such that $\mc{M}_1(x) = \mc{M}(x)$, and the last $\mc{M}_n$ is guaranteed to give $\mc{M}_n(x) \neq \mc{M}(x)$ because all neurons are removed.
Label the first MLP \texttt{YES}, and the last \texttt{NO}.
Perform a variant of binary search on the sequence as follows.
Evaluate the $\mc{M}_i$ halfway between \texttt{YES} and \texttt{NO} while removing all its neurons labeled 0.
If it satisfies the condition, label it \texttt{YES}, and repeat the same strategy with the sequence starting from the \texttt{YES} just labeled until the last $\mc{M}_n$.
If it does \textit{not} satisfy the condition, label it \texttt{NO} and repeat the same strategy with the sequence starting from the \texttt{YES} at the beginning of the original sequence until the \texttt{NO} just labeled.
This iterative procedure halves the sequence each time.
Halt when you find two adjacent $\la \texttt{YES}, \texttt{NO} \ra$ circuits (guaranteed to exist), and return the circuit set of the \texttt{YES} network V and the single neuron difference between \texttt{YES} and \texttt{NO} (the breaking point), $v \in V$.
The complexity of this algorithm is roughly $O(n \log n)$.

\end{proof}

\section{Gnostic Neurons Problem}
\noindent
{\sc Gnostic Neurons} (GN) \\
{\em Input}: A multi-layer perceptron $M$ of depth $cd$ with $\#n_{tot}$
              neurons and maximum layer width $cw$, connection-value 
              matrices $W_1, W_2, \ldots, W_{cd}$, neuron bias vector $B$, and two sets $\mc{X}$ and $\mc{Y}$ of input vectors of length $\#n_{in}$, and a positive integer $k$ such that $1 \leq k \leq \#n_{tot}$. \\
{\em Output}: a subset of neurons $V$ in $M$ of size $|V| \geq k$ such that $\forall_{v \in V}$ it is the case that $\forall_{\mb{x} \in \mc{X}}$ computing $M(\mb{x})$ produces activations $A^v_{\mb{x}} \geq t$ and $\forall_{\mb{y} \in \mc{Y}} : A^v_{\mb{y}} < t $.

\vspace*{0.1in}

\begin{theorem}
\label{ThmGN}
GN is in PTIME (i.e., polynomial-time tractable).
\end{theorem}

\begin{proof}
Consider the complexity of the following subroutines of an algorithm for GN.
Computing the activations of all neurons of $M$ for all $\mb{x} \in \mc{X}$ and all $\mb{y} \in \mc{Y}$ takes polynomial time in $|M|, |\mc{X}|$ and $|\mc{Y}|$.
Labeling neurons that pass or not the activation threshold takes time polynomial in $|M|, |\mc{X}|$ and $|\mc{Y}|$.
Finally, checking whether the set of neurons that fulfils the condition is of size at least $k$ can be done in polynomial time in $|M|$.
These subroutines can be put together to yield a polynomial-time algorithm for GN.
\end{proof}

\begin{remark}
    One could also add to the output of the computational problem the requirement that if we silence (or activate) the neuron, we should elicit (or abolish) a behavior. Note that checking these effects can be done in polynomial time in all of the input parts given above and also in the size of the behavior set (which should be added to the input in these variants).
\end{remark}

\section{Necessary Circuit Problem}

\noindent
{\sc Minimum Local Necessary Circuit} (MLNC) \\
{\em Input}: A multi-layer perceptron $M$ of depth $cd$ with $\#n_{tot}$
              neurons and maximum layer width $cw$, connection-value 
              matrices $W_1, W_2, \ldots, W_{cd}$, neuron bias vector $B$,
              a Boolean input vector $I$ of length $\#n_{in}$, and a 
              positive integer $k$ such that $1 \leq k \leq \#n_{tot}$. \\
{\em Question}: Is there a subset $N'$, $|N'| \leq k$, of the $|N|$ neurons
               in $M$ such that $N' \cap C \neq \emptyset$ for every 
	       sufficient circuit $C$ of $M$ relative to $I$?

\vspace*{0.1in}

\noindent
{\sc Minimum Global Necessary Circuit} (MGNC) \\
{\em Input}: A multi-layer perceptron $M$ of depth $cd$ with $\#n_{tot}$
              neurons and maximum layer width $cw$, connection-value 
              matrices $W_1, W_2, \ldots, W_{cd}$, neuron bias vector $B$,
              and a 
              positive integer $k$ such that $1 \leq k \leq \#n_{tot}$. \\
{\em Question}: Is there a subset $N'$, $|N'| \leq k$, of the $|N|$ neurons
               in $M$ such that for for every possible Boolean input
	       vector $I$ of length $\#n_{in}$, $N' \cap C \neq \emptyset$
	       for every sufficient circuit $C$ of $M$ relative 
	       to $I$?

\vspace*{0.1in}

We will use reductions from the Hitting Set problem 
to prove our results for the problems above.

\vspace*{0.1in}



\noindent
Regarding the Hitting Set problem, we shall assume an ordering on the sets and elements in $C$ and $S$,
respectively.

Our parameterized results are proved relative to the parameters in 
Table \ref{TabPrmMNC}. Lemmas \ref{LemPrmFPProp1} and \ref{LemPrmFPProp2} will be useful
in deriving additional parameterized results from proved ones.



\noindent
Our reductions will use specialized ReLU logic
gates described in \citealt[Lemma 13]{barcelo_model_2020}. These gates assume Boolean
neuron input and output values of 0 and 1 and are structured as follows:

\begin{enumerate}
\item NOT ReLU gate:  A ReLU gate with one input connection weight of value
       $-1$ and a 
       bias of 1. This gate has output 1 if the input is 0 and 0 otherwise.
\item $n$-way AND ReLU gate: A ReLU gate with $n$ input connection weights 
       of value 1 and a bias of $-(n - 1)$. This gate has output 1 if all 
       inputs have value 1 and 0 otherwise.
\item $n$-way OR ReLU gate: A combination of an $n$-way AND ReLU gate with
       NOT ReLU gates on all of its inputs and a NOT ReLU gate on its 
       output that uses DeMorgan's Second Law to implement 
       $(x_1 \vee x_2 \vee \ldots x_n)$ as $\neg(\neg x_1 \wedge \neg x_2 
       \wedge \ldots \neg x_n)$. This gate has 
       output 1 if any input has value 1 and 0 otherwise.
\end{enumerate}

\begin{table}[t]
\caption{Parameters for the minimum necessary circuit problem.
}
\vspace*{0.1in}
\label{TabPrmMNC}
\centering
\begin{tabular}{| c || l | }
\hline
Parameter     & Description \\
\hline\hline
$cd$        & \# layers in given MLP \\
\hline
$cw$        & max \# neurons in layer in given MLP \\
\hline
$\#n_{tot}$ & total \# neurons in given MLP \\
\hline
$\#n_{in}$  & \# input neurons in given MLP \\
\hline
$\#n_{out}$ & \# output neurons in given MLP \\
\hline
$B_{\max}$  & max neuron bias in given MLP \\
\hline
$W_{\max}$  & max connection weight in given MLP \\
\hline
$k$           & Size of requested neuron subset \\
\hline
\end{tabular} 
\end{table}

\subsection{Results for MLNC}
Membership of MLNC in $\Sigma^p_2$ can be proven via the definition of the polynomial hierarchy and the following alternating quantifier formula:
$$\exists [N \subseteq \mc{M}] \ \forall [\mc{C} \subseteq \mc{M}] : [\mc{C}(\mb{x}) = \mc{M}(\mb{x})] \implies N \cap \mc{C} \neq \emptyset $$

\begin{theorem}
If MLNC is polynomial-time tractable then $P = NP$.
\label{ThmMLNCHS}
\end{theorem}
\begin{proof}
Consider the following reduction from HS to MLNC. Given an instance $\la C,
S, k\ra$ of HS, construct the following instance $\la M, I, k'\ra$ of MLNC:
Let $M$ be an MLP based on $\#n_{tot} = |S| + |C| + 2$ neurons spread
across four layers:

\begin{enumerate}
\item {\bf Input neuron layer}: The single input neuron $n_{in}$ (bias $+1$).
\item {\bf Hidden element layer}: The element neurons $ns_1, ns_2, \ldots
        ns_{|S|}$ (all with bias 0).
\item {\bf Hidden set layer}: The set AND neurons $nc_1, nc_2, 
        \ldots, nc_{|C|}$ (such that neuron $nc_i$ has bias $-|c_i|$).
\item {\bf Output layer}: The single stepped output neuron $n_{out}$ 
       (bias 0).
\end{enumerate}

\noindent
The non-zero weight connections between adjacent layers are as follows:

\begin{itemize}
\item The input neuron has an edge of weight 0 coming from its
       input and is in turn connected to each of the element
       neurons with weight 1.
\item Each element neuron $ns_i$, $1 \leq i \leq |S|$, is connected to each
       set neuron $nc_j$, $1 \leq j \leq |C|$, such that $s_i \in c_j$ with
       weight 1.
\item Each set neuron $nc_i$, $1 \leq i \leq |C|$, is connected to the
       output neuron with weight 1.
\end{itemize}

\noindent
All other connections between neurons in adjacent layers have weight 0.
Finally, let $I = (0)$ and $k' = k$. Observe that this instance of MLNC can
be created in time polynomial in the size of the given instance of HS. 
Moreover, the output behaviour of the neurons in $M$ from the presentation 
of input $I$ until the output is generated is as follows:

\begin{center}
\begin{tabular}{| c || l |}
\hline
timestep & neurons (outputs) \\
\hline\hline
0 & --- \\
\hline
1 & $n_{in} (1)$ \\
\hline
2 & $ns_1, ns_2, \ldots, ns_{|S|}$ (1) \\
\hline
3 & $nc_1, nc_2, \ldots, nc_{|C|}$ (1) \\
\hline
4 & $n_{out} (1))$ \\
\hline
\end{tabular}
\end{center}
 
\noindent
Note the following about the behavior of $M$:

\begin{observation}
For any set neuron $nc_i$ to output 1, it must receive input 1 from all 
of its incoming element neurons connected with weight 1.
\label{Obs1}
\end{observation}

\begin{observation}
For the output neuron to output 1, it is sufficient to get input 1 from
any of its incoming set neurons with weight 1.
\label{Obs2}
\end{observation}

\noindent
Observations \ref{Obs1} and \ref{Obs2} imply that any sufficient circuit for
$M$ must contain at least one set neuron and all of its associated 
element neurons.

We now need to show the correctness of this reduction by proving that the 
answer for the given instance of HS is ``Yes'' if and only if the answer for
the constructed instance of MLNC is ``Yes''. We prove the two 
directions of this if and only if separately as follows:

\begin{description}
\item [$\Rightarrow$]: Let $S' = \{s'_1, s'_2, \ldots, s'_k\} \subseteq S$ 
       be a hitting set of size $k$ for $C$. By the construction above,
       the $k$ element neurons corresponding to the elements in $S'$
       collectively connect with weight 1 to all set neurons in $M$.
       By Observations \ref{Obs1} and \ref{Obs2},  this means that the
       set $N$ of these element neurons has a non-empty intersection with 
       every sufficient circuit for $M$, and hence that $N$ is a necessary 
       circuit for $M$ of size $k = k'$.

\item [$\Leftarrow$]: Let $N$ be a necessary circuit for $M$ of size $k'$.
       Let $N_S$ and $N_C$ be the subsets of $N$ that are element and 
       set neurons. We can create a set $N'$ consisting only of element
       neurons by replacing each set neuron $n_c$ in $N_C$ with an arbitrary
       element neuron that is not already in $N_S$ and is connected to $n_c$
       with weight 1. Observe that $N'$ (whose size may be less than $k'$ if
       any $n_c$ already had an associated element neuron in $N_S$) remains
       a necessary circuit for $M$. Moreover, as the element neurons in $N'$
       by definition have a non-empty intersection with each sufficient
       circuit for $M$, by Observations \ref{Obs1} and \ref{Obs2} above,
       the set $S'$ of elements in $S$ corresponding to the element neurons
       in $N'$ has a non-empty intersection with each set in $C$ and
       hence is a hitting set of size $N' \leq k' = k$
\end{description}

\noindent
As HS is $NP$-hard \citep{garey1979computers}, the reduction above establishes 
that MLNC is also $NP$-hard.  The result follows from the definition of 
$NP$-hardness.
\end{proof}

\vspace*{0.15in}

\begin{theorem}
If $\la cd, \#n_{in}, \#n_{out}, W_{\max}, k\ra$-MLNC is 
fixed-parameter tractable \linebreak then $FPT = W[1]$.
\label{ThmMLNC_fpi1}
\end{theorem}
\begin{proof}
Observe that in the instance of MLNC constructed in the reduction in the 
proof of Theorem \ref{ThmMLNCHS}, $\#n_{in} = \#n_{out} = W_{\max} = 1$, 
$cd = 4$, and $k'$  is a function of $k$ in the given instance of HS.
The result then follows from the facts that 
$\la k \ra$-HS is $W[2]$-hard (by a reduction from $\la k \ra$-{\sc
Dominating set}; \citealt{downeyParameterizedComplexity1999a}) and $W[1] \subseteq W[2]$.
\end{proof}

\begin{theorem}
$\la \#n_{tot} \ra$-MLNC is fixed-parameter tractable.
\label{ThmMLNC_fpt1}
\end{theorem}
\begin{proof}
Consider the algorithm that generates every possible subset $N'$ of size
at most $k$ of the neurons in MLP $M$ and for each such subset, 
generates every possible subset $N''$ of $M$, checks if $N''$ is
a sufficient circuit for $M$ relative to $I$ and, if so, checks if
$N'$ has a non-empty intersection with $N''$. If an $N'$ is found that
has a non-empty intersection with each sufficient circuit for $M$
relative to $I$, return ``Yes''; otherwise, return ``No''. The number of 
possible subsets $N'$ and $N''$ are both at most $2^{\#n_{tot}}$.
As all subsequent checking operations can be done 
in time polynomial in the size of the given 
instance of MLNC, the above is a fixed-parameter tractable
algorithm for MLNC relative to parameter-set $\{ \#n_{tot} \}$.
\end{proof}

\vspace*{0.15in}

\begin{theorem}
$\la cw, cd \ra$-MLNC is fixed-parameter tractable.
\label{ThmMLNC_fpt2}
\end{theorem}
\begin{proof}
Follows from the algorithm in the proof of Theorem \ref{ThmMLNC_fpt1} and
the observation that $\#n_{tot} \leq cw \times cd$.
\end{proof}

\vspace*{0.15in}

\noindent
Observe that the results in Theorems \ref{ThmMLNC_fpi1}--\ref{ThmMLNC_fpt2}
in combination with Lemmas \ref{LemPrmFPProp1} and \ref{LemPrmFPProp2}
suffice to establish the parameterized complexity status of MLNC relative
to many subsets of the parameters listed in Table \ref{TabPrmMNC}.

Let us now consider the polynomial-time cost approximability of MLNC. 

\begin{theorem}
If MLNC has a polynomial-time $c$-approximation algorithm for any constant 
$c > 0$ then $P = NP$.
\label{ThmMLNC_appi1}
\end{theorem}
\begin{proof}
Recall from the proof of correctness of the reduction in the proof of
Theorem \ref{ThmMLNCHS} that a given instance of HS has a hitting set
of size $k$ if and only if the constructed instance of MLNC has a 
necessary circuit of size $k' = k$. This implies
that, given a polynomial-time $c$-approximation algorithm $A$ for MLNC for 
some constant $c > 0$, we can create a polynomial-time $c$-approximation
algorithm for HS by applying the reduction to the given instance $x$ of HS
to construct an instance $x'$ of MLNC, applying $A$ to $x'$ to create an 
approximate solution $y'$, and then using $y'$ to create an approximate
solution $y$ for $x$ that has the same cost as $y'$. The result then
follows from \citealt[Problem SP7]{ausielloComplexityApproximation1999}, which states that if HS has a 
polynomial-time $c$-approximation algorithm for any constant $c > 0$ a
and is hence in approximation problem class APX then $P = NP$.
\end{proof}

\vspace*{0.15in}

\noindent
Note that this theorem also renders MLNC PTAS-inapproximable unless
$FPT = W[1]$.

\subsection{Results for MGNC}
Membership of MLNC in $\Sigma^p_2$ can be proven via the definition of the polynomial hierarchy and the following alternating quantifier formula:

$$\exists [N \subseteq \mc{M}] \ \forall [\mc{C} \subseteq \mc{M} \ s.t. \ \forall (\mb{x} \in \{0, 1\}^{\#n_{in}})] : N \cap \mc{C} \neq \emptyset $$

\begin{theorem}
If MGNC is polynomial-time tractable then $P = NP$.
\label{ThmMGNCHS}
\end{theorem}
\begin{proof}
Observe that in the instance of MLNC constructed by the reduction in the 
proof of Theorem \ref{ThmMLNCHS}, the input-connection weight 0 and
bias 1 of the input neuron force this neuron to output 1 for
both of the possible input vectors $(1)$ and $(0)$.
Hence, with slight modifications to the proof
of reduction correctness, this reduction also establishes the
$NP$-hardness of MGNC.
\end{proof}

\vspace*{0.15in}

\begin{theorem}
If $\la cd, \#n_{in}, \#n_{out}, W_{\max}, k\ra$-MGNC is 
fixed-parameter tractable then $FPT = W[1]$.
\label{ThmMGNC_fpi1}
\end{theorem}
\begin{proof}
Observe that in the instance of MGNC constructed in the reduction in the 
proof of Theorem \ref{ThmMGNCHS}, $\#n_{in} = \#n_{out} = W_{\max} = 1$, 
$cd = 4$, and $k'$ is a function of $k$ in the given instance 
of HS The result then follows from the facts that 
$\la k \ra$-HS is $W[2]$-hard (by a reduction from $\la k \ra$-{\sc
Dominating set}; \citealt{downeyParameterizedComplexity1999a}) and $W[1] \subseteq W[2]$.
\end{proof}

\begin{theorem}
$\la \#n_{tot} \ra$-MGNC is fixed-parameter tractable.
\label{ThmMGNC_fpt1}
\end{theorem}
\begin{proof}
Modify the algorithm in the proof of Theorem \ref{ThmMLNC_fpt1} such
that each potential sufficient circuit $N''$ is checked to ensure that 
$M(I) = N''(I)$ for
every possible Boolean input vector of length $\#n_{in}$. As the number of
such vectors is $2^{\#n_{in}} < 2^{\#n_{tot}}$, 
the above is a fixed-parameter tractable
algorithm for MGNC relative to parameter-set $\{ \#n_{tot} \}$.
\end{proof}

\vspace*{0.15in}

\begin{theorem}
$\la cw, cd \ra$-MGNC is fixed-parameter tractable.
\label{ThmMGNC_fpt2}
\end{theorem}
\begin{proof}
Follows from the algorithm in the proof of Theorem \ref{ThmMGNC_fpt1} and
the observation that $\#n_{tot} \leq cw \times cd$.
\end{proof}

\vspace*{0.15in}

\noindent
Observe that the results in Theorems \ref{ThmMGNC_fpi1}--\ref{ThmMGNC_fpt2}
in combination with Lemmas \ref{LemPrmFPProp1} and \ref{LemPrmFPProp2}
suffice to establish the parameterized complexity status of MGNC relative
to many subsets of the parameters listed in Table \ref{TabPrmMNC}.

Let us now consider the polynomial-time cost approximability of MGNC. 

\begin{theorem}
If MGNC has a polynomial-time $c$-approximation algorithm for any constant 
$c > 0$ then $P = NP$.
\label{ThmMGNC_appi1}
\end{theorem}
\begin{proof}
As the reduction in the proof of Theorem \ref{ThmMGNCHS} is essentially
the same as the reduction in the proof of Theorem \ref{ThmMLNCHS}, the
result follows by the same reasoning as given in the proof of Theorem
\ref{ThmMLNC_appi1}.
\end{proof}

\vspace*{0.15in}

\noindent
Note that this theorem also renders MGNC PTAS-inapproximable unless
$FPT = W[1]$.


\section{Circuit Ablation and Clamping Problems}

Given an MLP $M$ and a subset $N$ of the neurons in $M$, the MLP $M'$
induced by $N$ is said to be {\em active} if there is at least one path
between the the input and output neurons in $M'$; otherwise, $M'$ is
{\em inactive}. As we are interested in inductions that preserve or
violate output behaviour, all output neurons of $M$ must be preserved in 
$M'$; however, we only require that at least one input neuron be so
preserved. Unless otherwise stated, all inductions discussed wrt MLCA and
MGCA below will be assumed to result in active MLP.

\vspace*{0.1in}

\noindent
{\sc Minimum Local Circuit Ablation} (MLCA) \\
{\em Input}: A multi-layer perceptron $M$ of depth $cd$ with $\#n_{tot}$
              neurons and maximum layer width $cw$, connection-value 
              matrices $W_1, W_2, \ldots, W_{cd}$, neuron bias vector $B$,
              a Boolean input vector $I$ of length $\#n_{in}$, and a 
              positive integer $k$ such that $1 \leq k \leq \#n_{tot}$. \\
{\em Question}: Is there a subset $N'$, $|N'| \leq k$, of the $|N|$ neurons
               in $M$ such that $M(I) \neq M'(I)$ for the MLP $M'$ 
	       induced by $N \setminus N'$?

\vspace*{0.1in}

\noindent
{\sc Minimum Global Circuit Ablation} (MGCA) \\
{\em Input}: A multi-layer perceptron $M$ of depth $cd$ with $\#n_{tot}$
              neurons and maximum layer width $cw$, connection-value 
              matrices $W_1, W_2, \ldots, W_{cd}$, neuron bias vector $B$,
              and a 
              positive integer $k$ such that $1 \leq k \leq \#n_{tot}$. \\
{\em Question}: Is there a subset $N'$, $|N'| \leq k$, of the $|N|$ neurons
               in $M$ such that for the MLP $M'$ induced by $N \setminus 
	       N'$, $M(I) \neq M'(I)$ for every possible Boolean input
	       vector $I$ of length $\#n_{in}$?

\vspace*{0.1in}

\noindent
Given an MLP $M$ and a neuron $v$ in $M$, {\it $v$ is clamped to value
$val$} if the output of $v$ is always $val$ regardless of the inputs to $v$.
As one can trivially change the output of an MLP by clamping one or
more of its output neurons, we shall not allow the clamping of output
neurons in the problems below.

\vspace*{0.1in}

\noindent
{\sc Minimum Local Circuit Clamping} (MLCC) \\
{\em Input}: A multi-layer perceptron $M$ of depth $cd$ with $\#n_{tot}$
              neurons and maximum layer width $cw$, connection-value 
              matrices $W_1, W_2, \ldots, W_{cd}$, neuron bias vector $B$,
              a Boolean input vector $I$ of length $\#n_{in}$, a Boolean
	      value $val$, and a
              positive integer $k$ such that $1 \leq k \leq \#n_{tot}$. \\
{\em Question}: Is there a subset $N'$, $|N'| \leq k$, of the $|N|$ neurons
               in $M$ such that $M(I) \neq M'(I)$ for the MLP $M'$ in 
	       which all neurons in $N'$ are clamped to value $val$?

\vspace*{0.1in}

\noindent
{\sc Minimum Global Circuit Clamping} (MGCC) \\
{\em Input}: A multi-layer perceptron $M$ of depth $cd$ with $\#n_{tot}$
              neurons and maximum layer width $cw$, connection-value 
              matrices $W_1, W_2, \ldots, W_{cd}$, neuron bias vector $B$,
              a Boolean value $val$, and a positive integer $k$ such that 
	      $1 \leq k \leq \#n_{tot}$. \\
{\em Question}: Is there a subset $N'$, $|N'| \leq k$, of the $|N|$ neurons
               in $M$ such that for the MLP $M'$ in which all neurons in 
	       $N'$ are clamped to value $val$, $M(I) \neq M'(I)$ for every
	       possible Boolean input vector $I$ of length $\#n_{in}$?

\vspace*{0.1in}

\noindent
Following \citealt[page 4]{barcelo_model_2020}, all 
neurons in $M$ use the ReLU activation function and the output $x$ of each 
output neuron is stepped as necessary to be Boolean, i.e, $step(x) = 0$ if 
$x \leq 0$ and is $1$ otherwise.





\vspace*{0.1in}

\noindent
For a graph $G = (V, E)$, we shall assume an ordering on the vertices and edges in $V$ 
and $E$, respectively. For each vertex $v \in V$, let the complete neighbourhood $N_C(v)$ of $v$ be the set
composed of $v$ and the set of all vertices in $G$ that are adjacent to $v$
by a single edge, i.e., $v \cup \{ u ~ | ~ u ~ \in V ~ \rm{and} ~ (u,v) \in E\}$.

We will prove various classical and parameterized results for MLCA, MGCA,MLCC, and MGCC using reductions from {\sc Clique}. 
The parameterized results are proved relative to the parameters in Table \ref{TabPrmMCAC}. Lemmas \ref{LemPrmFPProp1} and \ref{LemPrmFPProp2} will be useful in deriving additional parameterized results from proved ones.



\noindent
Additional reductions from DS used to prove polynomial-time cost
inapproximability use specialized ReLU logic
gates described in \citealt[Lemma 13]{barcelo_model_2020}. These gates assume Boolean
neuron input and output values of 0 and 1 and are structured as follows:

\begin{enumerate}
\item NOT ReLU gate:  A ReLU gate with one input connection weight of value
       $-1$ and a 
       bias of 1. This gate has output 1 if the input is 0 and 0 otherwise.
\item $n$-way AND ReLU gate: A ReLU gate with $n$ input connection weights 
       of value 1 and a bias of $-(n - 1)$. This gate has output 1 if all 
       inputs have value 1 and 0 otherwise.
\item $n$-way OR ReLU gate: A combination of an $n$-way AND ReLU gate with
       NOT ReLU gates on all of its inputs and a NOT ReLU gate on its 
       output that uses DeMorgan's Second Law to implement 
       $(x_1 \vee x_2 \vee \ldots x_n)$ as $\neg(\neg x_1 \wedge \neg x_2 
       \wedge \ldots \neg x_n)$. This gate has 
       output 1 if any input has value 1 and 0 otherwise.
\end{enumerate}

\begin{table}[t]
\caption{Parameters for the minimum circuit ablation and clamping problems.
}
\vspace*{0.1in}
\label{TabPrmMCAC}
\centering
\begin{tabular}{| c || l | }
\hline
Parameter     & Description \\
\hline\hline
$cd$        & \# layers in given MLP \\
\hline
$cw$        & max \# neurons in layer in given MLP \\
\hline
$\#n_{tot}$ & total \# neurons in given MLP \\
\hline
$\#n_{in}$  & \# input neurons in given MLP \\
\hline
$\#n_{out}$ & \# output neurons in given MLP \\
\hline
$B_{\max}$  & max neuron bias in given MLP \\
\hline
$W_{\max}$  & max connection weight in given MLP \\
\hline
$k$           & Size of requested neuron subset \\
\hline
\end{tabular} 
\end{table}

\subsection{Results for Minimal Circuit Ablation}

The following hardness and inapproximability results are notable for holding
when the given MLP $M$ has three hidden layers. 

\subsubsection{Results for MLCA}

Towards proving NP-completeness, we first prove membership and then follow up with hardness.
Membership in NP can be proven via the definition of the polynomial hierarchy and the following alternating quantifier formula:

$$\exists [\mc{S} \subseteq \mc{M}] : [\mc{M} \setminus \mc{S}](\mb{x}) \neq \mc{M}(\mb{x})$$

\begin{theorem}
If MLCA is polynomial-time tractable then $P = NP$.
\label{ThmMLCACli}
\end{theorem}
\begin{proof}
Consider the following reduction from {\sc Clique} to MLCA. Given an 
instance $\la G = (V,E), k\ra$ of {\sc Clique}, construct the following 
instance $\la M, I, k'\ra$ of MLCA:
Let $M$ be an MLP based on $\#n_{tot} = 3|V| + |E| + 2$ neurons spread
across five layers:

\begin{enumerate}
\item {\bf Input neuron layer}: The single input neuron $n_{in}$ (bias $+1$).
\item {\bf Hidden vertex pair layer}: The vertex neurons $nvP1_1, nvP1_2, 
        \ldots nvP1_{|V|}$ and $nvP2_1, nvP2_2, \ldots nvP2_{|V|}$
	(all with bias 0).
\item {\bf Hidden vertex regulator layer}: The vertex neurons $nvR_1, nvR_2, \ldots nR_{|V|}$
        (all with bias 0).
\item {\bf Hidden edge layer}: The edge neurons $ne_1, ne_2, \ldots ne_{|E|}$
        (all with bias $-1$).
\item {\bf Output layer}: The single output neuron $n_{out}$ (bias $-(k(k - 1)/2 - 1)$).
\end{enumerate}

\noindent
The non-zero weight connections between adjacent layers are as follows:

\begin{itemize}
\item Each input neuron has an edge of weight 0 coming from its
       corresponding input and is in turn connected to each of the vertex 
       pair neurons with weight 1.
\item Each P1 (P2) vertex pair neuron $nvP1_i$ ($nvP2_i$), $1 \leq i \leq 
       |V|$, is connected to vertex regulator neuron $nvR_i$ with 
       weight $-2$ (1).
\item Each vertex regulator neuron $nv_i$, $1 \leq i \leq |V|$, is connected to each edge neuron 
       whose corresponding edge has an endpoint $v_i$ with weight 1.
\item Each edge neuron $ne_i$, $1 \leq i \leq |E|$, is connected to the output neuron
       $n_{out}$ with weight 1.
\end{itemize}

\noindent
All other connections between neurons in adjacent layers have weight 0.
Finally, let $I = (1)$ and $k' = k$.
Observe that this instance of MLCA can be created in time polynomial in the size of the 
given instance of {\sc Clique}. Moreover, the output behaviour of the neurons in 
$M$ from the presentation of input $I$ until the output is generated is as follows:

\begin{center}
\begin{tabular}{| c || l |}
\hline
timestep & neurons (outputs) \\
\hline\hline
0 & --- \\
\hline
1 & $n_{in} (1)$ \\
\hline
2 & $nvP1_1 (1), nvP1_2 (1), \ldots nvP1_{|V|} (1),
     nvP2_1 (1), nvP2_2 (1), \ldots nvP2_{|V|} (1)$ \\
\hline
3 & $nvR_1 (0), nvR_2 (0), \ldots nvR_{|V|} (0)$ \\
\hline
4 & $ne_1 (0), ne_2 (0), \ldots ne_{|E|} (0)$ \\
\hline
5 & $n_{out} (0))$ \\
\hline
\end{tabular}
\end{center}
 
We now need to show the correctness of this reduction by proving that the answer
for the given instance of {\sc Clique} is ``Yes'' if and only if the answer for the 
constructed instance of MLCA is ``Yes''. We prove the two 
directions of this if and only if separately as follows:

\begin{description}
\item [$\Rightarrow$]: Let $V' = \{v'_1, v'_2, \ldots, v'_k\} \subseteq V$ 
       be a clique in $G$ of size $k'' \geq k$ and $N'$ be the $k'' \geq 
       k' = k$-sized subset of the P1 vertex pair neurons corresponding to 
       the vertices in $V'$. Let $M'$ be the version of $M$ in which all 
       neurons in $N'$ are ablated. As each of these vertex pair neurons
       previously forced their associated vertex regulator
       neurons to output 0 courtesy of their connection-weight of $-2$,
       their ablation now allows these $k''$ vertex regulator neurons to
       output 1.  As $V'$ is a clique of size $k''$, 
       exactly $k''(k'' - 1)/2 \geq k(k - 1)/2$ edge neurons in $M'$ receive
       the requisite inputs of 1 on both of their endpoints from the vertex
       regulator neurons associated with the P1 vertex pair neurons in $N'$.
       This in turn ensures the output neuron produces output 1.  Hence, 
       $M(I) = 0 \neq 1 = M'(I)$.
\item [$\Leftarrow$]: Let $N'$ be a subset of $N$ of size at most $k' = k$
       such that for the MLP $M'$ induced by ablating all neurons in $N'$,
       $M(I) \neq M(I')$. As $M(I) = 0$ and circuit
       outputs are stepped to be Boolean, $M'(I) = 1$. Given the bias of the
       output neuron, this can only occur if at least $k(k - 1)/2$ edge 
       neurons in $M'$ have output 1 on input $I$, which requires that each
       of these neurons receives 1 from both of its endpoint vertex 
       regulator neurons. These vertex regulator neurons can
       only output 1 if all of their associated P1 vertex neurons have
       been ablated; moreover, there must be exactly 
       $k$ such neurons. This means that the vertices in
       $G$ corresponding to the P1 vertex pair neurons in $N'$ must form a
       clique of size $k$ in $G$.
\end{description}

\noindent
As {\sc Clique} is $NP$-hard \citep{garey1979computers}, the reduction above establishes 
that MLCA is also $NP$-hard.  The result follows from the definition of 
$NP$-hardness.
\end{proof}

\vspace*{0.15in}

\begin{theorem}
If $\la cd, \#n_{in}.\#n_{out}, W_{\max}, B_{\max}, k\ra$-MLCA is 
fixed-parameter tractable \linebreak then $FPT = W[1]$.
\label{ThmMLCA_fpi1}
\end{theorem}
\begin{proof}
Observe that in the instance of MLCA constructed in the reduction in the proof of Theorem \ref{ThmMLCACli}, $\#n_{in} = \#n_{out} = W_{\max} = 1$, $cd = 5$, and $B_{\max}$ and $k$ are function of $k$ in the given instance of {\sc Clique}. The result then follows from the fact that $\la k \ra$-{\sc Clique} is $W[1]$-hard \citep{downeyParameterizedComplexity1999a}.
\end{proof}

\begin{theorem}
$\la \#n_{tot} \ra$-MLCA is fixed-parameter tractable.
\label{ThmMLCA_fpt1}
\end{theorem}
\begin{proof}
Consider the algorithm that generates every possible subset $N'$ of size
at most $k$ of the neurons $N$ in MLP $M$ and for each such subset, creates
the MLP $M'$ induced from $M$ by ablating the neurons in $N'$ and
(assuming $M'$ is active) checks if $M'(I) \neq M(I)$. If such a subset is 
found, return ``Yes''; otherwise, return ``No''. The number of possible 
subsets $N'$ is at most $k \times \#n_{tot}^k \leq \#n_{tot} \times 
\#n_{tot}^{\#n_{tot}}$. As any such $M'$ can be generated from $M$, checked
or activity, and run on $I$ in time polynomial in the size of the given 
instance of MLCA, the above is a fixed-parameter tractable
algorithm for MLCA relative to parameter-set $\{ \#n_{tot} \}$.
\end{proof}

\vspace*{0.15in}

\begin{theorem}
$\la cw, cd \ra$-MLCA is fixed-parameter tractable.
\label{ThmMLCA_fpt2}
\end{theorem}
\begin{proof}
Follows from the algorithm in the proof of Theorem \ref{ThmMLCA_fpt1} and
the observation that $\#n_{tot} \leq cw \times cd$.
\end{proof}

\vspace*{0.15in}

\noindent
Observe that the results in Theorems \ref{ThmMLCA_fpi1}--\ref{ThmMLCA_fpt2}
in combination with Lemmas \ref{LemPrmFPProp1} and \ref{LemPrmFPProp2}
suffice to establish the parameterized complexity status of MLCA relative
to many subsets of the parameters listed in Table \ref{TabPrmMCAC}.

Let us now consider the polynomial-time cost approximability of MLCA. As
MLCA is a minimization problem, we cannot do this using reductions from a
maximization problem like {\sc Clique}. Hence we will instead use a
reduction from another minimization problem, namely DS.

\begin{theorem}
If MLCA is polynomial-time tractable then $P = NP$.
\label{ThmMLCADS}
\end{theorem}
\begin{proof}
Consider the following reduction from DS to MLCA.
Given an instance $\la G = (V,E), k\ra$ of DS, construct the following 
instance $\la M, I, k'\ra$ of MLCA:
Let $M$ be an MLP based on $\#n_{tot,g} = 3|V| + 1$ neurons spread
across four layers:

\begin{enumerate}
\item {\bf Input layer}: The input vertex neurons $nv_1, nv_2, \ldots nv_{|V|}$, all of which have bias 1.
\item {\bf Hidden vertex neighbourhood layer I}: The vertex neighbourhood AND neurons $nvnA_1, nvnA_2, \ldots nvnA_{|V|}$, where $nvnA_i$ is an $x$-way AND ReLU gates such that $x = |N_C(v_i)|$.
\item {\bf Hidden vertex neighbourhood layer II}: The vertex neighbourhood NOT neurons $nvnN_1, nvnN_2, \ldots nvnN_{|V|}$, all of which are NOT ReLU gates.
\item {\bf Output layer}: The single output neuron $n_{out}$, which is a
       $|V|$-way AND ReLU gate.
\end{enumerate}

\noindent
The non-zero weight connections between adjacent layers are as follows:

\begin{itemize}
\item Each input vertex neuron $nv_i$, $1 \leq i \leq |V|$, is connected 
       to its input line with weight 0 and to each vertex neighbourhood AND
       neuron $nvnA_j$ such that $v_i \in N_C(v_j)$ with weight 1.
\item Each vertex neighbourhood AND neuron $nvnA_i$, $1 \leq i \leq |V|$, is
       connected to its corresponding vertex neighbourhood NOT neuron 
       $nvnN_i$ with weight 1.
\item Each vertex neighbourhood NOT neuron $nvnN_i$, $1 \leq i \leq |V|$, is
       connected to the output neuron $n_{out}$ with weight 1.
\end{itemize}

\noindent
All other connections between neurons in adjacent layers have weight 0.
Finally, let $I$ be the $|V|$-length one-vector and $k' = k$. Observe that
this instance of MLCA can be created in time polynomial in the size of the
given instance of DS, Moreover, the output behaviour of the neurons in 
$M$ from the presentation of input $I$ until the output is generated is as 
follows:

\begin{center}
\begin{tabular}{| c || l |}
\hline
timestep & neurons (outputs) \\
\hline\hline
0 & --- \\
\hline
1 & $nv_1 (1), nv_2 (1), \ldots nv_{|V|} (1)$ \\
\hline
2 & $nvnA_1 (1), nvnA_2 (1), \ldots nvnA_{|V|} (1)$ \\
\hline
3 & $nvnN_1 (0), nvnN_2 (0), \ldots nvnN_{|V|} (0)$ \\
\hline
4 & $n_{out} (0)$ \\
\hline
\end{tabular}
\end{center}
 
We now need to show the correctness of this reduction by proving that the answer
for the given instance of DS is ``Yes'' if and only if the answer for the 
constructed instance of MLCA is ``Yes''. We prove the two 
directions of this if and only if separately as follows:

\begin{description}
\item [$\Rightarrow$]: Let $V' = \{v'_1, v'_2, \ldots, v'_k\} \subseteq V$ 
       be a dominating set in $G$ of size $k$ and $N'$ be the $k' = 
       k$-sized subset of the input vertex neurons in $M$ corresponding 
       to the vertices in $V'$. Create MLP $M'$ by ablated in $M$ the 
       neurons in $N'$. As $V'$ is
       a dominating set, each vertex neighbourhood AND neuron in $M'$ is
       missing a i-input from at least one input vertex neuron in $N'$,
       which in turn ensures that each vertex neighbourhood AND neuron 
       in $M'$ has output 0. This in turn ensures that $M$ produces output 
       1 on input $I$ such that $M(I) = 0 \neq 1 = M'(I)$. 
\item [$\Leftarrow$]: Let $N'$ be a $k'' \leq k' = k$-sized subset of the 
       set $N$ of neurons in $M$ whose ablation in $M$ creates an MLP $M'$
       such that $M(I) = 0 \neq M'(I)$. As all MLP outputs are stepped
       to be Boolean, this implies that $M'(I) = 1$. This can only happen 
       if all vertex neighbourhood NOT neurons output 1, which in turn can
       happen only if all vertex neighbourhood AND gates output 0. As
       $I = 1^{|V|}$, this can only happen if for each vertex neighbourhood
       AND neuron, at least one input vertex neuron previously producing a 
       1-input to that vertex neighbourhood AND neuron has been ablated in 
       creating $M'$. This in turn implies that the $k''$ vertices in 
       $G$ corresponding to the elements of $N'$ form a dominating set of 
       size $k'' \leq k$ for $G$.
\end{description}

\noindent
As DS is $NP$-hard \citep{garey1979computers}, the reduction above establishes that MLCA is
also $NP$-hard.  The result follows from the definition of $NP$-hardness.
\end{proof}

\vspace*{0.15in}

\begin{theorem}
If MLCA has a polynomial-time $c$-approximation algorithm for any constant $c > 0$ then
$FPT = W[1]$.
\label{ThmMLCA_appi1}
\end{theorem}
\begin{proof}
Recall from the proof of correctness of the reduction in the proof of
Theorem \ref{ThmMLCADS} that a given instance of DS has a dominating set of size $k$ if and only if the constructed instance of MLCA has a subset $N'$ of size $k' = k$ of the neurons in given MLP $M$ such that the ablation in $M$ of the neurons in $N'$ creates an MLP $M'$ such that $M(I) \neq M'(I)$.  
This implies that, given a polynomial-time $c$-approximation algorithm $A$ for MLCA for some constant $c > 0$, we can create a polynomial-time $c$-approximation algorithm for DS by applying the reduction to the given instance $x$ of DS to construct an instance $x'$ of MLCA, applying $A$ to $x'$ to create an approximate solution $y'$, and then using $y'$ to create an approximate solution $y$ for $x$ that has the same cost as $y'$. 
The result then follows from \citealt[Corollary 2]{chenConstantInapproximabilityParameterized2019}, which implies that if DS has a polynomial-time $c$-approximation algorithm for any constant $c > 0$ then $FPT = W[1]$.
\end{proof}

\vspace*{0.15in}

\noindent
Note that this theorem also renders MLCA PTAS-inapproximable unless
$FPT = W[1]$.

\subsubsection{Results for MGCA}

Membership in in $\Sigma^p_2$ can be proven via the definition of the polynomial hierarchy and the following alternating quantifier formula:

$$\exists [\mc{S} \subseteq \mc{M}] \ \forall [\mb{x} \in \{0, 1\}^{\#n_{in}}] : [\mc{M} \setminus \mc{S}](\mb{x}) \neq \mc{M}(\mb{x})$$

\begin{theorem}
If MGCA is polynomial-time tractable then $P = NP$.
\label{ThmMGCACli}
\end{theorem}
\begin{proof}
Observe that in the instance of MLCA constructed by the reduction in the 
proof of Theorem \ref{ThmMLCACli}, the input-connection weight 0 and
bias 1 of the input neuron force this neuron to output 1 for
both of the possible input vectors $(1)$ and $(0)$.
Hence, with slight modifications to the proof
of reduction correctness, this reduction also establishes the
$NP$-hardness of MGCA.
\end{proof}

\vspace*{0.15in}

\begin{theorem}
If $\la cd, \#n_{in}, \#n_{out}, W_{\max}, B_{\max}, k\ra$-MGCA is 
fixed-parameter tractable then $FPT = W[1]$.
\label{ThmMGCA_fpi1}
\end{theorem}
\begin{proof}
Observe that in the instance of MGCA constructed in the reduction in the 
proof of Theorem \ref{ThmMGCACli}, $\#n_{in} = \#n_{out} = W_{\max} = 1$, 
$cd = 5$, and $B_{\max}$ and $k$ are function of $k$ in the given instance 
of {\sc Clique}. The result then follows from the fact that 
$\la k \ra$-{\sc Clique} is $W[1]$-hard \citep{downeyParameterizedComplexity1999a}.
\end{proof}

\begin{theorem}
$\la \#n_{tot} \ra$-MGCA is fixed-parameter tractable.
\label{ThmMGCA_fpt1}
\end{theorem}
\begin{proof}
Modify the algorithm in the proof of Theorem \ref{ThmMLCA_fpt1} such
that each created MLP $M$ is checked to ensure that $M(I) \neq M'(I)$ for
every possible Boolean input vector of length $\#n_{in}$. As the number of
such vectors is $2^{\#n_{in}} \leq 2^{\#n_{tot}}$, 
the above is a fixed-parameter tractable
algorithm for MGCA relative to parameter-set $\{ \#n_{tot} \}$.
\end{proof}

\vspace*{0.15in}

\begin{theorem}
$\la cw, cd \ra$-MGCA is fixed-parameter tractable.
\label{ThmMGCA_fpt2}
\end{theorem}
\begin{proof}
Follows from the algorithm in the proof of Theorem \ref{ThmMGCA_fpt1} and
the observation that $\#n_{tot} \leq cw \times cd$.
\end{proof}

\vspace*{0.15in}

\noindent
Observe that the results in Theorems \ref{ThmMGCA_fpi1}--\ref{ThmMGCA_fpt2}
in combination with Lemmas \ref{LemPrmFPProp1} and \ref{LemPrmFPProp2}
suffice to establish the parameterized complexity status of MGCA relative
to many subsets of the parameters listed in Table \ref{TabPrmMCAC}.

Let us now consider the polynomial-time cost approximability of MGCA. As
MGCA is a minimization problem, we cannot do this using reductions from a
maximization problem like {\sc Clique}. Hence we will instead use a
reduction from another minimization problem, namely DS.

\begin{theorem}
If MGCA is polynomial-time tractable then $P = NP$.
\label{ThmMGCADS}
\end{theorem}
\begin{proof}
Observe that in the instance of MLCA constructed by the reduction in the 
proof of Theorem \ref{ThmMLCADS}, the input-connection weight 0 and
bias 1 of the vertex input neurons force each such neuron to output 1 for
input value 0 or 1.
Hence, with slight modifications to the proof
of reduction correctness, this reduction also establishes the
$NP$-hardness of MGCA.
\end{proof}

\vspace*{0.15in}

\begin{theorem}
If MGCA has a polynomial-time $c$-approximation algorithm for any constant 
$c > 0$ then $FPT = W[1]$.
\label{ThmMGCA_appi1}
\end{theorem}
\begin{proof}
As the reduction in the proof of Theorem \ref{ThmMGCADS} is essentially
the same as the reduction in the proof of Theorem \ref{ThmMLCADS}, the
result follows by the same reasoning as given in the proof of Theorem
\ref{ThmMLCA_appi1}.
\end{proof}

\vspace*{0.15in}

\noindent
Note that this theorem also renders MGCA PTAS-inapproximable unless
$FPT = W[1]$.

\subsection{Results for Minimal Circuit Clamping}

Towards proving NP-completeness, we first prove membership and then follow up with hardness.
Membership in NP can be proven via the definition of the polynomial hierarchy and the following alternating quantifier formula:

$$\exists [\mc{S} \subseteq \mc{M}] : \mc{M}_\mc{S}(\mb{x}) \neq \mc{M}(\mb{x})$$

The following hardness results are notable for holding
when the given MLP $M$ has only one hidden layer. 

\subsubsection{Results for MLCC}

\begin{theorem}
If MLCC is polynomial-time tractable then $P = NP$.
\label{ThmMLCCCli}
\end{theorem}
\begin{proof}
Consider the following reduction from {\sc Clique} to MLCC. Given an 
instance $\la G = (V,E), k\ra$ of {\sc Clique}, construct the following 
instance $\la M, I, val, k'\ra$ of MLCC:
Let $M$ be an MLP based on $\#n_{tot} = |V| + |E| + 1$ neurons spread
across three layers:

\begin{enumerate}
\item {\bf Input vertex layer}: The vertex neurons $nv_1, nv_2, \ldots nv_{|V|}$
        (all with bias $-2$).
\item {\bf Hidden edge layer}: The edge neurons $ne_1, ne_2, \ldots ne_{|E|}$
        (all with bias $-1$).
\item {\bf Output layer}: The single output neuron $n_{out}$ (bias $-(k(k - 1)/2 - 1)$).
\end{enumerate}

\noindent
Note that this MLP has only one hidden layer.
The non-zero weight connections between adjacent layers are as follows:

\begin{itemize}
\item Each vertex neuron $nv_i$, $1 \leq i \leq |V|$, is connected to each edge neuron 
       whose corresponding edge has an endpoint $v_i$ with weight 1.
\item Each edge neuron $ne_i$, $1 \leq i \leq |E|$, is connected to the output neuron
       $n_{out}$ with weight 1.
\end{itemize}

\noindent
All other connections between neurons in adjacent layers have weight 0.
Finally, let $I = 0^{\#n_{in}})$, $val = 1$, and $k' = k$.
Observe that this instance of MLCC can be created in time polynomial in the size of the 
given instance of {\sc Clique}. Moreover, the output behaviour of the neurons in 
$M$ from the presentation of input $I$ until the output is generated is as follows:

\begin{center}
\begin{tabular}{| c || l |}
\hline
timestep & neurons (outputs) \\
\hline\hline
0 & --- \\
\hline
1 & $nv_1 (0), nv_2 (0), \ldots nv_{|V|} (0)$ \\
\hline
2 & $ne_1 (0), ne_2 (0), \ldots ne_{|E|} (0)$ \\
\hline
3 & $n_{out} (0)$ \\
\hline
\end{tabular}
\end{center}
 
We now need to show the correctness of this reduction by proving that the answer
for the given instance of {\sc Clique} is ``Yes'' if and only if the answer for the 
constructed instance of MLCC is ``Yes''. We prove the two 
directions of this if and only if separately as follows:

\begin{description}
\item [$\Rightarrow$]: Let $V' = \{v'_1, v'_2, \ldots, v'_k\} \subseteq V$ 
       be a clique in $G$ of size $k'' \geq k$ and $N'$ be the $k'' \geq 
       k' = k$-sized
       subset of the input vertex neurons corresponding to the vertices in 
       $V'$. Let $M'$ be the version of $M$ in which all neurons in $N'$
       are clamped to value $val = 1$. As $V'$ is a clique of size $k''$, 
       exactly $k''(k'' - 1)/2 \geq k(k - 1)/2$ edge neurons in $M'$ receive
       the requisite inputs of 1 on both of their endpoints from the vertex
       neurons in $N'$. This in turn ensures the output
       neuron produces output 1.  Hence, $M(I) = 0 \neq 1 = M'(I)$.
\item [$\Leftarrow$]: Let $N'$ be a subset of $N$ of size at most $k' = k$
       such that for the MLP $M'$ induced by clamping all neurons in $N'$
       to value $val = 1$, $M(I) \neq M(I')$. As $M(I) = 0$ and circuit
       outputs are stepped to be Boolean, $M'(I) = 1$. Given the bias of the
       output neuron, this can only occur if at least $k(k - 1)/2$ edge 
       neurons in $M'$ have output 1 on input $I$, which requires that each
       of these neurons receives 1 from both of its endpoint vertex neurons.
       As $I = 0^{\#n_{in}}$, these 1-inputs could only have come from
       the clamped vertex neurons in $N'$; moreover, there must be exactly 
       $k$ such neurons. This means that the vertices in
       $G$ corresponding to the vertex neurons in $N'$ must form a
       clique of size $k$ in $G$.
\end{description}

\noindent
As {\sc Clique} is $NP$-hard \citep{garey1979computers}, the reduction above establishes 
that MLCC is also $NP$-hard.  The result follows from the definition of 
$NP$-hardness.
\end{proof}

\vspace*{0.15in}

\begin{theorem}
If $\la cd, \#n_{out}, W_{\max}, B_{\max}, k\ra$-MLCC is fixed-parameter 
tractable then \linebreak $FPT = W[1]$.
\label{ThmMLCC_fpi1}
\end{theorem}

\begin{proof}
Observe that in the instance of MLCC constructed in the reduction in the 
proof of Theorem \ref{ThmMLCCCli}, $\#n_{out} = W_{\max} = 1$, $cd = 3$, and
$B_{\max}$ and $k$ are function of $k$ in the given instance of 
{\sc Clique}. The result then follows from the fact that 
$\la k \ra$-{\sc Clique} is $W[1]$-hard \citep{downeyParameterizedComplexity1999a}.
\end{proof}

\begin{theorem}
$\la \#n_{tot} \ra$-MLCC is fixed-parameter tractable.
\label{ThmMLCC_fpt1}
\end{theorem}
\begin{proof}
Consider the algorithm that generates every possible subset $N'$ of size
at most $k$ of the neurons $N$ in MLP $M$ and for each such subset, creates
the MLP $M'$ induced from $M$ by clamping the neurons in $N'$ to $val$ and 
checks if $M'(I) \neq M(I)$. If such a subset is found, return 
``Yes''; otherwise, return ``No''. The number of possible subsets
$N'$ is at most $k \times \#n_{tot}^k \leq \#n_{tot} \times 
\#n_{tot}^{\#n_{tot}}$. As any such $M'$ can be
generated from $M$ and $M'$ can be run on $I$ in time polynomial in the 
size of the given instance of MLCC, the above is a fixed-parameter tractable
algorithm for MLCC relative to parameter-set $\{ \#n_{tot} \}$.
\end{proof}

\vspace*{0.15in}

\begin{theorem}
$\la cw, cd \ra$-MLCC is fixed-parameter tractable.
\label{ThmMLCC_fpt2}
\end{theorem}
\begin{proof}
Follows from the algorithm in the proof of Theorem \ref{ThmMLCC_fpt1} and
the observation that $\#n_{tot} \leq cw \times cd$.
\end{proof}

\vspace*{0.15in}

\noindent
Observe that the results in Theorems \ref{ThmMLCC_fpi1}--\ref{ThmMLCC_fpt2}
in combination with Lemmas \ref{LemPrmFPProp1} and \ref{LemPrmFPProp2}
suffice to establish the parameterized complexity status of MLCC relative
to many subsets of the parameters listed in Table \ref{TabPrmMCAC}.

Let us now consider the polynomial-time cost approximability of MLCC. As
MLCC is a minimization problem, we cannot do this using reductions from a
maximization problem like {\sc Clique}. Hence we will instead use a
reduction from another minimization problem, namely DS.

\begin{theorem}
If MLCC is polynomial-time tractable then $P = NP$.
\label{ThmMLCCDS}
\end{theorem}
\begin{proof}
Consider the following reduction from DS to MLCC.
Given an instance $\la G = (V,E), k\ra$ of DS, construct the following 
instance $\la M, I, val, k'\ra$ of MLCC:
Let $M$ be an MLP based on $\#n_{tot,g} = 3|V| + 1$ neurons spread
across four layers:

\begin{enumerate}
\item {\bf Input layer}: The input vertex neurons $nv_1, nv_2, \ldots nv_{|V|}$, all of which have bias 1.
\item {\bf Hidden vertex neighbourhood layer I}: The vertex neighbourhood AND neurons $nvnA_1, nvnA_2, \ldots nvnA_{|V|}$, where $nvnA_i$ is an $x$-way AND ReLU gates such that $x = |N_C(v_i)|$.
\item {\bf Hidden vertex neighbourhood layer II}: The vertex neighbourhood NOT neurons $nvnN_1, nvnN_2, \ldots nvnN_{|V|}$, all of which are NOT ReLU gates.
\item {\bf Output layer}: The single output neuron $n_{out}$, which is a
       $|V|$-way AND ReLU gate.
\end{enumerate}

\noindent
The non-zero weight connections between adjacent layers are as follows:

\begin{itemize}
\item Each input vertex neuron $nv_i$, $1 \leq i \leq |V|$, is connected 
       to its input line with weight 0 and to each vertex neighbourhood AND
       neuron $nvnA_j$ such that $v_i \in N_C(v_j)$ with weight 1.
\item Each vertex neighbourhood AND neuron $nvnA_i$, $1 \leq i \leq |V|$, is
       connected to its corresponding vertex neighbourhood NOT neuron 
       $nvnN_i$ with weight 1.
\item Each vertex neighbourhood NOT neuron $nvnN_i$, $1 \leq i \leq |V|$, is
       connected to the output neuron $n_{out}$ with weight 1.
\end{itemize}

\noindent
All other connections between neurons in adjacent layers have weight 0.
Finally, let $I$ be the $|V|$-length one-vector, $val = 0$, and $k' = k$. 
Observe that this instance of MLCC can be created in time polynomial in 
the size of the given instance of DS, Moreover, the output behaviour of the
neurons in $M$ from the presentation of input $I$ until the output is 
generated is as follows:

\begin{center}
\begin{tabular}{| c || l |}
\hline
timestep & neurons (outputs) \\
\hline\hline
0 & --- \\
\hline
1 & $nvN_1 (1), nvN_2 (1), \ldots nvN_{|V|} (1)$ \\
\hline
2 & $nvnA_1 (1), nvnA_2 (1), \ldots nvnA_{|V|} (1)$ \\
\hline
3 & $nvnN_1 (0), nvnN_2 (0), \ldots nvnN_{|V|} (0)$ \\
\hline
4 & $n_{out} (0)$ \\
\hline
\end{tabular}
\end{center}
 
We now need to show the correctness of this reduction by proving that the 
answer for the given instance of DS is ``Yes'' if and only if the answer 
for the constructed instance of MLCC is ``Yes''. We prove the two 
directions of this if and only if separately as follows:

\begin{description}
\item [$\Rightarrow$]: Let $V' = \{v'_1, v'_2, \ldots, v'_k\} \subseteq V$ 
       be a dominating set in $G$ of size $k$ and $N'$ be the $k' = 
       k$-sized subset of the input vertex neurons in $M$ corresponding 
       to the vertices in $V'$. Create MLP $M'$ by clamping in $M$ the 
       neurons in $N'$ to $val = 0$. As $V'$ is
       a dominating set, each vertex neighbourhood AND neuron in $M'$ is
       now missing a i-input from at least one input vertex neuron in $N'$,
       which in turn ensures that each vertex neighbourhood AND neuron 
       in $M'$ has output 0. This in turn ensures that $M$ produces output 
       1 on input $I$ such that $M(I) = 0 \neq 1 = M'(I)$.. 
\item [$\Leftarrow$]: Let $N'$ be a $k'' \leq k' = k$-sized subset of the 
       set $N$ of neurons in $M$ whose clamping to $val = 0$ in $M$ 
       creates an MLP $M'$ such that $M(I) = 0 \neq M'(I)$. As all 
       MLP outputs are stepped to be Boolean, this implies that $M'(I) = 
       1$. This can only happen if
       all vertex neighbourhood NOT neurons output 1, which in turn can
       happen only if all vertex neighbourhood AND gates output 0. As
       $I = 1^{|V|}$, this can only happen if for each vertex neighbourhood
       AND neuron, at least one input vertex neuron previously producing a 
       1-input to that vertex neighbourhood AND neuron has been clamped to
       0 in creating $M'$. This in turn implies that the $k''$ vertices in 
       $G$ corresponding to the elements of $N'$ form a dominating set of 
       size $k'' \leq k$ for $G$.
\end{description}

\noindent
As DS is $NP$-hard \citep{garey1979computers}, the reduction above establishes that MLCC 
is also $NP$-hard.  The result follows from the definition of 
$NP$-hardness.
\end{proof}

\vspace*{0.15in}

\begin{theorem}
If MLCC has a polynomial-time $c$-approximation algorithm for any constant $c > 0$ then
$FPT = W[1]$.
\label{ThmMLCC_appi1}
\end{theorem}
\begin{proof}
Recall from the proof of correctness of the reduction in the proof of
Theorem \ref{ThmMLCCDS} that a given instance of DS has a dominating set
of size $k$ if and only if the constructed instance of MLCC has a subset 
$N'$ of size $k' = k$ of the neurons in given MLP $M$ such that the 
clamping to $val = 0$ in $M$ of the neurons in $N'$ creates an MLP $M'$ 
such that $M(I) \neq M'(I)$.  This implies
that, given a polynomial-time $c$-approximation algorithm $A$ for MLCC for 
some constant $c > 0$, we can create a polynomial-time $c$-approximation
algorithm for DS by applying the reduction to the given instance $x$ of DS
to construct an instance $x'$ of MLCC, applying $A$ to $x'$ to create an 
approximate solution $y'$, and then using $y'$ to create an approximate
solution $y$ for $x$ that has the same cost as $y'$. The result then
follows from \citealt[Corollary 2]{chenConstantInapproximabilityParameterized2019}, which implies that if DS has a 
polynomial-time $c$-approximation algorithm for any constant $c > 0$ then 
$FPT = W[1]$.
\end{proof}

\vspace*{0.15in}

\noindent
Note that this theorem also renders MLCC PTAS-inapproximable unless
$FPT = W[1]$.

\subsubsection{Results for MGCC}

Membership in in $\Sigma^p_2$ can be proven via the definition of the polynomial hierarchy and the following alternating quantifier formula:
$$\exists [\mc{S} \subseteq \mc{M}] \ \forall [\mb{x} \in \{0, 1\}^{\#n_{in}}] : \mc{M}_\mc{S}(\mb{x}) \neq \mc{M}(\mb{x})$$

\begin{theorem}
If MGCC is polynomial-time tractable then $P = NP$.
\label{ThmMGCCCli}
\end{theorem}
\begin{proof}
Observe that in the instance of MLCC constructed by the reduction in the 
proof of Theorem \ref{ThmMLCCCli}, the biases of $-2$ in the input vertex 
neurons force these neurons to map any given Boolean input vector
onto $0^{\#n_{in}}$. Hence, with slight modifications to the proof
of reduction correctness, this reduction also establishes the
$NP$-hardness of MGCC.
\end{proof}

\vspace*{0.15in}

\begin{theorem}
If $\la cd, \#n_{out}, W_{\max}, B_{\max}, k\ra$-MGCC is fixed-parameter 
tractable then \linebreak $FPT = W[1]$.
\label{ThmMGCC_fpi1}
\end{theorem}
\begin{proof}
Observe that in the instance of MGCC constructed in the reduction in the 
proof of Theorem \ref{ThmMGCCCli}, $\#n_{out} = W_{\max} = 1$, $cd = 3$, and
$B_{\max}$ and $k$ are function of $k$ in the given instance of 
{\sc Clique}. The result then follows from the fact that 
$\la k \ra$-{\sc Clique} is $W[1]$-hard \citep{downeyParameterizedComplexity1999a}.
\end{proof}

\begin{theorem}
$\la \#n_{tot} \ra$-MLCC is fixed-parameter tractable.
\label{ThmMGCC_fpt1}
\end{theorem}
\begin{proof}
Modify the algorithm in the proof of Theorem \ref{ThmMLCC_fpt1} such
that each created MLP $M$ is checked to ensure that $M(I) \neq M'(I)$ for
every possible Boolean input vector of length $\#n_{in}$. As the number of
such vectors is $2^{\#n_{in}} \leq 2^{\#n_{tot}}$, 
the above is a fixed-parameter tractable
algorithm for MGCC relative to parameter-set $\{ \#n_{tot} \}$.
\end{proof}

\vspace*{0.15in}

\begin{theorem}
$\la cw, cd \ra$-MGCC is fixed-parameter tractable.
\label{ThmMGCC_fpt2}
\end{theorem}
\begin{proof}
Follows from the algorithm in the proof of Theorem \ref{ThmMGCC_fpt1} and
the observation that $\#n_{tot} \leq cw \times cd$.
\end{proof}

\vspace*{0.15in}

\noindent
Observe that the results in Theorems \ref{ThmMGCC_fpi1}--\ref{ThmMGCC_fpt2}
in combination with Lemmas \ref{LemPrmFPProp1} and \ref{LemPrmFPProp2}
suffice to establish the parameterized complexity status of MGCC relative
to many subsets of the parameters listed in Table \ref{TabPrmMCAC}.

Let us now consider the polynomial-time cost approximability of MGCC. As
MGCC is a minimization problem, we cannot do this using reductions from a
maximization problem like {\sc Clique}. Hence we will instead use a
reduction from another minimization problem, namely DS.

\begin{theorem}
If MGCC is polynomial-time tractable then $P = NP$.
\label{ThmMGCCDS}
\end{theorem}
\begin{proof}
Observe that in the instance of MLCC constructed by the reduction in the 
proof of Theorem \ref{ThmMLCCDS}, the input-connection weight 0 and
bias 1 of the vertex input neurons force each such neuron to output 1 for
input value 0 or 1. Hence, with slight modifications to the proof
of reduction correctness, this reduction also establishes the
$NP$-hardness of MGCC.
\end{proof}

\vspace*{0.15in}

\begin{theorem}
If MGCC has a polynomial-time $c$-approximation algorithm for any constant 
$c > 0$ then $FPT = W[1]$.
\label{ThmMGCC_appi1}
\end{theorem}
\begin{proof}
As the reduction in the proof of Theorem \ref{ThmMGCCDS} is essentially
the same as the reduction in the proof of Theorem \ref{ThmMLCCDS}, the
result follows by the same reasoning as given in the proof of Theorem
\ref{ThmMLCC_appi1}.
\end{proof}

\vspace*{0.15in}

\noindent
Note that this theorem also renders MGCC PTAS-inapproximable unless
$FPT = W[1]$.


\section{Circuit Patching Problem}

\noindent
{\sc Minimum local circuit patching } (MLCP) \\
{\em Input}: A multi-layer perceptron $M$ of depth $cd$ with $\#n_{tot}$
              neurons and maximum layer width $cw$, connection-value 
              matrices $W_1, W_2, \ldots, W_{cd}$, neuron bias vector $B$,
              Boolean input vectors $x$ and $y$ of length $\#n_{in}$, and a 
              positive integer $k$ such that $1 \leq k \leq (\#n_{tot} -
	      (\#n_{in} + \#n_{out}))$. \\
{\em Question}: Is there a subset $C$, $|C| \leq k$, of the internal neurons
              in $M$ such that for the MLP $M'$ created when $M$ is
              $y$-patched wrt $C$, i.e., $M'$ is created when $M/C$ is 
	      patched with activations from $M(x)$ and $C$ is patched with 
	      activations from $M(y)$, $M'(x) = M(y)$?

\vspace*{0.1in}

\noindent
{\sc Minimum Global Circuit Patching } (MGCP) \\
{\em Input}: A multi-layer perceptron $M$ of depth $cd$ with $\#n_{tot}$
              neurons and maximum layer width $cw$, connection-value
              matrices $W_1, W_2, \ldots, W_{cd}$, neuron bias vector $B$,
              Boolean input vector $y$ of length $\#n_{in}$, and a
              positive integer $k$ such that $1 \leq k \leq (\#n_{tot} -
     (\#n_{in} + \#n_{out}))$. \\
{\em Question}: Is there a subset $C$, $|C| \leq k$, of the internal neurons
              in $M$ such that,  for all possible input vectors x, for the
MLP $M'$ created when $M$ is
              $y$-patched wrt $C$, i.e., $M'$ is created when $M/C$ is
     patched with activations from $M(x)$ and $C$ is patched with
     activations from $M(y)$, $M'(x) = M(y)$?

\vspace*{0.1in}

\noindent
Following \citealt[page 4]{barcelo_model_2020}, all 
neurons in $M$ use the ReLU activation function and the output $x$ of each 
output neuron is stepped as necessary to be Boolean, i.e, $step(x) = 0$ if 
$x \leq 0$ and is $1$ otherwise.

\vspace*{0.1in}



\noindent
For a graph $G = (V, E)$, we shall assume an ordering on the vertices and edges in $V$ 
and $E$, respectively. For each vertex $v \in V$, let the complete neighbourhood $N_C(v)$ of $v$ be the set
composed of $v$ and the set of all vertices in $G$ that are adjacent to $v$
by a single edge, i.e., $v \cup \{ u ~ | ~ u ~ \in V ~ \rm{and} ~ (u,v) \in E\}$.

We will prove various classical and parameterized results for MLCP and MGCP
using 
reductions from \textsc{Dominating Set}. The parameterized results are proved relative to the 
parameters in Table \ref{TabPrmMCP}. Our reductions (Theorems 
\ref{ThmMLCPDS} and \ref{ThmMGCPDS}) 
uses specialized ReLU logic
gates described in \citealt[Lemma 13]{barcelo_model_2020}. These gates assume Boolean
neuron input and output values of 0 and 1 and are structured as follows:

\begin{enumerate}
\item NOT ReLU gate:  A ReLU gate with one input connection weight of value
       $-1$ and a 
       bias of 1. This gate has output 1 if the input is 0 and 0 otherwise.
\item $n$-way AND ReLU gate: A ReLU gate with $n$ input connection weights 
       of value 1 and a bias of $-(n - 1)$. This gate has output 1 if all 
       inputs have value 1 and 0 otherwise.
\item $n$-way OR ReLU gate: A combination of an $n$-way AND ReLU gate with
       NOT ReLU gates on all of its inputs and a NOT ReLU gate on its 
       output that uses DeMorgan's Second Law to implement 
       $(x_1 \vee x_2 \vee \ldots x_n)$ as $\neg(\neg x_1 \wedge \neg x_2 
       \wedge \ldots \neg x_n)$. This gate has 
       output 1 if any input has value 1 and 0 otherwise.
\end{enumerate}

\begin{table}[t]
\caption{Parameters for the minimum circuit patching problem.
}
\vspace*{0.1in}
\label{TabPrmMCP}
\centering
\begin{tabular}{| c || l | }
\hline
Parameter     & Description \\
\hline\hline
$cd$        & \# layers in given MLP \\
\hline
$cw$        & max \# neurons in layer in given MLP \\
\hline
$\#n_{tot}$ & total \# neurons in given MLP \\
\hline
$\#n_{in}$  & \# input neurons in given MLP \\
\hline
$\#n_{out}$ & \# output neurons in given MLP \\
\hline
$B_{\max}$  & max neuron bias in given MLP \\
\hline
$W_{\max}$  & max connection weight in given MLP \\
\hline
$k$           & Size of requested patching-subset of $M$ \\
\hline
\end{tabular} 
\end{table}


\subsection{Results for MLCP}
Towards proving NP-completeness, we first prove membership and then follow up with hardness.
Membership in NP can be proven via the definition of the polynomial hierarchy and the following alternating quantifier formula:

$$\exists [\mc{S} \subseteq \mc{M}] : \mc{M}_\mc{S}(\mb{x}) = \mc{M}(\mb{y})$$

\begin{theorem}
If MLCP is polynomial-time tractable then $P = NP$.
\label{ThmMLCPDS}
\end{theorem}
\begin{proof}
Consider the following reduction from DS to MLCP adapted from the reduction from DS to MSR in \Cref{ThmMSRDS}.
Given an instance $\la G = 
(V,E), k\ra$ of DS, construct the following instance $\la M, x, y, k'\ra$ 
of MLCP: Let $M$ be an MLP based on $\#n_{tot} = 4|V| + 1$ neurons spread
across five layers:

\begin{enumerate}
\item {\bf Input layer}: The input vertex neurons $nv_1, nv_2, \ldots nv_{|V|}$, all of which have bias 0.
\item {\bf Hidden vertex layer}: The hidden vertex neurons $nhv_1, nhv_2, \ldots nhv_{|V|}$, all of which are identity ReLU gates with bias 0.
\item {\bf Hidden vertex neighbourhood layer I}: The vertex neighbourhood AND neurons $nvnA_1, nvnA_2, \ldots nvnA_{|V|}$, where $nvnA_i$ is an $x$-way AND ReLU gates such that $x = |N_C(v_i)|$.
\item {\bf Hidden vertex neighbourhood layer II}: The vertex neighbourhood NOT neurons $nvnN_1, nvnN_2, \ldots nvnN_{|V|}$, all of which are NOT ReLU gates.
\item {\bf Output layer}: The single output neuron $n_{out}$, which is a
       $|V|$-way AND ReLU gate.
\end{enumerate}

\noindent
The non-zero weight connections between adjacent layers are as follows:

\begin{itemize}
\item Each input vertex neuron $nv_i$, $1 \leq i \leq |V|$, is connected 
       to its associated hidden vertex neuron $nhv_i$ with weight 1.
\item Each hidden vertex neuron $nhv_i$, $1 \leq i \leq |V|$, is connected 
       to each vertex neighbourhood AND neuron $nvnA_j$ such that $v_i \in 
       N_C(v_j)$ with weight 1.
\item Each vertex neighbourhood AND neuron $nvnA_i$, $1 \leq i \leq |V|$, is
       connected to its corresponding vertex neighbourhood NOT neuron 
       $nvnN_i$ with weight 1.
\item Each vertex neighbourhood NOT neuron $nvnN_i$, $1 \leq i \leq |V|$, is
       connected to the output neuron $n_{out}$ with weight 1.
\end{itemize}

\noindent
All other connections between neurons in adjacent layers have weight 0.
Finally, let $x$ and $y$ be the $|V|$-length one- and zero-vectors and 
$k' = k$. Observe that this instance of MLCP can be created in time 
polynomial in the size of the given instance of DS, Moreover, the output 
behaviour of the neurons in $M$ from the presentation of input $x$ until 
the output is generated is 

\begin{center}
\begin{tabular}{| c || l |}
\hline
timestep & neurons (outputs) \\
\hline\hline
0 & --- \\
\hline
1 & $nv_1 (1), nv_2 (1), \ldots nv_{|V|} (1)$ \\
\hline
2 & $nhv_1 (1), nhv_2 (1), \ldots nhv_{|V|} (1)$ \\
\hline
3 & $nvnA_1 (1), nvnA_2 (1), \ldots nvnA_{|V|} (1)$ \\
\hline
4 & $nvnN_1 (0), nvnN_2 (0), \ldots nvnN_{|V|} (0)$ \\
\hline
5 & $n_{out} (0)$ \\
\hline
\end{tabular}
\end{center}
 
\noindent
and the output behaviour of the neurons in $M$ from the presentation of 
input $y$ until the output is generated is 

\begin{center}
\begin{tabular}{| c || l |}
\hline
timestep & neurons (outputs) \\
\hline\hline
0 & --- \\
\hline
1 & $nv_1 (0), nv_2 (0), \ldots nv_{|V|} (0)$ \\
\hline
2 & $nhv_1 (0), nhv_2 (0), \ldots nhv_{|V|} (0)$ \\
\hline
3 & $nvnA_1 (0), nvnA_2 (0), \ldots nvnA_{|V|} (0)$ \\
\hline
4 & $nvnN_1 (1), nvnN_2 (1), \ldots nvnN_{|V|} (1)$ \\
\hline
5 & $n_{out} (1)$ \\
\hline
\end{tabular}
\end{center}
 
We now need to show the correctness of this reduction by proving that the answer
for the given instance of DS is ``Yes'' if and only if the answer for the 
constructed instance of MLCP is ``Yes''. We prove the two 
directions of this if and only if separately as follows:

\begin{description}
\item [$\Rightarrow$]: Let $V' = \{v'_1, v'_2, \ldots, v'_k\} \subseteq V$ 
       be a dominating set in $G$ of size $k$ and $C$ be the $k' = k$-sized
       subset of the hidden vertex neurons in $M$ corresponding to the 
       vertices in $V'$. As $V'$ is
       a dominating set, each vertex neighbourhood AND neuron receives
       input 0 from at least one hidden vertex neuron in $C$ when
       $M$ is $y$-patched wrt $C$.  This ensures that each 
       vertex neighbourhood AND neuron has output 0, which in turn
       ensures that each vertex neighbourhood NOT neuron has output
       1 and for $M'$ created by $y$-patching $M$ wrt $C$, $M'(x) = M(y) 
       = 1$.
\item [$\Leftarrow$]: Let $C$ be a $k' = k$-sized subset of the internal
       neurons of $M$ such that when $M'$ is created by $y$-patching $M$
       wrt $C$, $M'(x) = M(y) = 1$. The output of $M'$ on $X$ can be 1 (and
       hence equal to the output of $M$ on $y$) only if
       all vertex neighbourhood NOT neurons output 1, which in turn can
       happen only if all vertex neighbourhood AND gates output 0. However,
       as all elements of $x$ have value 1, this means that 
       each vertex neighbourhood AND neuron must be connected to
       at least one patched hidden vertex neuron (all of which have 
       output 0 courtesy of $y$), which in turn implies that the
       $k' = k$ vertices in $G$ corresponding to the patched hidden
       vertex neurons in $C$ form a dominating set of size $k$ for $G$.
\end{description}

\noindent
As DS is $NP$-hard \citep{garey1979computers}, the reduction above establishes that MLCP is
also $NP$-hard.  The result follows from the definition of $NP$-hardness.
\end{proof}

\begin{theorem}
If $\la cd, \#n_{out}, W_{\max}, B_{\max}, k\ra$-MLCP is fixed-parameter 
tractable then \linebreak $FPT = W[1]$.
\label{ThmMLCP_fpi1}
\end{theorem}
\begin{proof}
Observe that in the instance of MLCP constructed in the reduction in the 
proof of Theorem \ref{ThmMLCPDS}, $\#n_{out} = W_{\max} = 1$, $cd = 4$, and 
$B_{\max}$ and $k$ are function of $k$ in the given instance of 
DS. The result then follows from the facts that 
$\la k \ra$-DS is $W[2]$-hard \citep{downeyParameterizedComplexity1999a} and $W[1] \subseteq W[2]$.
\end{proof}

\begin{theorem}
$\la \#n_{tot} \ra$-MLCP is fixed-parameter tractable.
\label{ThmMLCP_fpt1}
\end{theorem}
\begin{proof}
Consider the algorithm that generates all possible subset $C$ of the
internal neurons in $M$ and for each such subset, checks if $M'$ created
by $y$-patching $M$ wrt $C$ is such that $M'(x) = M(y)$.  If such a $C$ is 
found, return ``Yes''; otherwise, return ``No''. The number of possible 
subsets $C$ is at most $2^{(\#n_{tot}}$.  Given this, as $M$ can be patched
relative to $C$ and run on $x$ and $y$ in time polynomial in the size of the
given instance of MLCP, the above is a fixed-parameter tractable algorithm 
for MLCP
relative to parameter-set $\{ \#n_{tot} \}$.
\end{proof}

\vspace*{0.15in}




\noindent
Observe that the results in Theorems \ref{ThmMLCP_fpi1} and \ref{ThmMLCP_fpt1}
in combination with Lemmas \ref{LemPrmFPProp1} and \ref{LemPrmFPProp2}
suffice to establish the parameterized complexity status of MLCP relative
to many subset of the parameters listed in Table \ref{TabPrmMCP}.

Let us now consider the polynomial-time cost approximability of MLCP. 

\begin{theorem}
If MLCP has a polynomial-time $c$-approximation algorithm for any constant $c > 0$ then
$FPT = W[1]$.
\label{ThmMLCP_appi1}
\end{theorem}
\begin{proof}
Recall from the proof of correctness of the reduction in the proof of
Theorem \ref{ThmMLCPDS} that a given instance of DS has a dominating set
of size $k$ if and only if the constructed instance of $MLCP$ has a subset 
$C$ of the internal neurons in $M$ of size $k' = k$ such that for the MLP 
$M'$ created from $M$ by $y$-patching $M$ wrt $C$, $M'(x) = M(y)$ This 
implies that, given a polynomial-time $c$-approximation algorithm $A$ for 
MLCP for some constant $c > 0$, we can create a polynomial-time 
$c$-approximation algorithm for DS by applying the reduction to the given 
instance $I$ of DS to construct an instance $I'$ of MLCP, applying $A$ to 
$I'$ to create an approximate solution $S'$, and then using $S'$ to create 
an approximate solution $S$ for $I$ that has the same cost as $S'$. The 
result then follows from \citealt[Corollary 2]{chenConstantInapproximabilityParameterized2019}, which implies that if 
DS has a polynomial-time $c$-approximation algorithm for any constant 
$c > 0$ then $FPT = W[1]$.
\end{proof}

\vspace*{0.15in}

\noindent
Note that this theorem also renders MLCP PTAS-inapproximable unless
$FPT = W[1]$.

\subsection{Results for MGCP}

Membership in in $\Sigma^p_2$ can be proven via the definition of the polynomial hierarchy and the following alternating quantifier formula:
$$\exists [\mc{S} \subseteq \mc{M}] \ \forall [\mb{x} \in \{0, 1\}^{\#n_{in}}] : \mc{M}_\mc{S}(\mb{x}) = \mc{M}(\mb{y})$$

\begin{theorem}
If MGCP is polynomial-time tractable then $P = NP$.
\label{ThmMGCPDS}
\end{theorem}
\begin{proof}
Modify the reduction in the proof of Theorem \ref{ThmMLCPDS} such that
each hidden vertex neuron has input weight 0 and bias 1; this will force
all hidden vertex neurons to output 1 for all input vectors $x$ instead of
just when $x$ is the all-one vector. 
Hence, with slight modifications to the proof
of reduction correctness for the reduction in the proof of Theorem
\ref{ThmMLCPDS}, this modified reduction establishes the
$NP$-hardness of MGCP.
\end{proof}

\begin{theorem}
If $\la cd, \#n_{out}, W_{\max}, B_{\max}, k\ra$-MGCP is fixed-parameter 
tractable then \linebreak $FPT = W[1]$.
\label{ThmMGCP_fpi1}
\end{theorem}
\begin{proof}
Observe that in the instance of MGCP constructed in the reduction in the 
proof of Theorem \ref{ThmMGCPDS}, $\#n_{out} = W_{\max} = 1$, $cd = 4$, and 
$B_{\max}$ and $k$ are function of $k$ in the given instance of 
DS. The result then follows from the facts that 
$\la k \ra$-DS is $W[2]$-hard \citep{downeyParameterizedComplexity1999a} and $W[1] \subseteq W[2]$.
\end{proof}

\begin{theorem}
$\la \#n_{tot} \ra$-MGCP is fixed-parameter tractable.
\label{ThmMGCP_fpt1}
\end{theorem}
\begin{proof}
Modify the algorithm in the proof of Theorem \ref{ThmMLCP_fpt1} such that 
each circuit $M'$ created by $y$-patching $M$ is checked to ensure that 
$M'(x) = M(y)$ for every possible Boolean input vector $x$ of length 
$\#n_{in}$. As the number of such vectors is $2^{\#n_{in}} < 2^{\#n_{tot}}$,
the above is a fixed-parameter tractable
algorithm for MGCP relative to parameter-set $\{ \#n_{tot} \}$.
\end{proof}

\vspace*{0.15in}

\noindent
Observe that the results in Theorems \ref{ThmMGCP_fpi1} and \ref{ThmMGCP_fpt1}
in combination with Lemmas \ref{LemPrmFPProp1} and \ref{LemPrmFPProp2}
suffice to establish the parameterized complexity status of MGCP relative
to many subset of the parameters listed in Table \ref{TabPrmMCP}.

Let us now consider the polynomial-time cost approximability of MGCP. 

\begin{theorem}
If MGCP has a polynomial-time $c$-approximation algorithm for any constant $c > 0$ then
$FPT = W[1]$.
\label{ThmMGCP_appi1}
\end{theorem}
\begin{proof}
As the reduction in the proof of Theorem \ref{ThmMGCPDS} is essentially
the same as the reduction in the proof of Theorem \ref{ThmMLCPDS}, the
result follows by the same reasoning as given in the proof of Theorem
\ref{ThmMLCP_appi1}.
\end{proof}

\vspace*{0.15in}

\noindent
Note that this theorem also renders MGCP PTAS-inapproximable unless
$FPT = W[1]$.


\section{Quasi-Minimal Circuit Patching Problem}
\noindent
{\sc Quasi-Minimal Circuit Patching} (QMCP) \\
{\em Input}: A multi-layer perceptron $M$ of depth $cd$ with $\#n_{tot}$
              neurons and maximum layer width $cw$, connection-value 
              matrices $W_1, W_2, \ldots, W_{cd}$, neuron bias vector $B$, an input vector $\mb{y}$ and a set $\mc{X}$ of input vectors of length $\#n_{in}$. \\
{\em Output}: a subset $\mc{C}$ in $\mc{M}$ and a neuron $v \in \mc{C}$, such that for the $\mc{M}^*$ induced by patching $\mc{C}$ with activations from $\mc{M}(\mb{y})$ and $\mc{M} \setminus \mc{C}$ with activations from $\mc{M}(\mb{x})$, $\forall_{\mb{x} \in \mc{X}}: \mc{M}^*(\mb{x}) = \mc{M}(\mb{y})$, and for $\mc{M}'$ induced by patching identically except for $v \in \mc{C}$, $\exists_{\mb{x} \in \mc{X}}: \mc{M}'(\mb{x}) \neq \mc{M}(\mb{y})$.

\vspace*{0.1in}

\begin{theorem}
\label{ThmQMCP}
QMCP is in PTIME (i.e., polynomial-time tractable).
\end{theorem}

\begin{proof}
Consider the following algorithm for QMCP.
Build a sequence of MLPs by taking $\mc{M}$ with all neurons labeled 0, and generating subsequent $\mc{M}_i$ in the sequence by labeling an additional neuron with 1 each time (this choice can be based on any heuristic strategy, for instance, one based on gradients).
The first MLP, $\mc{M}_1$, obtained by patching all neurons labeled 1 (i.e., none) is such that $\mc{M}_1(x) \neq \mc{M}(y)$, and the last $\mc{M}_n$ is guaranteed to give $\mc{M}_n(x) = \mc{M}(y)$ because all neurons are patched.
Label the first MLP \texttt{NO}, and the last \texttt{YES}.
Perform a variant of binary search on the sequence as follows.
Evaluate the $\mc{M_i}$ halfway between \texttt{NO} and \texttt{YES} while patching all its neurons labeled 1.
If it satisfies the condition, label it \texttt{YES}, and repeat the same strategy with the sequence starting from the first $\mc{M}_i$ until the \texttt{YES} just labeled.
If it does \textit{not} satisfy the condition, label it \texttt{NO} and repeat the same strategy with the sequence starting from the \texttt{NO} just labeled until the \texttt{YES} at the end of the original sequence.
This iterative procedure halves the sequence each time.
Halt when you find two adjacent $\la \texttt{NO}, \texttt{YES} \ra$ patched networks (guaranteed to exist), and return the patched neuron set of the \texttt{YES} network V and the single neuron difference between \texttt{YES} and \texttt{NO} (the breaking point), $v \in V$.
The complexity of this algorithm is roughly $O(n \log n)$.

\end{proof}

\section{Circuit Robustness Problem}

\begin{definition}
Given an MLP $M$, a subset $H$ of the elements in $M$, an integer $k 
\leq |H|$, and an input $I$ to $M$, $M$ is $k$-robust relative to 
$H$ for $I$ if for each subset $H' \subseteq H$, $|H'| \leq k$, 
$M(I) = (M/H')(I)$.
\end{definition}

\vspace*{0.1in}

\noindent
{\sc Maximum Local Circuit Robustness} (MLCR) \\
{\em Input}: A multi-layer perceptron $M$ of depth $cd$ with $\#n_{tot}$
              neurons and maximum layer width $cw$, connection-value 
              matrices $W_1, W_2, \ldots, W_{cd}$, neuron bias vector $B$,
	      a subset $H$ of the neurons in $M$,
              a Boolean input vector $I$ of length $\#n_{in}$, and a 
              positive integer $k$ such that $1 \leq k \leq |H|$. \\
{\em Question}: Is $M$ $k$-robust relative to $H$ for $I$?

\vspace*{0.1in}

\noindent
{\sc Restricted Maximum Local Circuit Robustness} (MLCR$^*$) \\
{\em Input}: A multi-layer perceptron $M$ of depth $cd$ with $\#n_{tot}$
              neurons and maximum layer width $cw$, connection-value 
              matrices $W_1, W_2, \ldots, W_{cd}$, neuron bias vector $B$,
              a Boolean input vector $I$ of length $\#n_{in}$, and a 
              positive integer $k$ such that $1 \leq k \leq |M|$. \\
{\em Question}: Is $M$ $k$-robust relative to $H = M$ for $I$?

\vspace*{0.1in}

\noindent
{\sc Maximum Global Circuit Robustness} (MGCR) \\
{\em Input}: A multi-layer perceptron $M$ of depth $cd$ with $\#n_{tot}$
              neurons and maximum layer width $cw$, connection-value 
              matrices $W_1, W_2, \ldots, W_{cd}$, neuron bias vector $B$,
	      a subset $H$ of the neurons in $M$, and a 
              positive integer $k$ such that $1 \leq k \leq |H|$. \\
{\em Question}: Is $M$ $k$-robust relative to $H$ for every possible
              Boolean input vector $I$ of length $\#n_{in}$?

\vspace*{0.1in}

\noindent
{\sc Restricted Maximum Global Circuit Robustness} (MGCR$^*$) \\
{\em Input}: A multi-layer perceptron $M$ of depth $cd$ with $\#n_{tot}$
              neurons and maximum layer width $cw$, connection-value 
              matrices $W_1, W_2, \ldots, W_{cd}$, neuron bias vector $B$,
              and a positive integer $k$ such that $1 \leq k \leq |M|$. \\
{\em Question}: Is $M$ $k$-robust relative to $H = M$ for every possible
              Boolean input vector $I$ of length $\#n_{in}$?

\vspace*{0.1in}

\noindent
Following \citealt[page 4]{barcelo_model_2020}, all 
neurons in $M$ use the ReLU activation function and the output $x$ of each 
output neuron is stepped as necessary to be Boolean, i.e, $step(x) = 0$ if 
$x \leq 0$ and is $1$ otherwise.

We will use previous results for \textsc{Minimum Local/Global Circuit Ablation}, \textsc{Clique} and \textsc{Vertex Cover} to prove our
results for the problems above.

\vspace*{0.1in}

\noindent
For a graph $G = (V, E)$, we shall assume an ordering on the vertices and edges in $V$ 
and $E$, respectively. 

We will prove various classical and parameterized results for MLCR, 
MLCR$^*$, MGCR, and MGCR$^*$. The
parameterized results are proved relative to the parameters in 
Table \ref{TabPrmMCR}. Lemmas \ref{LemPrmFPProp1} and \ref{LemPrmFPProp2} will be useful in deriving additional parameterized results from proved ones.



\begin{table}[t]
\caption{Parameters for the minimum circuit robustness problem.
}
\vspace*{0.1in}
\label{TabPrmMCR}
\centering
\begin{tabular}{| c || l | c | }
\hline
Parameter     & Description & Appl. \\
\hline\hline
$cd$        & \# layers in given MLP & All \\
\hline
$cw$        & max \# neurons in layer in given MLP & All \\
\hline
$\#n_{tot}$ & total \# neurons in given MLP & All \\
\hline
$\#n_{in}$  & \# input neurons in given MLP & All \\
\hline
$\#n_{out}$ & \# output neurons in given MLP & All \\
\hline
$B_{\max}$  & max neuron bias in given MLP & All \\
\hline
$W_{\max}$  & max connection weight in given MLP & All \\
\hline\hline
$k$           & Requested level of robustness & All \\
\hline
$|H|$           & Size of investigated region & M$\{$L,G$\}$CR \\
\hline
\end{tabular} 
\end{table}

\noindent
Several of our proofs involve
problems that are hard for $coW[1]$ and $coNP$, the complement
classes of $W[1]$ and $NP$ \citep[Section 7.1]{garey1979computers}. Given a decision
problem {\bf X} let co-{\bf X} be the complement problem in which all
instances answers are switched. 

\begin{observation}
MLCR$^*$ is the complement of MLCA. 
\label{ObsMLCR*MLCA}
\end{observation}

\begin{proof}
Let $X = \la M, I, k\ra$ be a shared input of MLCR$^*$ and MLCA. If the
the answer to MLCA on input $X$ is ``Yes'', this means that there is a
subset $H$ of size $\leq k$ of the neurons of $M$ that can be ablated
such that $M(I) \neq (M/H)(I)$. This implies that the answer to MLCR$^*$ 
on input $X$ is ``No''. Conversely, if the answer to MLCA on input $X$ 
is ``No'', this means that there is no subset $H$ of size $\leq k$ of the 
neurons of $M$ that can be ablated such that $M(I) \neq (M/H)(I)$. This 
implies that the answer to MLCR$^*$ on input $X$ is ``Yes'. The
observation follows from the definition of complement problem.
\end{proof}

\vspace*{0.15in}

\noindent
The following lemmas will be of use in
the derivation and interpretation of results involving complement problems
and classes.

\begin{lemma}
\citep[Section 7.1]{garey1979computers}
Given a decision problem {\bf X}, if {\bf X} is $NP$-hard then
co-{\bf X} is $coNP$-hard.
\label{LemcoNPI}
\end{lemma}

\begin{lemma}
Given a decision problem {\bf X}, if {\bf X} is $coNP$-hard and 
{\bf X} is polynomial-time solvable then $P = NP$.
\label{LemcoNPII}
\end{lemma}

\begin{proof}
Suppose decision problem {\bf X} is $coNP$-hard and solvable in
polynomial-time by algorithm $A$. By the definition
of problem, class hardness, for every problem {\bf Y} in $coNP$ there is
a polynomial-time many-one reduction $\Pi$ from {\bf Y} to {\bf X}; let
$A_{\Pi}$ be the polynomial-time algorithm encoded in $\Pi$. We
can create a polynomial-time algorithm $A'$ for co-{\bf Y} by
running $A_{\Pi}$ on a given input, running $A$, and then complementing
the produced output, i.e., ``Yes'' $\Rightarrow$ ``No'' and ``No''
$\Rightarrow$ ``Yes''. However, as co{\bf X} is $NP$-hard by Lemma
\ref{LemcoNPI}, this implies that $P = NP$.
\end{proof}

\begin{lemma}
\citep[Lemma 8.23]{flumParameterizedComplexityTheory2006}
Let $\cal{C}$ be a parameterized complexity class. Given a parameterized 
decision problem {\bf X}, if {\bf X} is $\cal{C}$-hard then
co-{\bf X} is $co\cal{C}$-hard.
\label{LemcoWI}
\end{lemma}

\begin{lemma}
Given a parameterized decision problem {\bf X}, if {\bf X} is 
fixed-parameter tractable then co-{\bf X} is fixed-parameter tractable.
\label{LemcoWIa}
\end{lemma}

\begin{proof}
Given a fixed-parameter tractable algorithm $A$ for {\bf X}, we
can create a fixed-parameter tractable algorithm $A'$ for co-{\bf X} by
running $A$ on a given input and then complementing
the produced output, i.e., ``Yes'' $\Rightarrow$ ``No'' and ``No''
$\Rightarrow$ ``Yes''. 
\end{proof}

\begin{lemma}
Given a parameterized decision problem {\bf X}, if {\bf X} is 
$coW[1]$-hard and fixed-parameter tractable then $FPT = W[1]$.
\label{LemcoWII}
\end{lemma}

\begin{proof}
Suppose decision problem {\bf X} is $coW[1]$-hard and solvable in
polynomial-time by algorithm $A$. By the definition of problem class 
hardness, for every problem {\bf Y} in $coW[1]$ there is
a parameterized reduction $\Pi$ from {\bf Y} to {\bf X}; let
$A_{\Pi}$ be the fixed-parameter tractable algorithm encoded in $\Pi$. We
can create a polynomial-time algorithm $A'$ for co-{\bf Y} by
running $A_{\Pi}$ on a given input, running $A$, and then complementing
the produced output, i.e., ``Yes'' $\Rightarrow$ ``No'' and ``No''
$\Rightarrow$ ``Yes''. However, as co{\bf X} is $W[1]$-hard by Lemma
\ref{LemcoWI}, this implies that $P = NP$.
\end{proof}

\vspace*{0.1in}

We will also be deriving polynomial-time inapproximability results for
optimization versions of MLCR, MLCR$^*$, MGCR, and MGCR$^*$, i.e.,

\begin{itemize}
\item Max-MLCR, which asks for the maximum value $k$ such that $M$ is
       $k$-robust relative to $H$ for $I$.
\item Max-MLCR$^*$, which asks for the maximum value $k$ such that $M$ is
       $k$-robust relative to $H = M$ for $I$.
\item Max-MGCR, which asks for the maximum value $k$ such that $M$ is
       $k$-robust relative to $H$ for every possible Boolean input vector 
       $I$ of length $\#n_{in}$
\item Max-MGCR$^*$, which asks for the maximum value $k$ such that $M$ is
       $k$-robust relative to $H = M$ for every possible Boolean input 
       vector $I$ of length $\#n_{in}$
\end{itemize}

\noindent
The derivation of such results is complicated by both the $coNP$-hardness 
of the decision versions of these problems (and the scarcity of 
inapproximability results for $coNP$-hard problems which one could 
transfer to our problems by $L$-reductions) and the fact that the
optimization versions of our problems are evaluation problems that
return numbers rather than graph structures (which makes the use of
instance-copy and gap inapproximability proof techniques 
[\citealt[Chapter 6]{garey1979computers}] extremely difficult). We shall sidestep many
of these issues by deriving our results within the $OptP$ framework for
analyzing evaluation problems developed in \citealt{Kre88,gasarchOptP1995}.
In particular, we shall find the following of use.

\begin{definition}
(Adapted from \cite[page 493]{Kre88})
Let $f, g : \Sigma^* \rightarrow \mathcal{Z}$. A {\bf metric reduction}
from $f$ to $g$ is a pair $(T_1, T_2)$ of polynomial-time computable 
functions where $T_1 : \Sigma^* \rightarrow \Sigma^*$ and $T_2 : \Sigma^* 
\times \mathcal{Z} \rightarrow \mathcal{Z}$ such that $f(x) = T_2(x, 
g(T_1(x)))$ for all $x \in \Sigma^*$.
\end{definition}

\begin{lemma}
(Corollary of \cite[Theorem 4.3]{Kre88})
Given an evaluation problem $\Pi$ that is $OptP[O(\log n)]$-hard under
metric reductions,  if $\Pi$ has a $c$-additive
approximation algorithm for some $c \in o(poly)$\footnote{
Note that $o(poly)$ is the set of all functions $f$ that are strictly
upper bounded by all polynomials of $n$, i.e., $f(n) \leq c \times g(n)$
for $n \geq n_0$ for all $c > 0$ and $g(n) \in \cup_k n^k = n^{O(1)}$.
} 
then $P = NP$.
\label{LemOptPlogInapprox}
\end{lemma}

\noindent
Some of our metric reductions  use specialized ReLU logic
gates described in \citealt[Lemma 13]{barcelo_model_2020}. These gates assume Boolean
neuron input and output values of 0 and 1 and are structured as follows:

\begin{enumerate}
\item NOT ReLU gate:  A ReLU gate with one input connection weight of value
       $-1$ and a 
       bias of 1. This gate has output 1 if the input is 0 and 0 otherwise.
\item $n$-way AND ReLU gate: A ReLU gate with $n$ input connection weights 
       of value 1 and a bias of $-(n - 1)$. This gate has output 1 if all 
       inputs have value 1 and 0 otherwise.
\item $n$-way OR ReLU gate: A combination of an $n$-way AND ReLU gate with
       NOT ReLU gates on all of its inputs and a NOT ReLU gate on its 
       output that uses DeMorgan's Second Law to implement 
       $(x_1 \vee x_2 \vee \ldots x_n)$ as $\neg(\neg x_1 \wedge \neg x_2 
       \wedge \ldots \neg x_n)$. This gate has 
       output 1 if any input has value 1 and 0 otherwise.
\end{enumerate}

The hardness (inapproximability) results in this section hold
when the given MLP $M$ has three (six) hidden layers.

\subsection{Results for MLCR-special and MLCR}

Let us first consider problem MLCR$^*$.

\begin{theorem}
If MLCR$^*$ is polynomial-time tractable then $P = NP$.
\label{ThmMLCR*MLCA}
\end{theorem}
\begin{proof}
As MLCR$^*$ is the complement of MLCA by Observation \ref{ObsMLCR*MLCA}
and MLCA is $NP$-hard by the proof of \Cref{ThmMLCACli}, Lemma \ref{LemcoNPI} implies that MLCR$^*$ is $coNP$-hard. The result then follows from Lemma \ref{LemcoNPII}.
\end{proof}

\vspace*{0.15in}

\begin{theorem}
If $\la cd, \#n_{in}, \#n_{out}, W_{\max}, B_{\max}, k\ra$-MLCR$^*$ is 
fixed-parameter tractable \linebreak then $FPT = W[1]$.
\label{ThmMLCR*_fpi1}
\end{theorem}
\begin{proof}
The $coW[1]$-hardness of  $\la cd, \#n_{in}, \#n_{out}, W_{\max}, B_{\max}, 
k\ra$-MLCR$^*$ follows from the $W[1]$-hardness of  $\la cd, \#n_{in},
\#n_{out}, W_{\max}, B_{\max}, k\ra$-MLCA (\Cref{ThmMLCA_fpi1}),
Observation \ref{ObsMLCR*MLCA}, and Lemma \ref{LemcoWI}. The result
then follows from Lemma \ref{LemcoWII}.
\end{proof}

\begin{theorem}
$\la \#n_{tot} \ra$-MLCR$^*$ is fixed-parameter tractable.
\label{ThmMLCR*_fpt1}
\end{theorem}
\begin{proof}
The result follows from the fixed-parameter tractability of
$\la \#n_{tot} \ra$-MLCA \Cref{ThmMLCA_fpt1}, Observation 
\ref{ObsMLCR*MLCA}, and Lemma \ref{LemcoWIa}.
\end{proof}

\vspace*{0.15in}

\begin{theorem}
$\la cw, cd \ra$-MLCR$^*$ is fixed-parameter tractable.
\label{ThmMLCR*_fpt2}
\end{theorem}
\begin{proof}
The result follows from the fixed-parameter tractability of
$\la cw, cd\ra$-MLCA \Cref{ThmMLCA_fpt2}, Observation 
\ref{ObsMLCR*MLCA}, and Lemma \ref{LemcoWIa}.
\end{proof}

\vspace*{0.15in}

\noindent
Observe that the results in Theorems 
\ref{ThmMLCR*_fpi1}--\ref{ThmMLCR*_fpt2}
in combination with Lemmas \ref{LemPrmFPProp1} and \ref{LemPrmFPProp2}
suffice to establish the parameterized complexity status of MLCR$^*$ 
relative to many subsets of the parameters listed in Table \ref{TabPrmMCR}.

We can derive a polynomial-time additive inapproximability result for
Max-MLCR$^*$ using the following chain of metric reductions based on
two evaluation problems:

\begin{itemize}
\item Min-VC, which asks for the minimum value $k$ such that $G$ has
       a vertex cover of size $k$.
\item Min-MLCA, which asks for the minimum value $k$ such that there is
       a $k$-size subset $N'$ of the $|N|$ neurons in $M$ such that 
       $M(I) \neq (M/N')(I)$.
\end{itemize}

\begin{lemma}
Min-VC metric reduces to Min-MLCA.
\label{LemMinVC_MinMLCA}
\end{lemma}
\begin{proof}
Consider the following reduction from Min-VC to Min-MLCA. Given an 
instance $X = \la G = (V,E)\ra$ of Min-VS, construct the following 
instance $X' = \la M, I\ra$ of Min-MLCA:
Let $M$ be an MLP based on $\#n_{tot} = 3|V| + 2|E| + 2$ neurons spread
across six layers:

\begin{enumerate}
\item {\bf Input neuron layer}: The single input neuron $n_{in}$ (bias $+1$).
\item {\bf Hidden vertex pair layer}: The vertex neurons $nvP1_1, nvP1_2, 
        \ldots nvP1_{|V|}$ and $nvP2_1, nvP2_2, \ldots nvP2_{|V|}$
	(all with bias 0).
\item {\bf Hidden vertex AND layer}: The vertex neurons $nvA_1, nvA_2, 
        \ldots nA_{|V|}$, all of which are 2-way AND ReLU gates.
\item {\bf Hidden edge AND layer}: The edge neurons $neA_1, neA_2, \ldots 
         neA_{|E|}$, all of which are 2-way AND ReLU gates.
\item {\bf Hidden edge NOT layer}: The edge neurons $neN_1, neN_2, \ldots 
        neN_{|E|}$, all of which are NOT ReLU gates.
\item {\bf Output layer}: The single output neuron $n_{out}$, which is
        an $|E|$-way AND ReLU gate.
\end{enumerate}

\noindent
The non-zero weight connections between adjacent layers are as follows:

\begin{itemize}
\item Each input neuron has an edge of weight 0 coming from its
       corresponding input and is in turn connected to each of the vertex 
       pair neurons with weight 1.
\item Each vertex pair neuron $nvP1_i$ ($nvP2_i$), $1 \leq i \leq 
       |V|$, is connected to vertex AND neuron $nvR_i$ with 
       weight 2 (0).
\item Each vertex AND neuron $nvA_i$, $1 \leq i \leq |V|$, is connected to 
       each edge AND neuron 
       whose corresponding edge has an endpoint $v_i$ with weight 1.
\item Each edge AND neuron $neA_i$, $1 \leq i \leq |E|$, is connected to 
       edge NOT neuron $neN_i$ with weight 1.
\item Each edge NOT neuron $neN_i$, $1 \leq i \leq |E|$, is connected to the output neuron
       $n_{out}$ with weight 1.
\end{itemize}

\noindent
All other connections between neurons in adjacent layers have weight 0.
Finally, let $I = (1)$. Observe that this instance of Min-MLCA can be 
created in time polynomial in the size of the given instance of Min-VC. 
Moreover, the output behaviour of the neurons in $M$ from the presentation
of input $I$ until the output is generated is as follows:

\begin{center}
\begin{tabular}{| c || l |}
\hline
timestep & neurons (outputs) \\
\hline\hline
0 & --- \\
\hline
1 & $n_{in} (1)$ \\
\hline
2 & $nvP1_1 (1), nvP1_2 (1), \ldots nvP1_{|V|} (1),
     nvP2_1 (1), nvP2_2 (1), \ldots nvP2_{|V|} (1)$ \\
\hline
3 & $nvA_1 (1), nvA_2 (1), \ldots nvA_{|V|} (1)$ \\
\hline
4 & $neA_1 (1), neA_2 (1), \ldots neA_{|V|} (1)$ \\
\hline
5 & $neN_1 (0), neN_2 (0), \ldots neN_{|V|} (0)$ \\
\hline
6 & $n_{out} (0))$ \\
\hline
\end{tabular}
\end{center}
 
\noindent
Note that, given the 0 (2) connection-weights of P2 (P1) vertex pair 
neurons to vertex AND neurons, it is the outputs of P1 vertex pair neurons
in timestep 2 that enables vertex AND neurons to output 1 in timestep 3.

We now need to show the correctness of this reduction by proving that the 
answer for the given instance of Min-VC is $k$ if and only if the answer 
for the constructed instance of Min-MLCA is $k$. We prove the two 
directions of this if and only if separately as follows:

\begin{description}
\item [$\Rightarrow$]: Let $V' = \{v'_1, v'_2, \ldots, v'_k\} \subseteq V$ 
       be a minimum-size vertex cover in $G$ of size $k$ and $N'$ be the 
       $k$-sized sized subset of the P1 vertex pair neurons corresponding 
       to the vertices in $V'$. Let $M'$ be the version of $M$ in which all
       neurons in $N'$ are ablated. As each of these vertex pair neurons
       previously allowed their associated vertex AND
       neurons to output 1, 
       their ablation now allows these $k$ vertex AND neurons to
       output 0.  As $V'$ is a vertex cover of size $k$, 
       all edge AND neurons in $M'$ receive
       inputs of 0 on at least one of their endpoints from the vertex
       AND neurons associated with the P1 vertex pair neurons in $N'$.
       This in turn ensures that all of the edge NOT neurons and the 
       output neuron produces output 1.  Hence, $M(I) = 0 \neq 1 = M'(I)$.
\item [$\Leftarrow$]: Let $N'$ be a minimum-sized subset of $N$ of size $k$
       such that for the MLP $M'$ induced by ablating all neurons in $N'$,
       $M(I) \neq M(I')$. As $M(I) = 0$ and circuit outputs are stepped to
       be Boolean, $M'(I) = 1$. Given the bias of the output neuron, this 
       can only occur if all $|E|$ edge NOT (AND) neurons in $M'$ have 
       output 1 (0) on input $I$, the latter of which requiring that each
       edge AND neuron receives 0 from at least one of its endpoint vertex
       AND neurons. These vertex AND neurons can only output 0 if all of 
       their associated P1 vertex neurons have been ablated.  This means 
       that the vertices in $G$ corresponding to the P1 vertex pair neurons
       in $N'$ must form a vertex cover of size $k$ in $G$.
\end{description}

\noindent
As this proves that Min-VC($X$) $=$ Min-MLCA($X'$), the reduction above
is a metric reduction from Min-VC to Min-MLCA.
\end{proof}

\begin{lemma}
Min-MLCA metric reduces to Max-MLCR$^*$.
\label{LemMinMLCA_MaxMLCR*}
\end{lemma}
\begin{proof}
As MLCA is the complement problem of MLCR$^*$ (Observation
\ref{ObsMLCR*MLCA}), we already have a trivial reduction from  MLCA to
MLCR$^*$ on their common input $X = \la M, I\ra$. We can then show that
$k$ is the minimum value such that $M$ has a $k$ circuit ablation relative
to $I$ if and only if $k - 1$ is the maximum value such that $M$ is 
$k-1$-robust relative to $I$:

\begin{description}
\item[$\Rightarrow$] If $k$ is is the minimum value such that $M$ has a 
       $k$-sized circuit ablation relative to $I$ then no subset of $M$ of
       size $k - 1$ can be a circuit ablation of $M$ relative to $I$, and 
       $M$ is $(k - 1)$-robust relative to $I$. Moreover, $M$ cannot be 
       $k$-robust relative to $I$ as that would contradict the existence 
       of a $k$-sized circuit ablation for $M$ relative to $I$. Hence, 
       $k - 1$ is the maximum robustness value for $M$ relative to $I$.
\item[$\Leftarrow$] If $k - 1$ is the maximum value such that $M$ is 
       $(k - 1)$-robust relative to $I$ then there must be a subset $H$ of
       $M$ of size $k$ that ensures $M$ is not $k-1$-robust, i.e., 
       $(M/H)(I) \neq M(I)$. Such an $H$ is a $k$-sized circuit ablation of
       $M$ relative to $I$. Moreover, there cannot be a $(k - 1)$-sized 
       circuit ablation of $M$ relative to $I$ as that would contradict the
       $(k - 1)$-robustness of $M$ relative to $I$. Hence, $k$ is the
       minimum size of circuit ablations for $M$ relative to $I$.
\end{description}

\noindent
As this proves that Min-MLCA($X$) $=$ Max-MLCR$^*$($X$) $+ 1$, the 
reduction above is a metric reduction from Min-MLCA to Max-MLCR$^*$.
\end{proof}

\begin{theorem}
If Max-MLCR$^*$has a $c$-additive approximation algorithm for some 
$c \in o(poly)$ then $P = NP$.
\label{ThmMLCR*Inapprox}
\end{theorem}

\begin{proof}
The $OptP[O(\log n)]$-hardness of Max-MLCR$^*$ follows from the
$OptP[O(\log n)]$-hardness of Min-VC \cite[Theorem Theorem 3.3]{gasarchOptP1995} and
the metric reductions in Lemmas \ref{LemMinVC_MinMLCA} and
\ref{LemMinMLCA_MaxMLCR*}. The result then follows from
Lemma \ref{LemOptPlogInapprox}
\end{proof}

\vspace*{0.15in}

Let us now consider problem MLCR.

\begin{lemma}
MLCR$^*$ many-one polynomial-time reduces to MLCR.
\label{LemMLCR*MLCR}
\end{lemma}

\begin{proof}
Follows from the trivial reduction in which an instance $\la M, I, k\ra$
of MLCR$^*$ is transformed into an instance $\la M, H = M, I, k\ra$ of
MLCR.
\end{proof}

\begin{theorem}
If MLCR is polynomial-time tractable then $P = NP$.
\label{ThmMLCR_MLCR*}
\end{theorem}
\begin{proof}
Follows from the $coNP$-hardness of MLCR$^*$ (Theorem \ref{ThmMLCR*MLCA}),
the reduction in Lemma \ref{LemMLCR*MLCR}, and Lemma \ref{LemcoNPII}.
\end{proof}

\vspace*{0.15in}

\begin{theorem}
If $\la cd, \#n_{in}, \#n_{out}, W_{\max}, B_{\max}, k\ra$-MLCR is 
fixed-parameter tractable \linebreak then $FPT = W[1]$.
\label{ThmMLCR_fpi1}
\end{theorem}
\begin{proof}
Follows from the $coW[1]$-hardness of  $\la cd, \#n_{in}, \#n_{out}, 
W_{\max}, B_{\max}, k\ra$-MLCR$^*$ (Theorem \ref{ThmMLCR*_fpi1}), 
the reduction in Lemma \ref{LemMLCR*MLCR}, and Lemma \ref{LemcoWII}.
\end{proof}

\begin{theorem}
$\la |H| \ra$-MLCR is fixed-parameter tractable.
\label{ThmMLCR_fpt1}
\end{theorem}
\begin{proof}
Consider the algorithm that generates every possible subset $H'$ of size
at most $k$ of the neurons $N$ in $H$ and for each such subset (assuming 
$M/H'$ is active) checks if $(M/H')(I) \neq M(I)$. If such a subset is 
found, return ``No''; otherwise, return ``Yes''. The number of possible 
subsets $H'$ is at most $k \times |H|^k \leq |H| \times 
|H|^{|H|}$. As any such $M'$ can be generated from $M$, checked
or activity, and run on $I$ in time polynomial in the size of the given 
instance of MLCR, the above is a fixed-parameter tractable
algorithm for MLCR relative to parameter-set $\{ |H| \}$.
\end{proof}

\vspace*{0.15in}

\begin{theorem}
$\la cw, cd \ra$-MLCR is fixed-parameter tractable.
\label{ThmMLCR_fpt2}
\end{theorem}
\begin{proof}
Follows from the algorithm in the proof of Theorem \ref{ThmMLCR_fpt1} and
the observation that $|H| \leq cw \times cd$.
\end{proof}

\vspace*{0.15in}

\begin{theorem}
$\la \#n_{tot} \ra$-MLCR is fixed-parameter tractable.
\label{ThmMLCR_fpt3}
\end{theorem}
\begin{proof}
Follows from the algorithm in the proof of Theorem \ref{ThmMLCR_fpt1} and
the observation that $|H| \leq \#n_{tot}$.
\end{proof}

\vspace*{0.15in}

\begin{theorem}
If Max-MLCR has a $c$-additive approximation algorithm for some 
$c \in o(poly)$ then $P = NP$.
\label{ThmMLCRInapprox}
\end{theorem}

\begin{proof}
As Max-MLCR$^*$ is a special case of Max-MLCR, if Max-MLCR has a 
$c$-additive approximation algorithm for some $c$ then so does
Max-MLCR$^*$. The result then follows from Theorem \ref{ThmMLCR*Inapprox}
\end{proof}

\vspace*{0.15in}

\subsection{Results for MGCR-special and MGCR}

Consider the following variant of MGCA:

\vspace*{0.1in}

\noindent
{\sc Special Circuit Ablation} (SCA) \\
{\em Input}: A multi-layer perceptron $M$ of depth $cd$ with $\#n_{tot}$
              neurons and maximum layer width $cw$, connection-value 
              matrices $W_1, W_2, \ldots, W_{cd}$, neuron bias vector $B$,
              and a 
              positive integer $k$ such that $1 \leq k \leq \#n_{tot}$. \\
{\em Question}: Is there a subset $N'$, $|N'| \leq k$, of the $|N|$ neurons
               in $M$ such that for the MLP $M'$ induced by $N \setminus 
	       N'$, $M(I) \neq M'(I)$ for some Boolean input
	       vector $I$ of length $\#n_{in}$?

\vspace*{0.1in}

\begin{theorem}
If SCA is polynomial-time tractable then $P = NP$.
\label{ThmSCA_MLCA}
\end{theorem}
\begin{proof}
Observe that in the instance of MLCA constructed by the reduction from
{\sc Clique} in the proof of \Cref{ThmMLCA_fpi1}, the 
input-connection weight 0 and bias 1 of the input neuron force this neuron 
to output 1 for both of the possible input vectors $(1)$ and $(0)$. This 
means that the answer to the given instance of {\sc Clique} is ``Yes'' if 
and only if the answer to the constructed instance of MLCA relative to any 
of its possible input vectors is ``Yes''. Hence, with slight modifications 
to the proof of reduction correctness, this reduction also establishes the
$NP$-hardness of SCA.
\end{proof}

\vspace*{0.15in}

\begin{theorem}
If $\la cd, \#n_{in}, \#n_{out}, W_{\max}, B_{\max}, k\ra$-SCA is 
fixed-parameter tractable then $FPT = W[1]$.
\label{ThmSCA_fpi1}
\end{theorem}
\begin{proof}
Observe that in the instance of SCA constructed in the reduction in the 
proof of Theorem \ref{ThmSCA_MLCA}, $\#n_{in} = \#n_{out} = W_{\max} = 1$, 
$cd = 5$, and $B_{\max}$ and $k$ are function of $k$ in the given instance 
of {\sc Clique}. The result then follows from the fact that 
$\la k \ra$-{\sc Clique} is $W[1]$-hard \citep{downeyParameterizedComplexity1999a}.
\end{proof}

\begin{theorem}
$\la \#n_{tot} \ra$-SCA is fixed-parameter tractable.
\label{ThmSCA_fpt1}
\end{theorem}
\begin{proof}
Modify the algorithm in the proof of \Cref{ThmMLCA_fpt1} such
that each created MLP $M$ is checked to ensure that $M(I) \neq M'(I)$ for every possible Boolean input vector of length $\#n_{in}$. As the number of such vectors is $2^{\#n_{in}} \leq 2^{\#n_{tot}}$,  the above is a fixed-parameter tractable algorithm for SCA relative to parameter-set $\{ \#n_{tot} \}$.
\end{proof}

\vspace*{0.15in}

\begin{theorem}
$\la cw, cd \ra$-SCA is fixed-parameter tractable.
\label{ThmSCA_fpt2}
\end{theorem}
\begin{proof}
Follows from the algorithm in the proof of Theorem \ref{ThmSCA_fpt1} and
the observation that $\#n_{tot} \leq cw \times cd$.
\end{proof}

\vspace*{0.15in}

\noindent
Our results above for SCA above gain importance for us here courtesy of
the following observation.

\begin{observation}
MGCR$^*$ is the complement of SCA. 
\label{ObsMGCR*SCA}
\end{observation}

\begin{proof}
Let $X = \la M, k\ra$ be a shared input of MGCR$^*$ and SCA. If the
the answer to SCA on input $X$ is ``Yes'', this means that there is a
subset $H$ of size $\leq k$ of the neurons of $M$ that can be ablated
such that $M(I) \neq (M/H)(I)$ for some input vector $I$. This implies that
the answer to MGCR$^*$ on input $X$ is ``No''. Conversely, if the answer to
SCA on input $X$ is ``No'', this means that there is no subset $H$ of size 
$\leq k$ of the neurons of $M$ that can be ablated such that $M(I) \neq 
(M/H)(I)$ for any input vector $I$. This 
implies that the answer to MGCR$^*$ on input $X$ is ``Yes'. The
observation follows from the definition of complement problem.
\end{proof}

\begin{theorem}
If MGCR$^*$ is polynomial-time tractable then $P = NP$.
\label{ThmMGCR*SCA}
\end{theorem}
\begin{proof}
As MGCR$^*$ is the complement of SCA by Observation \ref{ObsMGCR*SCA}
and SCA is $NP$-hard (Theorem \ref{ThmSCA_MLCA}), Lemma \ref{LemcoNPI} 
implies that MGCR$^*$ is $coNP$-hard. The result then follows from Lemma
\ref{LemcoNPII}.
\end{proof}

\vspace*{0.15in}

\begin{theorem}
If $\la cd, \#n_{in}, \#n_{out}, W_{\max}, B_{\max}, k\ra$-MGCR$^*$ is 
fixed-parameter tractable \linebreak then $FPT = W[1]$.
\label{ThmMGCR*_fpi1}
\end{theorem}
\begin{proof}
The $coW[1]$-hardness of  $\la cd, \#n_{in}, \#n_{out}, W_{\max}, B_{\max}, 
k\ra$-MGCR$^*$ follows from the $W[1]$-hardness of  $\la cd, \#n_{in},
\#n_{out}, W_{\max}, B_{\max}, k\ra$-SCA (Theorem \ref{ThmSCA_fpi1}),
Observation \ref{ObsMGCR*SCA}, and Lemma \ref{LemcoWI}. The result
then follows from Lemma \ref{LemcoWII}.
\end{proof}

\begin{theorem}
$\la \#n_{tot} \ra$-MGCR$^*$ is fixed-parameter tractable.
\label{ThmMGCR*_fpt1}
\end{theorem}
\begin{proof}
The result follows from the fixed-parameter tractability of
$\la \#n_{tot} \ra$-SCA (Theorem \ref{ThmSCA_fpt1}), Observation 
\ref{ObsMGCR*SCA}, and Lemma \ref{LemcoWIa}.
\end{proof}

\vspace*{0.15in}

\begin{theorem}
$\la cw, cd \ra$-MGCR$^*$ is fixed-parameter tractable.
\label{ThmMGCR*_fpt2}
\end{theorem}
\begin{proof}
The result follows from the fixed-parameter tractability of
$\la cw, cd\ra$-SCA (Theorem \ref{ThmSCA_fpt2}), Observation 
\ref{ObsMGCR*SCA}, and Lemma \ref{LemcoWIa}.
\end{proof}

\vspace*{0.15in}

\noindent
Observe that the results in Theorems 
\ref{ThmMGCR*_fpi1}--\ref{ThmMGCR*_fpt2}
in combination with Lemmas \ref{LemPrmFPProp1} and \ref{LemPrmFPProp2}
suffice to establish the parameterized complexity status of MGCR$^*$ 
relative to many subsets of the parameters listed in Table \ref{TabPrmMCR}.

We can derive a polynomial-time additive inapproximability result for
Max-MGCR$^*$ using the following chain of metric reductions based on
two evaluation problems:

\begin{itemize}
\item Min-VC, which asks for the minimum value $k$ such that $G$ has
       a vertex cover of size $k$.
\item Min-SCA, which asks for the minimum value $k$ such that there is
       a $k$-size subset $N'$ of the $|N|$ neurons in $M$ such that 
       $M(I) \neq (M/N')(I)$ for some Boolean input vector $I$ of
       length$\#n_{in}$.
\end{itemize}

\begin{lemma}
Min-VC metric reduces to Min-SCA.
\label{LemMinVC_MinSCA}
\end{lemma}
\begin{proof}
Observe that in the instance of Min-MLCA constructed by the reduction from
Min-VC in the proof of Theorem \ref{LemMinVC_MinMLCA}, the
input-connection weight 0 and bias 1 of the input neuron force this neuron 
to output 1 for both of the possible input vectors $(1)$ and $(0)$. This 
means that the answer to the constructed instance of Min-MLCA is the
same relative to any of its possible input vectors. Hence, with slight 
modifications to the proof of reduction correctness, this reduction is
also a metric reduction from Min-VC to Min-SCA such that for 
given instance $X$ of Min-VC and constructed instance $X'$ of Min-SCA,
Min-VC($X$) $=$ Min-SCA($X'$).
\end{proof}

\begin{lemma}
Min-SCA metric reduces to Max-MGCR$^*$.
\label{LemMinSCA_MaxMGCR*}
\end{lemma}
\begin{proof}
As SCA is the complement problem of MGCR$^*$ (Observation
\ref{ObsMGCR*SCA}), we already have a trivial reduction from  SCA to
MGCR$^*$ on their common input $X = \la M\ra$. We can then show that
$k$ is the minimum value such that $M$ has a $k$ circuit ablation relative
to some possible $I$ if and only if $k - 1$ is the maximum value such 
that $M$ is $k-1$-robust relative to all possible $I$:

\begin{description}
\item[$\Rightarrow$] If $k$ is is the minimum value such that $M$ has a 
       $k$-sized circuit ablation relative to some possible $I$ then no 
       subset of $M$ of size $k - 1$ can be a circuit ablation of $M$ 
       relative to any possible $I$, and $M$ is $(k - 1)$-robust relative 
       to all possible $I$. Moreover, $M$ cannot be $k$-robust relative to
       all possible  $I$ as that would contradict the existence of a 
       $k$-sized circuit ablation for $M$ relative to some possible $I$. 
       Hence, $k - 1$ is the maximum robustness value for $M$ relative to
       all possible $I$.
\item[$\Leftarrow$] If $k - 1$ is the maximum value such that $M$ is 
       $(k - 1)$-robust relative to all possible $I$ then there must be a 
       subset $H$ of $M$ of size $k$ that ensures $M$ is not $k-1$-robust
       relative to all possible $I$, i.e., $(M/H)(I) \neq M(I)$ for some 
       possible $I$. Such an $H$ is a $k$-sized circuit ablation of
       $M$ relative to that $I$. Moreover, there cannot be a 
       $(k - 1)$-sized circuit ablation of $M$ relative to some possible
       $I$ as that would contradict the $(k - 1)$-robustness of $M$ 
       relative to all possible $I$. Hence, $k$ is the minimum size of 
       circuit ablations for $M$ relative to some possible $I$.
\end{description}

\noindent
As this proves that Min-SCA($X$) $=$ Max-MGCR$^*$($X$) $+ 1$, the 
reduction above is a metric reduction from Min-SCA to Max-MGCR$^*$.
\end{proof}

\begin{theorem}
If Max-MGCR$^*$has a $c$-additive approximation algorithm for some 
$c \in o(poly)$ then $P = NP$.
\label{ThmMGCR*Inapprox}
\end{theorem}

\begin{proof}
The $OptP[O(\log n)]$-hardness of Max-MGCR$^*$ follows from the
$OptP[O(\log n)]$-hardness of Min-VC \citep[Theorem Theorem 3.3]{gasarchOptP1995} and
the metric reductions in Lemmas \ref{LemMinVC_MinSCA} and
\ref{LemMinSCA_MaxMGCR*}. The result then follows from
Lemma \ref{LemOptPlogInapprox}
\end{proof}

\vspace*{0.15in}

Let us now consider problem MGCR.

\begin{lemma}
MGCR$^*$ many-one polynomial-time reduces to MGCR.
\label{LemMGCR*MGCR}
\end{lemma}

\begin{proof}
Follows from the trivial reduction in which an instance $\la M, k\ra$
of MGCR$^*$ is transformed into an instance $\la M, H = M, k\ra$ of
MGCR.
\end{proof}

\begin{theorem}
If MGCR is polynomial-time tractable then $P = NP$.
\label{ThmMGCR_MLCR*}
\end{theorem}
\begin{proof}
Follows from the $coNP$-hardness of MGCR$^*$ (Theorem \ref{ThmMGCR*SCA}),
the reduction in Lemma \ref{LemMGCR*MGCR}, and Lemma \ref{LemcoNPII}.
\end{proof}

\vspace*{0.15in}

\begin{theorem}
If $\la cd, \#n_{in}, \#n_{out}, W_{\max}, B_{\max}, k\ra$-MGCR is 
fixed-parameter tractable \linebreak then $FPT = W[1]$.
\label{ThmMGCR_fpi1}
\end{theorem}
\begin{proof}
Follows from the $coW[1]$-hardness of  $\la cd, \#n_{in}, \#n_{out}, 
W_{\max}, B_{\max}, k\ra$-MGCR$^*$ (Theorem \ref{ThmMGCR*_fpi1}), 
the reduction in Lemma \ref{LemMGCR*MGCR}, and Lemma \ref{LemcoWII}.
\end{proof}

\begin{theorem}
$\la \#n_{in}, |H| \ra$-MGCR is fixed-parameter tractable.
\label{ThmMGCR_fpt1}
\end{theorem}
\begin{proof}
Modify the algorithm in the proof of Theorem \ref{ThmMLCR_fpt1} such
that each created MLP $M$ is checked to ensure that $M(I) \neq (H/H')(I)$ 
for every possible Boolean input vector of length $\#n_{in}$. As the number
of such vectors is $2^{\#n_{in}}$, the above is a fixed-parameter tractable
algorithm for MGCR relative to parameter-set $\{ \#n_{in}, |H| \}$.
\end{proof}

\vspace*{0.15in}

\begin{theorem}
$\la cw, cd \ra$-MGCR is fixed-parameter tractable.
\label{ThmMGCR_fpt2}
\end{theorem}
\begin{proof}
Follows from the algorithm in the proof of Theorem \ref{ThmMGCR_fpt1} and
the observations that $\#n_{tot} \leq cw \times cd$ and
$|H| \leq cw \times cd$.
\end{proof}

\vspace*{0.15in}

\begin{theorem}
$\la \#n_{tot} \ra$-MGCR is fixed-parameter tractable.
\label{ThmMGCR_fpt3}
\end{theorem}
\begin{proof}
Follows from the algorithm in the proof of Theorem \ref{ThmMGCR_fpt1} and
the observations that $\#n_{tot} \leq \#n_{tot}$ and $|H| \leq \#n_{tot}$.
\end{proof}

\vspace*{0.15in}

\begin{theorem}
If Max-MGCR has a $c$-additive approximation algorithm for some 
$c \in o(poly)$ then $P = NP$.
\label{ThmMGCRInapprox}
\end{theorem}

\begin{proof}
As Max-MGCR$^*$ is a special case of Max-MGCR, if Max-MGCR has a 
$c$-additive approximation algorithm for some $c$ then so does
Max-MGCR$^*$. The result then follows from Theorem \ref{ThmMGCR*Inapprox}
\end{proof}

\section{Sufficient Reasons Problem}

\noindent
{\sc Minimum sufficient reason} (MSR) \\
{\em Input}: A multi-layer perceptron $M$ of depth $cd$ with $\#n_{tot}$
              neurons and maximum layer width $cw$, connection-value 
              matrices $W_1, W_2, \ldots, W_{cd}$, neuron bias vector $B$,
              a Boolean input vector $I$ of length $\#n_{in}$, and a 
              positive integer $k$ such that $1 \leq k \leq \#n_{in}$. \\
{\em Question}: Is there a $k$-sized subset $I'$ of $I$ such that for each
              possible completion $I''$ of $I'$, $I$ and $I''$ are 
              behaviorally equivalent with respect to $M$?

\vspace*{0.1in}







\noindent
For a graph $G = (V, E)$, we shall assume an ordering on the vertices and edges in $V$ 
and $E$, respectively. For each vertex $v \in V$, let the complete neighbourhood $N_C(v)$ of $v$ be the set
composed of $v$ and the set of all vertices in $G$ that are adjacent to $v$
by a single edge, i.e., $v \cup \{ u ~ | ~ u ~ \in V ~ \rm{and} ~ (u,v) \in E\}$.

We will prove various classical and parameterized results for MSR 
using reductions from {\sc Clique}. The
parameterized results are proved relative to the parameters in 
Table \ref{TabPrm}. An additional reduction from DS 
(Theorem \ref{ThmMSRDS}) use specialized ReLU logic
gates described in \citealt[Lemma 13]{barcelo_model_2020}. These gates assume Boolean
neuron input and output values of 0 and 1 and are structured as follows:

\begin{enumerate}
\item NOT ReLU gate:  A ReLU gate with one input connection weight of value
       $-1$ and a 
       bias of 1. This gate has output 1 if the input is 0 and 0 otherwise.
\item $n$-way AND ReLU gate: A ReLU gate with $n$ input connection weights 
       of value 1 and a bias of $-(n - 1)$. This gate has output 1 if all 
       inputs have value 1 and 0 otherwise.
\item $n$-way OR ReLU gate: A combination of an $n$-way AND ReLU gate with
       NOT ReLU gates on all of its inputs and a NOT ReLU gate on its 
       output that uses DeMorgan's Second Law to implement 
       $(x_1 \vee x_2 \vee \ldots x_n)$ as $\neg(\neg x_1 \wedge \neg x_2 
       \wedge \ldots \neg x_n)$. This gate has 
       output 1 if any input has value 1 and 0 otherwise.
\end{enumerate}

\begin{table}[b]
\caption{Parameters for the minimum sufficient reason problem.
}
\vspace*{0.1in}
\label{TabPrm}
\centering
\begin{tabular}{| c || l | }
\hline
Parameter     & Description \\
\hline\hline
$cd$        & \# layers in given MLP \\
\hline
$cw$        & max \# neurons in layer in given MLP \\
\hline
$\#n_{tot}$ & total \# neurons in given MLP \\
\hline
$\#n_{in}$  & \# input neurons in given MLP \\
\hline
$\#n_{out}$ & \# output neurons in given MLP \\
\hline
$B_{\max}$  & max neuron bias in given MLP \\
\hline
$W_{\max}$  & max connection weight in given MLP \\
\hline
$k$           & Size of requested subset of input vector \\
\hline
\end{tabular} 
\end{table}

\subsection{Results for MSR}

The following hardness results are notable for holding
when the given MLP $M$ has only one hidden layer. As such, they complement and significantly tighten results given in \citep{barcelo_model_2020,WM+21}, respectively.

\begin{theorem}
If MSR is polynomial-time tractable then $P = NP$.
\label{ThmMSRCli}
\end{theorem}
\begin{proof}
Consider the following reduction from {\sc Clique} to MSR.
Given an instance $\la G = (V,E), k\ra$ of {\sc Clique}, construct the following 
instance $\la M, I, k'\ra$ of MSR:
Let $M$ be an MLP based on $\#n_{tot} = |V| + |E| + 1$ neurons spread
across three layers:

\begin{enumerate}
\item {\bf Input vertex layer}: The vertex neurons $nv_1, nv_2, \ldots nv_{|V|}$
        (all with bias 0).
\item {\bf Hidden edge layer}: The edge neurons $ne_1, ne_2, \ldots ne_{|E|}$
        (all with bias $-1$).
\item {\bf Output layer}: The single output neuron $n_{out}$ (bias $-(k(k - 1)/2 - 1)$).
\end{enumerate}

\noindent
Note that this MLP has only one hidden layer.
The non-zero weight connections between adjacent layers are as follows:

\begin{itemize}
\item Each vertex neuron $nv_i$, $1 \leq i \leq |V|$, is connected to each edge neuron 
       whose corresponding edge has an endpoint $v_i$ with weight 1.
\item Each edge neuron $ne_i$, $1 \leq i \leq |E|$, is connected to the output neuron
       $n_{out}$ with weight 1.
\end{itemize}

\noindent
All other connections between neurons in adjacent layers have weight 0.
Finally, let $I = (1)$ and $k' = k$.
Observe that this instance of MSR can be created in time polynomial in the size of the 
given instance of {\sc Clique}. Moreover, the output behaviour of the neurons in 
$M$ from the presentation of input $I$ until the output is generated is as follows:

\begin{center}
\begin{tabular}{| c || l |}
\hline
timestep & neurons (outputs) \\
\hline\hline
0 & --- \\
\hline
1 & $nv_1 (1), nv_2 (1), \ldots nv_{|V|} (1)$ \\
\hline
2 & $ne_1 (1), ne_2 (1), \ldots ne_{|E|} (1)$ \\
\hline
3 & $n_{out} (|E| - (k(k - 1)/2 - 1))$ \\
\hline
\end{tabular}
\end{center}
 
\noindent
Note that it is the stepped output of $n_{out}$ in timestep 3 that yields output 1.

We now need to show the correctness of this reduction by proving that the answer
for the given instance of {\sc Clique} is ``Yes'' if and only if the answer for the 
constructed instance of MSR is ``Yes''. We prove the two 
directions of this if and only if separately as follows:

\begin{description}
\item [$\Rightarrow$]: Let $V' = \{v'_1, v'_2, \ldots, v'_k\} \subseteq V$ 
       be a clique in $G$ of size $k$ and $I'$ be the $k' = k$-sized
       subset of $I$ corresponding to the vertices in $V'$. As $V'$ is
       a clique of size $k$, exactly $k(k - 1)/2$ edge neurons in the
       constructed MLP $M$ receive the
       requisite inputs of 1 on both of their endpoints from the vertex
       neurons associated with $I'$. This in turn ensures the output
       neuron produces output 1.  No other possible inputs to the vertex
       neurons not corresponding to elements of $I'$ can change the outputs 
       of these activated edge neurons (and 
       hence the output neuron as well) from 1 to 0. Hence, all completions
       of $I'$ cause $M$ to output 1 and are behaviorally equivalent to $I$
       with respect to $M$.
\item [$\Leftarrow$]: Let $I'$ be a $k' = k$-sized subset of $I$ such that
       all possible completions of $I'$ are behaviorally equivalent to 
       $I$ with respect to $M$, i.e., all such completions cause $M$ to
       output 1. Consider the completion $I''$ of $I'$ in which all non-$I'$
       elements have value 0. The output of $M$ on $I''$ can be 1 (and
       hence equal to the output of $M$ on $I$) only if at least 
       $k(k - 1)/2$ edge neurons have output 1. As all non-$I'$ elements of
       $I''$ have value 0, this means that both endpoints of each of these 
       edge neuron must be connected to elements of $I'$ with output 1, 
       which in turn implies that the $k' = k$ vertices in $G$ corresponding
       to the elements of $I'$ form a clique of size $k$ for $G$.
\end{description}

\noindent
As {\sc Clique} is $NP$-hard \citep{garey1979computers}, the reduction above establishes 
that MSR is also $NP$-hard.  The result follows from the definition of 
$NP$-hardness.
\end{proof}

\begin{theorem}
If $\la cd, \#n_{out}, W_{\max}, B_{\max}, k\ra$-MSR is fixed-parameter 
tractable then \linebreak $FPT = W[1]$.
\label{ThmMSR_fpi1}
\end{theorem}
\begin{proof}
Observe that in the instance of MSR constructed in the reduction in the 
proof of Theorem \ref{ThmMSRCli}, $\#n_{out} = W_{\max} = 1$, $cd = 3$, and 
$B_{\max}$ and $k$ are function of $k$ in the given instance of 
{\sc Clique}. The result then follows from the fact that 
$\la k \ra$-{\sc Clique} is $W[1]$-hard \citep{downeyParameterizedComplexity1999a}.
\end{proof}

\begin{theorem}
$\la \#n_{in} \ra$-MSR is fixed-parameter tractable.
\label{ThmMSR_fpt1}
\end{theorem}
\begin{proof}
Consider the algorithm that generates each possible subset $I'$ of of $I$ 
of size $k$ and for each such $I'$, checks if all possible completions
of $I'$ are behaviorally equivalent to $I$ with respect to $M$. If such an
$I'$ is found, return ``Yes''; otherwise, return ``No''. The number of
possible $I'$ is at most $(\#n_{in})^k \leq (\#n_{in})^{\#n_{in}}$ and the number of possible
completions of any such $I'$ is less than $2^{\#n_{in}}$. Given this, as $M$ can be
run on each completion of $I'$ is time polynomial in the size of the given 
instance of MSR, the above is a fixed-parameter tractable algorithm for MSR
relative to parameter-set $\{ \#n_{in} \}$.
\end{proof}

\begin{theorem}
$\la \#n_{tot} \ra$-MSR is fixed-parameter tractable.
\label{ThmMSR_fpt2}
\end{theorem}
\begin{proof}
Follows from the algorithm in the proof of Theorem \ref{ThmMSR_fpt1} and
the observation that $\#n_{in} \leq \#n_{tot}$.
\end{proof}

\begin{theorem}
$\la cw \ra$-MSR is fixed-parameter tractable.
\label{ThmMSR_fpt3}
\end{theorem}
\begin{proof}
Follows from the algorithm in the proof of Theorem \ref{ThmMSR_fpt1} and
the observation that $\#n_{in} \leq cw$.
\end{proof}

\vspace*{0.15in}

\noindent
Observe that the results in Theorems \ref{ThmMSR_fpi1}--\ref{ThmMSR_fpt3}
in combination with Lemmas \ref{LemPrmFPProp1} and \ref{LemPrmFPProp2}
suffice to establish the parameterized complexity status of MSR relative
to every subset of the parameters listed in Table \ref{TabPrm}.

Let us now consider the polynomial-time cost approximability of MSR. As
MSR is a minimization problem, we cannot do this using reductions from a
maximization problem like {\sc Clique}. Hence we will instead use a
reduction from another minimization problem, namely DS.

\begin{theorem}
If MSR is polynomial-time tractable then $P = NP$.
\label{ThmMSRDS}
\end{theorem}
\begin{proof}
Consider the following reduction from DS to MSR.
Given an instance $\la G = (V,E), k\ra$ of DS, construct the following 
instance $\la M, I, k'\ra$ of MSR:
Let $M$ be an MLP based on $\#n_{tot,g} = 3|V| + 1$ neurons spread
across four layers:

\begin{enumerate}
\item {\bf Input layer}: The input vertex neurons $nv_1, nv_2, \ldots nv_{|V|}$, all of which have bias 0.
\item {\bf Hidden vertex neighbourhood layer I}: The vertex neighbourhood AND neurons $nvnA_1, nvnA_2, \ldots nvnA_{|V|}$, where $nvnA_i$ is an $x$-way AND ReLU gates such that $x = |N_C(v_i)|$.
\item {\bf Hidden vertex neighbourhood layer II}: The vertex neighbourhood NOT neurons $nvnN_1, nvnN_2, \ldots nvnN_{|V|}$, all of which are NOT ReLU gates.
\item {\bf Output layer}: The single output neuron $n_{out}$, which is a
       $|V|$-way AND ReLU gate.
\end{enumerate}

\noindent
The non-zero weight connections between adjacent layers are as follows:

\begin{itemize}
\item Each input vertex neuron $nv_i$, $1 \leq i \leq |V|$, is connected 
       to each vertex neighbourhood AND neuron $nvnA_j$ such that $v_i \in 
       N_C(v_j)$ with weight 1.
\item Each vertex neighbourhood AND neuron $nvnA_i$, $1 \leq i \leq |V|$, is
       connected to its corresponding vertex neighbourhood NOT neuron 
       $nvnN_i$ with weight 1.
\item Each vertex neighbourhood NOT neuron $nvnN_i$, $1 \leq i \leq |V|$, is
       connected to the output neuron $n_{out}$ with weight 1.
\end{itemize}

\noindent
All other connections between neurons in adjacent layers have weight 0.
Finally, let $I$ be the $|V|$-length zero-vector and $k' = k$. Observe that
this instance of MSR can be created in time polynomial in the size of the
given instance of DS, Moreover, the output behaviour of the neurons in 
$M$ from the presentation of input $I$ until the output is generated is as 
follows:

\begin{center}
\begin{tabular}{| c || l |}
\hline
timestep & neurons (outputs) \\
\hline\hline
0 & --- \\
\hline
1 & $nvN_1 (0), nvN_2 (0), \ldots nvN_{|V|} (0)$ \\
\hline
2 & $nvnA_1 (0), nvnA_2 (0), \ldots nvnA_{|V|} (0)$ \\
\hline
3 & $nvnN_1 (1), nvnN_2 (1), \ldots nvnN_{|V|} (1)$ \\
\hline
4 & $n_{out} (1)$ \\
\hline
\end{tabular}
\end{center}
 
We now need to show the correctness of this reduction by proving that the answer for the given instance of DS is ``Yes'' if and only if the answer for the constructed instance of MSR is ``Yes''.
We prove the two directions of this if and only if separately as follows:

\begin{description}
\item [$\Rightarrow$]: Let $V' = \{v'_1, v'_2, \ldots, v'_k\} \subseteq V$ be a dominating set in $G$ of size $k$ and $I'$ be the $k' = k$-sized subset of $I$ corresponding to the vertices in $V'$. As $V'$ is
       a dominating set, each vertex neighbourhood AND neuron receives
       input 0 from at least one input vertex neuron in the set of input 
       vertex neurons associated with $I'$, which in turn ensures that each 
       vertex neighbourhood AND neuron has output 0. This in turn ensures
       that $M$ produces output 1. No other possible inputs to the vertex
       neighbourhood AND neurons can change the output of these neurons from
       0 to 1. Hence, all completions of $I'$ cause $M$ to output 1 and are
       behaviorally equivalent to $I$ with respect to $M$.
\item [$\Leftarrow$]: Let $I'$ be a $k' = k$-sized subset of $I$ such that
       all possible completions of $I'$ are behaviorally equivalent to 
       $I$ with respect to $M$, i.e., all such completions cause $M$ to
       output 1. Consider the completion $I''$ of $I'$ in which all non-$I'$
       elements have value 1. The output of $M$ on $I''$ can be 1 (and
       hence equal to the output of $M$ on $I$) only if
       all vertex neighbourhood NOT neurons output 1, which in turn can
       happen only if all vertex neighbourhood AND gates output 0. However,
       as all non-$I'$ elements of $I''$ have value 1, this means that 
       each vertex neighbourhood AND neuron must be connected to
       at least one element of $I'$, which in turn implies that the
       $k' = k$ vertices in $G$ corresponding to the elements of $I'$ form
       a dominating set of size $k$ for $G$.
\end{description}

\noindent
As DS is $NP$-hard \citep{garey1979computers}, the reduction above establishes that MSR is also $NP$-hard.
The result follows from the definition of $NP$-hardness.
\end{proof}

\begin{theorem}
If MSR has a polynomial-time $c$-approximation algorithm for any constant $c > 0$ then $FPT = W[1]$.
\label{ThmMSR_appi1}
\end{theorem}
\begin{proof}
Recall from the proof of correctness of the reduction in the proof of
Theorem \ref{ThmMSRDS} that a given instance of DS has a dominating set
of size $k$ if and only if the constructed instance of $MSR$ has a subset $I'$ of $I$ of size $k' = k$ such that every possible completion of $I'$ is behaviorally equivalent to $I$ with respect to $M$. 
This implies that, given a polynomial-time $c$-approximation algorithm $A$ for MSR for some constant $c > 0$, we can create a polynomial-time $c$-approximation algorithm for DS by applying the reduction to the given instance $x$ of DS to construct an instance $x'$ of MSR, applying $A$ to $x'$ to create an approximate solution $y'$, and then using $y'$ to create an approximate solution $y$ for $x$ that has the same cost as $y'$. The result then follows from \citealt[Corollary 2]{chenConstantInapproximabilityParameterized2019}, which implies that if DS has a polynomial-time $c$-approximation algorithm for any constant $c > 0$ then $FPT = W[1]$.
\end{proof}

\vspace*{0.15in}

\noindent
Note that this theorem also renders MSR PTAS-inapproximable unless
$FPT = W[1]$.


\section{Probabilistic approximation schemes}
Let us now consider three other types of polynomial-time approximability 
that may be acceptable in situations where always getting the 
correct output for an input is not required:

\begin{enumerate}
\item
algorithms that always run in polynomial time but are frequently correct in that they 
produce the correct output for a given input in all but a small number of 
cases (i.e., the number of errors for input size $n$ is bounded by function $err(n)$) 
\cite{hemaspaandraSIGACTNewsComplexity2012};
\item
algorithms that always run in polynomial time but are frequently correct in that they 
produce the correct output for a given input 
with high probability \cite{motwaniRandomizedAlgorithms1995}; and
\item
algorithms that run in polynomial time with high probability but are always correct
\citep{gillComputationalComplexityProbabilistic1977}.
\end{enumerate}

\noindent
Unfortunately, none of these options are in general open to us, courtesy of the following result.

\vspace*{0.06in}

\begin{theorem}
None of the hard problems in \Cref{tbl:results} are polynomial-time approximable in senses 
(1--3).
\label{ThmProbIbapprox}
\end{theorem}
\begin{proof}
(Sketch) Holds relative to several strongly-believed or established
complexity-class relation conjectures courtesy of the $NP$-hardness of
the problems and the reasoning in the proof of \cite[Result E]{War22}.
\end{proof} 

\section{Supplementary discussion}
\textbf{Search space size versus intrinsic complexity.} Some of our hardness results can be surprising (e.g., fixed-parameter intractability indicating that taming intuitive network and circuit parameters is not enough to make queries feasible).
Other findings might be unsurprising/surprising for the wrong reasons.
Often intractability is assumed based on observing that a problem of interest has an exponential search space.
But this is not a sufficient condition for intractability.
For instance, although the Minimum Spanning Tree problem has an exponential search space, there is enough structure in it that can be exploited to get optimal solutions tractably.
A more directly relevant example is our Quasi-Minimal Circuit problems, which also have exponential search spaces. 
This is a scenario where the typical reasoning in the literature would lead us astray. 
Given our tractability results, jumping to intractability conclusions would miss valuable opportunities to design tractable algorithms with guarantees.

\textbf{Worst-case analysis.} Given our limited knowledge of the problem space of interpretability, worst-case analysis is appropriate to explore what problems might be solvable without requiring any additional assumptions \citep[e.g.,][]{bassanLocalVsGlobal2024a,barcelo_model_2020} and experimental results suggest it captures a lower bound on real-world complexity \citep[e.g.,][]{friedmanInterpretabilityIllusionsGeneralization2024,shiHypothesisTestingCircuit2024,yuFunctionalFaithfulnessWild2024}.  
One possibly fruitful avenue would be to conduct an empirical and formal characterization of learned weights in search of structure that could potentially distinguish conditions of (in)tractability.
This could inform future average-case analyses on plausible distributional assumptions.

\textbf{Strategies for exploring the viability of interpretability queries.} Although we find that many queries of interest are intractable in the general case (and empirical results are in line with this characterization), this should not paralyze real-world efforts to interpret models.
As our exploration of the current complexity landscape shows, reasonable relaxations, restrictions and problem variants can yield tractable queries for circuits with useful properties.
Consider a few out of many possible avenues to continue these explorations.

\textbf{(i)} Faced with an intractable query, we can investigate which parameters of the problem (e.g., network, circuit aspects) might be responsible for the core hardness of the general problem.
If these problematic parameters can be kept small in real-world applications, this can yield a fixed-parameter tractable query which can be answered efficiently in practice.
We have explored some of these parameters, but many more could be, as any aspect of the problem can be parameterized.
For this, a close dialogue between theorists and experimentalists will be crucial, as often empirical regularities suggest which parameters might be fruitful to explore theoretically, and experiments can test whether theoretically conjectured parameters are or can be kept small in practice.

\textbf{(ii)} Generating altogether different circuit query variants is another way of making interpretability feasible.
Our formalization of quasi-minimal circuit problems illustrates the search for viable algorithmic options with examples of tractable problems for inner interpretability.
When the use case is well defined, efficient queries that return circuits with useful affordances for applications can be designed. 
Some circuits (e.g., quasi-minimal circuits, but likely others) might mimic the affordances for prediction/control that ideal circuits have, while shedding the intractability that plagues the latter.

\textbf{(iii)} It could be fruitful to investigate properties of the network output.
Although for some problems, our constructions use step functions in the output layer \citep[following the literature;][]{bassanLocalVsGlobal2024a,barcelo_model_2020}, for many problems we do not or provide alternative proofs without them.
This suggests this is not likely a significant source of complexity.
Another aspect could be the binary input/output, although continuous input/output does not necessarily matter complexity-wise.
Sometimes it does, as in the case of Linear Programming \citep[PTIME;][]{karmarkarNewPolynomialtimeAlgorithm1984} versus 0-1 Integer Programming \citep[NP-complete][]{garey1979computers}, and sometimes it does not, as in Euclidean Steiner Tree \citep[NP-hard, not in NP for technical reasons;][]{garey1979computers} versus Rectilinear Steiner Tree \citep[NP-complete;][]{garey1979computers}.
Still, this is an interesting direction for future work, as it suggests studying the output as an axis of approximation.

\textbf{(iv)} A different path is to design queries that partially rely on mid-level abstractions \citep{vilasPositionInnerInterpretability2024} to bridge the gap between circuits and human-intelligible algorithms \citep[e.g., key-value mechanisms;][]{gevaTransformerFeedForwardLayers2022a, vilasAnalyzingVisionTransformers2024}.

\textbf{(v)} It is in principle possible that real-world trained neural networks possess an internal structure that is somehow benevolent to general (ideal) circuit queries (e.g., redundancy).
In such optimistic scenarios, general-purpose heuristics might work well.
The empirical evidence available, however, speaks against this possibility.
In any case, it will always be important to characterize any ‘benevolent structure’ in the problems such that we can leverage it explicitly to design algorithms with useful guarantees. 




\end{document}